\documentclass[11pt,a4paper,openright,twoside]{report}

\usepackage[american]{babel}
\usepackage[utf8]{inputenc}

\usepackage{fancyhdr}
\usepackage{indentfirst}
\usepackage{graphicx}
\usepackage{newlfont}
\usepackage{xcolor}

\usepackage{stmaryrd}
\usepackage{enumitem}

\usepackage{amssymb}
\usepackage{amsmath}
\usepackage{latexsym}
\usepackage{amsthm}
\usepackage{mathabx}
\usepackage{namedtensor}

\usepackage{hyperref}

\usepackage{caption}
\usepackage{subcaption}

\usepackage[left=2cm,right=2cm,bottom=3cm]{geometry}

\newcounter{mycounter}
\theoremstyle{plain}
\newtheorem{proposition}[mycounter]{Proposition}
\newtheorem{definition}[mycounter]{Definition}

\newtheorem{principle}[mycounter]{Principle}

\theoremstyle{definition}
\newtheorem{remark}[mycounter]{Remark}
\newtheorem{example}[mycounter]{Example}

\newcommand\dblquote[1]{\textquotedblleft #1\textquotedblright}

\begin{document}


\setlist[itemize]{nosep}
\setlist[enumerate]{nosep}


\begin{titlepage}
  \begin{center}
    {
      {\Large
        {\textsc{
          Alma Mater Studiorum $\cdot$ Università di Bologna
        }}
      }
    }
    \rule[0.1cm]{15.8cm}{0.1mm}
    \rule[0.5cm]{15.8cm}{0.6mm}
    {\small
      {\bf 
        SCUOLA DI SCIENZE\\
        Corso di Laurea in Matematica
      }
    }
    \end{center}
    \vspace{15mm}
    \begin{center}
      {\LARGE{\bf TOWARDS A CATEGORICAL }}\\
      \vspace{3mm}
      {\LARGE{\bf FOUNDATION OF DEEP LEARNING:}}\\
      \vspace{3mm}
      {\LARGE{\bf A SURVEY}}\\
      \vspace{19mm}
      {\large
        {\bf
          Thesis in Artificial Intelligence
        }
      }
    \end{center}
    \vspace{40mm}
    \par
    \noindent
    \begin{minipage}[t]{0.47\textwidth}
      {\large
        {\bf
          Advisor:\\
          FABIO ZANASI
        }
      }
    \end{minipage}
    \hfill
    \begin{minipage}[t]{0.47\textwidth}\raggedleft
      {\large
        {\bf 
          Student:\\
          FRANCESCO RICCARDO CRESCENZI
        }
      }
    \end{minipage}
    \vspace{30mm}
    \begin{center}
    {\large
      {\bf 
        Academic Year 2024-2025
      }
    }
  \end{center}
\end{titlepage}


\begin{titlepage}
\thispagestyle{empty}                   
\topmargin=6.5cm                        
\raggedleft                             
\large 
                             
\em                                     
To my beloved\\
Benedetta                   
\newpage                                

\clearpage{\pagestyle{empty}\cleardoublepage}
\end{titlepage}
\pagenumbering{roman}


\chapter*{Abstract}   
\addcontentsline{toc}{chapter}{Abstract}
\rhead[\fancyplain{}{\bfseries Abstract}]{\fancyplain{}{\bfseries\thepage}}\lhead[\fancyplain{}{\bfseries\thepage}]{\fancyplain{}{\bfseries Abstract}}

\section*{English}

The unprecedented pace of machine learning research has lead to incredible advances, but also poses hard challenges. At present, the field lacks strong theoretical underpinnings, and many important achievements stem from \textit{ad hoc} design choices which are hard to justify in principle and whose effectiveness often goes unexplained. Research debt is increasing and many papers are found not to be reproducible. 

This thesis is a survey that covers some recent work attempting to study machine learning categorically. Category theory is a branch of abstract mathematics that has found successful applications in many fields, both inside and outside mathematics. Acting as a \textit{lingua franca} of mathematics and science, category theory might be able to give a unifying structure to the field of machine learning. This could solve some of the aforementioned problems. 

In this work, we mainly focus on the application of category theory to deep learning. Namely, we discuss the use of categorical optics to model gradient-based learning, the use of categorical algebras and integral transforms to link classical computer science to neural networks, the use of functors to link different layers of abstraction and preserve structure, and, finally, the use of string diagrams to provide detailed representations of neural network architectures.

\section*{Italiano}

La velocità senza precedenti con la quale avanza la ricerca nel campo del machine learning ha portato a eccezionali scoperte, ma pone anche sfide impegnative per il futuro.
Al momento, il campo manca di forti basi teoriche, e molte scoperte significative sono frutto di scelte \textit{ad hoc}, difficili da giustificare in teoria, la cui efficacia spesso non si riesce a spiegare. Contemporaneamente, il debito di ricerca continua ad aumentare e molti tentativi di replicazione si concludono in fallimento. 

Questa tesi è una rassegna di alcuni recenti lavori che tentano di analizzare il machine learning dal punto di vista categorico. La teoria delle categorie è una branca della matematica astratta che è stata applicata con successo in molti campi sia dentro che fuori della matematica. Agendo come una lingua franca della matematica e delle scienze, la teoria delle categorie potrebbe dare una struttura unificante al campo del machine learning, il che potrebbe risolvere alcuni dei problemi menzionati.

In queste pagine, ci concentreremo prevalentemente sull'applicazione della teoria delle categorie al campo del deep learning. In particolare, trattaremo l'uso di ottiche categorie per modellare il gradient-based learning, l'uso di algebre categoriche e integral transforms per collegare informatica classica e reti neurali, l'uso di funtori per collegare livelli diversi di astrazione e preservare struttura, e, infine, l'uso di diagrammi a stringhe per rappresentare dettagliatamente reti neurali.

\clearpage{\pagestyle{empty}\cleardoublepage}


\chapter*{Introduction}   
\addcontentsline{toc}{chapter}{Introduction}
\rhead[\fancyplain{}{\bfseries Introduction}]{\fancyplain{}{\bfseries\thepage}}\lhead[\fancyplain{}{\bfseries\thepage}]{\fancyplain{}{\bfseries Introduction}}

\subsection*{The scattered state of machine learning research}

In the last seventy years, machine learning has gone from being a curiosity to being one of the most exciting frontiers of engineering. The ever-increasing amount of research carried out in the field has lead to incredible advancements and, nowadays, machine learning models have a substantial impact many areas of science (\cite{hajkowicz2023artificial}) and society (\cite{khogali2023blended}). Despite its unquestionable successes, the field also faces significant challenges on many levels. Setting aside ethical and societal considerations, which are not discussed in this work, there are also significant scientific concerns. First of all, machine learning, and deep learning in particular, lack strong theoretical underpinnings: at the moment, there is no general mathematical theory of machine learning, and the field looks more like alchemy than science (\cite{gavranovic2024fundamental}, \cite{rahimi2017machine}). Scientists and engineers discover and optimize models by trial and error with no direction and no frame of reference. This leads to a lot wasted time and resources and clearly hampers future progress.

At the same time, bad incentives in the machine learning community lead to issues with the research process itself (\cite{shiebler2021category}). \cite{olah2017research} points out the chaotic state of research and likens it to technical debt in computer engineering, coining the phrase \dblquote{research debt}: bad notation, unclear explanations, and unproven conjectures clutter the machine learning research panorama and make it difficult to actually produce new research. These factors, together with erroneous or missing statistical analysis, also lead to widespread reproducibility issues (\cite{gundersen2022sources}). For instance, \cite{raff2019step} found that more than half of the 255 machine learning papers they considered were not reproducible. 

These issues are especially significant in deep learning, where models often reveal themselves to be brittle black boxes: minor changes in hyperparameters or architectures have large effects on the performance of deep neural networks, and it is often impossible to explain their inner workings proactively (\cite{gavranovic2024fundamental}).
Despite this, deep learning models are being deployed at scale, with little concern for their flaws.  

\subsection*{Category theory as a \textit{lingua franca} of science}

In the most general terms possible, category theory is the branch of mathematics that deals with structure. From its start in algebraic topology, category theory has risen to be one of the most foundational areas of mathematics. Category theory can be seen as a \textit{lingua franca} of mathematics that unites an ever expanding number of mathematical fields under the same fundamental notions. Many have likened the rise of category theory as an extension of the \textit{Erlangen Programme} that unified geometry at the end of the nineteenth century and, from this point of view, categories might be seen as the pulsating heart of mathematics.

More recently, the field of applied category theory has developed, with the aim to applying category theory more generally as a unifying language of science and even engineering. Category theory has found applications in resource theory (\cite{coecke2016mathematical}), database theory (\cite{spivak2012functorial}), quantum mechanics (\cite{abramsky2009categorical}), and much more. Applied category theory can be seen as the study of compositionality and is thus useful wherever compositional structure arise, regardless of the nature of the objects or phenomena at hand. Compositionality here is defined as the property of systems and relationships that \dblquote{can be combined to form new systems or relationships} (\cite{fong2018seven}).

Compositionality is a good property that should be strived for because it offers insight into the fundamental structure of the systems at hand: large compositional systems can be broken up into smaller systems that are easier to understand; conversely, small compositional systems can be assembled into larger systems without requiring any paradigm shift. Machine learning models such as Bayesian networks and neural networks are inherently modular, and so it makes sense to investigate their compositionality using category theory. More generally, many authors think that category theory could provide a unifying structure for machine learning, and thus could help solve or mitigate many of the problems discussed above. Hence, starting from the seminal paper \cite{fong2019backprop}, a large amount of research has explored the intersection between machine learning and category theory. 

\subsection*{General overview}

This thesis was originally meant to be a general survey of the intersection between machine learning and category theory, in the style of \cite{shiebler2021category}. However, we soon realized that the field has grown so much in the last few years that it has become impossible to do justice to every interesting approach in the space and time available to us. Thus, we opted to focus on categorical approaches to deep learning and we decided to focus this thesis around four main ideas, described in the same number of chapters. Each chapter provides detailed descriptions of a few relevant approaches, and then compares such approaches with other related works, which are only briefly touched upon. Brief summaries of the chapters are listed below.

\begin{itemize}
  \item \textbf{Chapter 1: Parametric Optics for Gradient-Based Learning.} \textit{Gradient-based learning can be described and implemented within categories of adequately chosen parametric optics.}
  \item \textbf{Chapter 2: From Classical Computer Science to Neural Networks.} \textit{Category theory can build a bridge between classical computer science and neural networks, offering insight into currently known neural network architectures and informing the design of novel ones.}
  \item \textbf{Chapter 3: Functor Learning.} \textit{Learning functors between categories instead of just morphisms between objects allows the models at hand to preserve structure, which can lead to better learning outcomes.}
  \item \textbf{Chapter 4: Detailed Representations of Neural Networks.} \textit{Appropriate classes of string diagrams can be used to represent neural network architectures with all the detail necessary for implementation.}
\end{itemize}

We emphasize that this work only covers a small amount of the available research: for instance, we do not even touch upon the enormous and enormously relevant field of categorical probabilistic learning, nor do we discuss categorical approaches to explainable artificial intelligence. We do not even cover every categorical approach to neural networks. Nevertheless, we believe that this thesis can offer an interesting taste of the research that is going on in the field.

\subsection*{Target audience and introductory readings}

This work is aimed towards people who already know the basics of both category theory and deep learning. The readers are not required to be extremely proficient in neither, but we will assume that they already have some familiarity with neural network architectures, gradient-based learning, basic categorical definitions, and the theory of symmetric monoidal categories. Regrettably, we do not have enough space to provide our own supplementary material on these subjects, but we encourage interested readers to consult the repository \cite{gavranovic2023category}, for an introduction to category theory, and the online textbook \cite{zhang2021dive}, for an introduction to deep learning. \cite{fong2018seven} is a particularly good starting point of applied category theory. We also recommend that the readers consult the repository \cite{gavranovic2023intersection}, which collects numerous papers on the intersection between machine learning and category theory, some of which are discussed in the following chapters.

\begin{remark}
  For simplicity's sake, we ignore any size concern in our treatment of categorical concepts.
\end{remark}

\clearpage{\pagestyle{empty}\cleardoublepage}


\tableofcontents
\rhead[\fancyplain{}{\bfseries\leftmark}]{\fancyplain{}{\bfseries\thepage}} \lhead[\fancyplain{}{\bfseries\thepage}]{\fancyplain{}{\bfseries Table of Contents}}
\clearpage{\pagestyle{empty}\cleardoublepage}


\listoffigures  
\rhead[\fancyplain{}{\bfseries\leftmark}]{\fancyplain{}{\bfseries\thepage}} \lhead[\fancyplain{}{\bfseries\thepage}]{\fancyplain{}{\bfseries List of figures}}
\clearpage{\pagestyle{empty}\cleardoublepage}


\chapter{Parametric Optics for Gradient-Based Learning}
\lhead[\fancyplain{}{\bfseries\thepage}]{\fancyplain{}{\bfseries\rightmark}}
\pagenumbering{arabic}

Despite the unquestionable success that gradient-based deep learning has enjoyed in recent years, the field is still both young and poorly understood. As mentioned in the introduction, the lack of theoretical underpinnings means that good performance is highly dependent on \textit{ad hoc} choices and empirical heuristics leading to brittleness and poorly understood phenomena (\cite{shiebler2021category}, \cite{gavranovic2024fundamental}). The ever-growing complexity of deep learning models poses significant challenges both in terms of optimization (\cite{elliott2018simple}) and architectural design (\cite{gavranovicposition}), and, while there are a number of general purpose deep learning libraries that automatically implement backpropagation and provide tools for designing a wide variety of neural networks, these tools often rely on inelegant machinery difficult to parallelize (\cite{elliott2018simple}). Given the ever-increasing role gradient based learning plays in the sciences, in industry, and in everyday life, solving these issues is of the utmost importance. 

Hence, it would be desirable to develop a mathematically structured framework for gradient-based learning able to act as a bridge between low-level automatic differentiation and high level architectural specifications (\cite{gavranovic2024fundamental}). The great number of architectures developed in recent years and the inherently modular structure of deep neural networks call for a model which is general (that is, not dependent on a specific differentiation algorithm or a specific optimizer) and compositional (that is, we should be able to predict the behavior of the entire model if the behavior of each part is known). \cite{cruttwell2022categorical} and \cite{gavranovic2024fundamental} propose a promising combination of differential categories, parametrization and optics as a full-featured gradient-based framework able to challenge established tools and attack open problems. In this chapter, we illustrate such framework  and part of its mathematical foundations. 

\section{Categorical toolkit}

Learning neural networks have two important properties: they depend on parameters and information flows through them bidirectionally (forward propagation and back propagation). Any aspiring categorical model of gradient-based learning must take these two aspects into consideration. A number of authors (among which \cite{cruttwell2022categorical} and \cite{gavranovic2024fundamental}) propose the $\mathbf{Para}$ construction as a categorical model of parameter dependence and various categories of \textit{optics} as the right categorical abstraction for bidirectionality.

\subsection{Actegories}

Before we can deal with parametric maps, we need to find a way to glue input spaces to parameter spaces, so that such maps have well-defined domains. One common strategy is to provide the category at hand with a monoidal structure. However, monoidal products can only combine elements within the same underlying category. Since (co)parameters are sometimes taken from spaces that are different in nature from the input and output spaces, a more general mathematical tool is needed: namely, \textit{actegories} (see the survey \cite{capucci2022actegories} for a thorough treatment of the subject).  Actegories are actions of symmetric monoidal categories on other categories. For brevity's sake, we will only give an incomplete definition (see \cite{capucci2022actegories} or \cite{gavranovic2024fundamental} for further information).

\begin{definition}[Actegory]
  Let $(\mathcal{M},I,\otimes)$ be a strict symmetric monoidal category. A $\mathcal{M}$-actegory is a tuple $(\mathcal{C}, \bullet, \eta, \mu)$, where $\mathcal{C}$ is a category, $\bullet: \mathcal{M} \times \mathcal{C} \to \mathcal{C}$ is a functor, and $\eta$ and $\mu$ are natural isomorphisms enforcing $C \stackrel{\eta_C}{\cong}I \bullet C$ and $(M \otimes N) \bullet C \stackrel{\mu_{M,N,C}}{\cong} (M \bullet (N \bullet C))$.
  The isomorphisms $\eta$ and $\mu$ must also satisfy coherence conditions. If $\eta$ and $\mu$ are identical transformations, we say that the actegory is strict.
\end{definition}

\begin{remark}
  Although the requirement for strictness is somewhat restrictive, we will proceed under the assumption that the actegories we encounter are strict to streamline notation.
\end{remark}

We will also be interested in actegories that interact with the monoidal structure of the underlying category.

\begin{definition}[Monoidal actegory]
  Let $(\mathcal{M},I,\otimes)$ be a strict symmetric monoidal category and let $(\mathcal{C}, \bullet, \eta, \mu)$ be a strict actegory. Suppose $\mathcal{C}$ has a monoidal structure $(J,\boxtimes)$. Then we say that $(\mathcal{C}, \bullet)$ is monoidal if the underlying functor $\bullet$ is monoidal and the underlying natural transformations $\eta$ and $\mu$ are also monoidal.
\end{definition}

We may also be interested in studying the interaction between actegorical structures and endofunctors. This interaction can happen owing to a natural transformation known as strength. We will not provide coherence diagrams in the definition below for the sake of brevity, but the interested reader can find more detail in \cite{gavranovicposition}. The paper also provides a definition of actegorical strong monad, which is a very similar concept.

\begin{definition}[Actegorical strong functor]
  Let $(\mathcal{C}, \bullet)$ be an $\mathcal{M}$-actegory. A strong actegorical endofunctor on $(\mathcal{C}, \bullet)$ is a pair $(F, \sigma)$ where $F: \mathcal{C} \to \mathcal{C}$ is an endofunctor and $\sigma$ is a natural transformation with components $\sigma_{P,A}: P \bullet F(A) \to F(P \bullet A)$ which satisfies a few coherence conditions that we do not list here.
\end{definition}

\subsection{The $\mathbf{Para}$ construction}

Suppose we have an $\mathcal{M}$-actegory $(\mathcal{C}, \bullet)$. We wish to study maps in $\mathcal{C}$ which are parametrized using objects of $\mathcal{M}$, that is, maps in the form $P \bullet A \to B$. We are not just interested in the maps by themselves, but also in their compositional structure. Thus, we abstract away the details by defining a new category $\mathbf{Para}_{\bullet}(\mathcal{C})$ (first introduced in simplified form by \cite{fong2019backprop}). Since we also want to formalize the role of reparametrization, we actually construct $\mathbf{Para}_{\bullet}(\mathcal{C})$ as a bicategory, so that a $0$-cell $A$ can serve as an input/output space, a $1$-cell $(P,f)$ can serve as a parametric map, and, finally, a $2$-cell $r$ can serve as a reparametrization.

\begin{figure}[h]
  \begin{center}
    \begin{subfigure}{0.3\textwidth}
      \includegraphics[width=\textwidth]{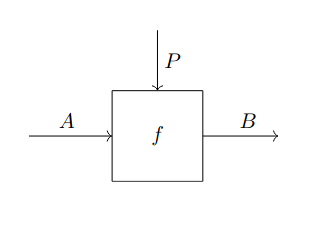}
      \caption{}
    \end{subfigure}
    \begin{subfigure}{0.5\textwidth}
      \includegraphics[width=\textwidth]{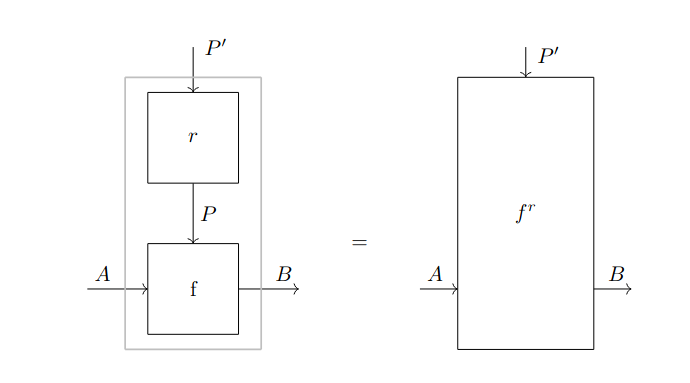}
      \caption{}
    \end{subfigure}
    \begin{subfigure}{0.7\textwidth}
      \includegraphics[width=\textwidth]{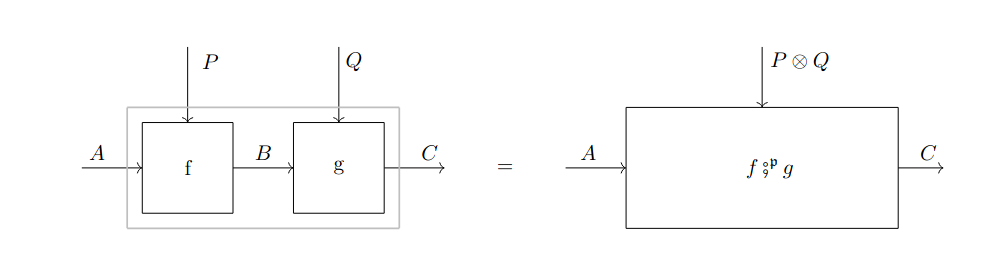}
      \caption{}
    \end{subfigure}                 
    \caption[Parametric morphisms]{String diagrams representing (a) a parametric morphism, (b) a reparametrization of a parametric morphism, (c) a composition of parametric morphisms. (Images taken from \cite{gavranovic2024fundamental}.)}\label{fig: paramorphisms}
  \end{center}
\end{figure}

\begin{definition}[$\mathbf{Para}_{\bullet}(\mathcal{C})$]
  \label{def: para}
  Let $(\mathcal{C}, \bullet)$ be an $\mathcal{M}$-actegory. Then, we define $\mathbf{Para}_{\bullet}(\mathcal{C})$ as the bicategory whose components are as follows.
  \begin{itemize}
    \item The $0$-cells are the objects of $\mathcal{C}$.
    \item The $1$-cells are pairs $(P,f): A \to B$, where $P : \mathcal{C}$ and $f: P \bullet A \to B$.
    \item The $2$-cells come in the form $r: (P,f) \Rightarrow (Q,g)$, where $r: P \to Q$ is a morphism in $\mathcal{C}$. $r$ must also satisfy a naturality condition.
    \item The $1$-cell composition law is
    \[(P,f) \fatsemi (Q,g) = (Q \otimes P, (Q \bullet f)  \fatsemi g).\]
    \item The horizontal and vertical $2$-cell composition laws are respectively given by parallel and sequential composition in $\mathcal{M}$. 
  \end{itemize}
  It is quite handy to represent the cells of $\mathbf{Para}_{\bullet}(\mathcal{C})$ using the string diagram notation illustrated in \textit{Fig. \ref{fig: paramorphisms}}. The $\mathbf{Para}$ construction has a dual $\mathbf{coPara}$ construction whose $1$-cells $f:\mathbf{coPara}_{\bullet}(\mathcal{C})(A,B)$ take the form $(P,f)$, where $f: A \to P \bullet B$. Cells in $\mathbf{coPara}_{\bullet}(\mathcal{C})$ can also be represented with appropriate string diagrams. The reader can find a complete definition in \cite{gavranovic2024fundamental}.
\end{definition}

It is shown in \cite{gavranovic2024fundamental} that $\mathbf{Para}_{\bullet}(\mathcal{C})$ is actually a $2$-category if the underlying actegory is strict. Assuming this is the case (as we do in this thesis), we can use a functor $F: \mathbf{Cat} \to \mathbf{Set}$ to quotient out the $2$-categorical structure and turn $\mathbf{Para}_{\bullet}(\mathcal{C})$ into a $1$-category $F_{*}(\mathbf{Para}_{\bullet}(\mathcal{C}))$. Here, $F_{*}: \mathbf{2Cat} \to \mathbf{Set}$ is the change of enrichment basis functor induced by $F$. This meaningfully recovers the $1$-categorical perspective of \cite{fong2019backprop}.

Both $\mathbf{Para}_{\bullet}(\mathcal{C})$ and $\mathbf{coPara}_{\bullet}(\mathcal{C})$ can be given a monoidal structure if $(\mathcal{C}, \bullet)$ is a monoidal actegory. This is extremely important because it allows us to compose (co)parametric morphisms both in sequence and in parallel. Once again, more detail can be found in \cite{gavranovic2024fundamental}.

\begin{remark}
  Another way to parametrize morphisms is the coKleisli construction. As noted by \cite{gavranovic2024fundamental}, the main difference between $\mathbf{coKl}$ and $\mathbf{Para}$ is that the parametrization offered by $\mathbf{coKl}$ is global, while the parametrization offered by $\mathbf{Para}$ is local: all morphisms in $\mathbf{coKl}(X \times -)$ must take a parameter in $X$, while different morphisms of $\mathbf{Para}(\mathcal{C})$ admit different parameter spaces. Nevertheless, the two constructions are related, and the former can be embedded into the latter.
\end{remark}

If we take a parametrized category $\mathbf{Para}_\bullet(\mathcal{C})$ and we restrict our attention to morphisms parametrized with the monoidal identity $I$, we get back the original category $\mathcal{C}$. This is expressed by the following proposition (\cite{gavranovic2024fundamental}).

\begin{proposition}
  \label{prop: embeddingth}
  Let $(\mathcal{C}, \bullet)$ be an $\mathcal{M}$-actegory. Then, there exists an identity-on-objects pseudofunctor $\gamma: \mathcal{C} \to \mathbf{Para}_\bullet(\mathcal{C})$ that maps $f \mapsto (I,f)$. If $\mathcal{M}$ is strict, this is a $2$-functor.
\end{proposition}

\subsection{Optics}

Modelling bidirectional flows of information is not only useful in machine learning, but also in game theory, database theory, and more. As such, categorical tools for bidirectionality have been sought after for a long time: in particular, a great deal of effort has been devoted to the development of lens theory. Lenses have then been generalized into optics (see e.g. \cite{riley2018categories}) to subsume other tools such as prisms and traversals into a single framework. Finally, there have also been various attempts to generalize optics (see e.g. \cite{clarke2024profunctor} for a definition of mixed optics). We will introduce lenses and optics, and focus on the generalization of optics introduced by \cite{gavranovic2024fundamental}: \textit{weighted optics}. 

As stated in \cite{gavranovic2024fundamental}, there is no standard definition of lens, and different authors opt for different \textit{ad hoc} definitions that best suit their purposes. We will borrow the perspective of \cite{cruttwell2022categorical}.

\begin{figure}[h]
  \begin{center}
    \begin{subfigure}{0.3\textwidth}
      \includegraphics[width=\textwidth]{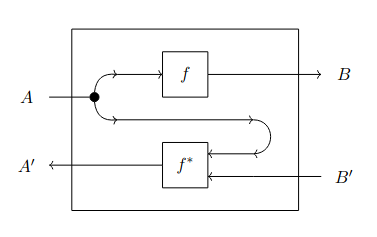}
      \caption{}
    \end{subfigure}
    \begin{subfigure}{0.6\textwidth}
      \includegraphics[width=\textwidth]{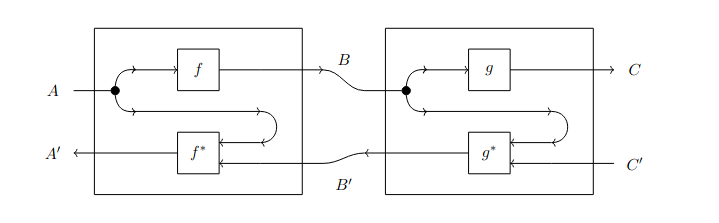}
      \caption{}
    \end{subfigure}                 
    \caption[Lenses]{String diagrams representing (a) a lens $\left(\begin{smallmatrix} f \\ f^* \end{smallmatrix}\right)$, (b) the composition of two lenses $\left(\begin{smallmatrix} f \\ f^* \end{smallmatrix}\right)$ and $\left(\begin{smallmatrix} g \\ g^* \end{smallmatrix}\right)$. (Images taken from \cite{cruttwell2022categorical}.)}
    \label{fig: lenses}
  \end{center}
\end{figure}

\begin{definition}[Lenses]
  \label{def: lenses}
  Let $\mathcal{C}$ be a Cartesian category. Then, $\mathbf{Lens}(\mathcal{C})$ is the category defined by the following data:
  \begin{itemize}
    \item an object of $\mathbf{Lens}(\mathcal{C})$ is a pair $\left(\begin{smallmatrix} A \\ A' \end{smallmatrix}\right)$ of objects of $\mathcal{C}$; 
    
    \item a $\left(\begin{smallmatrix} A \\ A' \end{smallmatrix}\right) \to \left(\begin{smallmatrix} B \\ B' \end{smallmatrix}\right)$ morphism (or lens) is a pair $\left(\begin{smallmatrix} f \\ f' \end{smallmatrix}\right)$ of morphisms of $\mathcal{C}$ such that $f: A \to B$ and $f': A \times B' \to A'$; $f$ is known as the forward pass of the lens $\left(\begin{smallmatrix} f \\ f' \end{smallmatrix}\right)$, whereas $f'$ is known as the backward pass;
    
    \item given $\left(\begin{smallmatrix} A \\ A' \end{smallmatrix}\right) : \mathbf{Lens}(\mathcal{C})$, the associated identity lens is $\left(\begin{smallmatrix} 1_A \\ \pi_1 \end{smallmatrix}\right)$;
    
    \item the composition of $\left(\begin{smallmatrix} f \\ f' \end{smallmatrix}\right): \left(\begin{smallmatrix} A \\ A' \end{smallmatrix}\right) \to \left(\begin{smallmatrix} B \\ B' \end{smallmatrix}\right)$ and $\left(\begin{smallmatrix} g \\ g' \end{smallmatrix}\right): \left(\begin{smallmatrix} B \\ B' \end{smallmatrix}\right) \to \left(\begin{smallmatrix} C \\ C' \end{smallmatrix}\right)$ is
    \[\left(\begin{smallmatrix} f \fatsemi g \\ \langle \pi_0, \langle \pi_0 \fatsemi f, \pi_1 \rangle \fatsemi g' \rangle \fatsemi f' \end{smallmatrix}\right).\]
  \end{itemize}
  Lenses can be represented using the string diagrams illustrated in \textit{Fig. \ref{fig: lenses}}.
\end{definition}

Lenses are a powerful tool, but they cannot be used to model all situations: for instance, lenses cannot be used if we wish to be able to choose not to interact with the environment depending on the input, or if we would like to reuse values computed in the forward pass for further computation in the backward pass.

Optics generalize lenses by weakening the link between forward and backward passes, and by replacing the Cartesian structure of the underlying category with a simpler symmetric monoidal structure. In an optic over $\mathcal{C}$, an object $M : \mathcal{C}$ acts as an inaccessible residual space transferring information between the upper components and the lower component.
We provide the definition given by \cite{riley2018categories}\footnote{\cite{riley2018categories}also provides a more versatile (but more sophisticated) definition of optics that relies on coends. Under the coend formalism,
\[\mathbf{Optic}(\mathcal{C})\left(\left(\begin{smallmatrix} A \\ A' \end{smallmatrix}\right),\left(\begin{smallmatrix} B \\ B' \end{smallmatrix}\right)\right) = \int^{M : \mathcal{C}}\mathcal{C}(A,M \otimes B) \times \mathcal{C}(M \otimes B',A').\]}.

\begin{definition}[Optics]
  Let $(\mathcal{C}, I, \otimes, \lambda, \rho, \alpha)$ be a symmetric monoidal category (we make the unitors and associators explicit for later use). Then, $\mathbf{Optic}(\mathcal{C})$ is the category defined by the following data:
  \begin{itemize}
    \item an object of $\mathbf{Optic}(\mathcal{C})$ is a pair $\left(\begin{smallmatrix} A \\ A' \end{smallmatrix}\right)$ of objects of $\mathcal{C}$; 
    \item a $\left(\begin{smallmatrix} A \\ A' \end{smallmatrix}\right) \to \left(\begin{smallmatrix} B \\ B' \end{smallmatrix}\right)$ morphism (or optic) is a pair $\left(\begin{smallmatrix} f \\ f' \end{smallmatrix}\right)$ of morphisms of $\mathcal{C}$ such that $f: A \to M \otimes B$ and $f': M \otimes B' \to A'$, where $M : \mathcal{C}$ is known as residual space; such pairs $\left(\begin{smallmatrix} f \\ f' \end{smallmatrix}\right)$ are also quotiented by an equivalence relation that allows for reparametrization of the residual space and effectively makes it inaccessible;
    \item the identity on $\left(\begin{smallmatrix} A \\ A' \end{smallmatrix}\right)$ is the optic represented by $\left(\begin{smallmatrix}\lambda_A^{-1}\\ \lambda_{A'}\end{smallmatrix}\right)$.
  \end{itemize}
  Refer to \cite{riley2018categories} or \cite{gavranovic2024fundamental} for the more information about the composition of optics and the representation of optics with string diagrams.
\end{definition}

Lenses come up as a special case of optics (\cite{riley2018categories}), and optics do solve some of the issues we have with lenses. However, optics are not perfect either: for instance, \cite{gavranovic2024fundamental} points out that optics cannot be used in cases where we ask that the forward pass and backward pass are different kind of maps, as they are both forced to live in the same category. Thus, a further layer of generalization is useful: namely, weighted optics.

\subsection{Weighted optics}

Before we define weighted optics, we need to introduce a new tool to our toolbox: the category of elements of a functor.

\begin{definition}[Elements of a functor]
  Let $F: \mathcal{C} \to \mathbf{Set}$ be a functor. We define $\mathbf{El}(F)$ as the category with the following data: (i) the objects of $\mathbf{El}(F)$ are pairs $(C, x)$ where $C : \mathcal{C}$ and $x : F(C)$; (ii) the $(C, x) \to (D, y)$ morphisms in $\mathbf{El}(F)$ are the morphisms $f: C \to D$ in $\mathcal{C}$ such that $F(f)(x) = y$.
\end{definition}

\cite{gavranovic2024fundamental} studies $\mathcal{B}$-actegories $(\mathcal{C}, \bullet)$, which are then reparametrized so that the acting category becomes $\mathcal{E} = \mathbf{El}(W)$ for some weight functor $W: \mathcal{B}^{\mathrm{op}} \to \mathbf{Set}$ (which is to be specified). The reparametrization takes place thanks to the opposite of the forgetful functor $\pi_W: \mathcal{E} \to \mathcal{B}^{\mathrm{op}}$, which maps $(B,x) \mapsto B$. Hence, we consider the action
\[\bullet^{\pi_W^{\mathrm{op}}} = \mathbf{El}(W)^{\mathrm{op}} \times \mathcal{C} \stackrel{\pi_W^{\mathrm{op}} \times \mathcal{C}}{\longrightarrow} \mathcal{B} \times \mathcal{C} \stackrel{\bullet}{\longrightarrow} \mathcal{C}.\]
We are finally ready to define weighted optics\footnote{Weighted optics also admit a coend definition. Refer to \cite{gavranovic2024fundamental} for more information.}.

\begin{definition}[Weighted $\mathbf{coPara}$]
  If $W$ is a weight functor as above and $(\mathcal{C}, \bullet)$ is a $\mathcal{B}$-actegory, we define
  \[\mathbf{coPara}^W_{\bullet}(\mathcal{C}) = \pi_{0*}(\mathbf{coPara}_{\bullet^{\pi_W^{\mathrm{op}}}}(\mathcal{C})),\]
  where $\pi_{0*}$ is the enrichment base change functor generated by the connected component functor $\pi_{0}: \mathbf{Cat} \to \mathbf{Set}$. More explicitly, $\pi_{0*}$ quotients out the connections provided by reparametrizations.
\end{definition}

\begin{definition}[Weighted optics]
  Suppose $(\mathcal{C}, \bullet)$ is an $\mathcal{M}$-actegory, suppose $(\mathcal{D}, \sqbullet)$ is an $\mathcal{M}'$-actegory, and suppose $W: \mathcal{M}^{\mathrm{op}} \times \mathcal{M}' \to \mathbf{Set}$ is a lax monoidal functor.
  We define the category of $W$-weighted optics over the product actegory $\left(\mathcal{C}\times\mathcal{D}^{\mathrm{op}},\left(\begin{smallmatrix}\bullet\\\sqbullet^{\mathrm{op}}\end{smallmatrix}\right)\right)$ as
  \[\mathbf{Optic}^W_{\left(\begin{smallmatrix}\bullet\\\sqbullet\end{smallmatrix}\right)} = \mathbf{coPara}^W_{\left(\begin{smallmatrix}\bullet\\\sqbullet^{\mathrm{op}}\end{smallmatrix}\right)}(\mathcal{C}\times\mathcal{D}^{\mathrm{op}}).\]
\end{definition}

\begin{figure}[h]
  \begin{center}
    \includegraphics[width=0.7\textwidth]{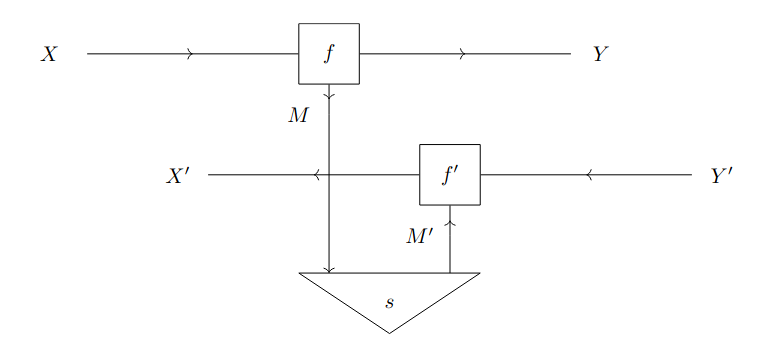}             
    \caption[Weighted optic]{String diagram representing the inner workings of a weighted optic. (Image taken from \cite{gavranovic2024fundamental}.)}
    \label{fig: weightedoptic}
  \end{center}
\end{figure}

The definition is very dense and deserves some explanation. Fist of all, we assume that $W$ maps $\left(\begin{smallmatrix}M \\ M'\end{smallmatrix}\right)$ to a set of maps $s: M \to M'$. If that's the case, a $\left(\begin{smallmatrix}X\\X'\end{smallmatrix}\right) \to \left(\begin{smallmatrix}Y\\Y'\end{smallmatrix}\right)$ map is a triplet $\left(\left(\begin{smallmatrix}M\\M'\end{smallmatrix}\right),s,\left(\begin{smallmatrix}f\\f'\end{smallmatrix}\right)\right)$, where $M$ is the forward residual, $M'$ is the backward residual, $s: M \to M'$ links the two residuals, $f: X \to M \bullet Y$ is the forward pass, and $f': M' \bullet Y' \to X'$ is the backward pass.
The triplets are also quotiented with respect to reparametrization, which makes the residual spaces effectively inaccessible (as it happens in the case of ordinary optics). We can get a clear "operational" understanding of how a weighted optic works looking at an associated string diagram (see \textit{Fig. \ref{fig: weightedoptic}}): data from $X$ flows through the forward map, which computes an output in $Y$ and a forward residual in $M$. Such forward residual is then converted into a backward residual in $M'$ by the map $s$, which is provided by the weight functor. Finally, the backward residual is used by $f'$, together with input from $Y'$, in order to compute a value in $X'$. A full account of the composition law for weighted optics can be found on \cite{gavranovic2024fundamental}.
As stated in \cite{gavranovic2024fundamental}, since $\mathbf{coPara}$ can be given a monoidal structure, we can also give $\mathbf{Optic}^W_{\left(\begin{smallmatrix}\bullet\\\sqbullet\end{smallmatrix}\right)}$ one such structure as long as the underlying actegories are monoidal and the weight functor $W$ is braided monoidal.

The advantages of weighted optics over ordinary optics are clear: when dealing with weighted optics, we are no longer forced to take reverse maps from the same category as the forward maps. The action on the category of forward spaces is now separated from the action on the category of backward spaces, and the link between the two actions is provided by an external functor. Such modular approach provides a great deal of conceptual clarity and flexibility, more than regular optics or lenses can provide on their own. It is also shown in \cite{gavranovic2024fundamental} that weighted optics are indeed a generalization of optics. In particular, it is shown that the lenses in \textit{Def. \ref{def: lenses}} are the specialized weighted optics obtained when $\mathcal{C} = \mathcal{D}$ is Cartesian and the actegories are given by the Cartesian product. More generally, \cite{gavranovic2024fundamental} claims that - to the best of the author's knowledge - all definitions of lenses currently used in the literature are subsumed by the definition of mixed optics (see \cite{clarke2024profunctor}), which are themselves a special case of weighted optics. Hence, all lenses are weighted optics.

\cite{gavranovic2024fundamental} goes on to apply the $\mathbf{Para}$ construction onto weighted optics, obtaining \textit{parametric weighted optics}, which are proposed as a full-featured model for deep learning. The author conjectures that "weighted optics provide a good denotational and operational semantics for differentiation". In its full, generality, this is still an unproven conjecture. However, restricting our attention to a special class $\mathbf{Lens}_A$ of lenses with an additive backward passes yields a formal theory of \dblquote{backpropagation through structure} (\cite{gavranovic2024fundamental}), which will be illustrated in the rest of the chapter, after a short digression on differential categories.

\subsection{Differential categories}
\label{subsec: diffcats}

Modelling gradient-based learning obviously requires a setting where differentiation can take place. Although it is tempting to directly employ smooth functions over Euclidean spaces, recent research has shown that there are tangible advantages in working with generalized differential combinators that extend the notion of derivative to polynomial circuits (\cite{wilson2022categories}, \cite{wilson2021reverse}), manifolds (\cite{peng2021hyperbolic}), complex spaces (\cite{bassey2021survey}), and so on.  Thus, it makes sense to work with an abstract notion of derivative which can then be appropriately implemented depending on the requirements at hand.

One approach to this problem involves the explicit definition of two kinds of \textit{differential categories}: Cartesian differential categories (first introduced in \cite{blute2006differential}) and Cartesian reverse differential categories (first introduced by \cite{cockett2019reverse}). The former allow for forward differentiation, while the latter allow for reverse differentiation. We will omit the defining axioms for the sake of brevity, but the reader can find complete definitions in \cite{cockett2019reverse}.

\begin{definition}[Cartesian differential category]
  \label{def: cdc}
  A Cartesian differential category (CDC) $\mathcal{C}$ is a Cartesian left-additive category where a differential combinator $\mathrm{D}$ is defined. Such differential combinator must take a morphism $f: A \to B$ and return a morphism $\mathrm{D}[f]: A \times A \to B$, which is known as the derivative of $f$. The combinator $\mathrm{D}$ must satisfy a number of axioms. 
\end{definition}

\begin{definition}[Cartesian reverse differential category]
  A Cartesian reverse differential category (CRDC) $\mathcal{C}$ is a Cartesian left-additive category where a reverse differential combinator $\mathrm{R}$ is defined. Such reverse differential combinator must take a morphism $f: A \to B$ and return a morphism $\mathrm{R}[f]: A \times B \to A$, which is known as the reverse derivative of $f$. The combinator $\mathrm{R}$ must satisfy a number of axioms. 
\end{definition}

\begin{example}
  \label{ex: smooth}
  Consider the category $\mathbf{Smooth}$ of Euclidean spaces and smooth functions.
  $\mathbf{Smooth}$ is a both a CDC and a CRDC. In fact, if $\mathcal{J}_f$ is the Jacobian matrix of a smooth morphism $f$,
  \[\mathrm{D}[f]: (x,v) \mapsto \mathcal{J}_f(x)v\]
  and
  \[\mathrm{R}[f]: (x,y) \mapsto \mathcal{J}_f(x)^Ty\]
  induce well-defined combinators $\mathrm{D}$ and $\mathrm{R}$. This is only a partial coincidence: as shown in \cite{cockett2019reverse} CRDCs are always CDCs under a canonical choice of differential combinator. The converse, however, is generally false.
\end{example}

As it turns out, forward differentiation tends to be less efficient when dealing with neural networks that come up in practice (\cite{elliott2018simple}), so CDCs are not extremely useful when studying deep learning. CRDCs, on the other hand, have been applied to great success (see e.g. \cite{cruttwell2022categorical}). As shown in \cite{wilson2022categories}, a large supply of CRDCs can be obtained by providing the generators of a finitely presented Cartesian left-additive category with associated reverse derivatives (as long as the choices of reverse derivative are consistent).
Moreover, CRDCs have been recently generalized by \cite{gavranovic2024fundamental} to coalgebras associated with copointed endofunctors, which could also increase the number of known CRDCs in the future. The rest of this section is devoted to this generalization.

It is shown in \cite{gavranovic2024fundamental} that there is a particular class of weighted optics which is useful for reverse differentiation, being able to represent both maps (through forward passes) and the associated reverse derivatives (through backward passes). Moreover, such weighted optics can be represented as lenses in the sense of \textit{Def \ref{def: lenses}}, which means that their inner workings can be pictured in a simple, intuitive way.

\begin{definition}[Additively closed Cartesian left-additive category]
  A Cartesian left-additive category $\mathcal{C}$ is an additively closed Cartesian left-additive category (ACCLAC) if and only if the following are true:
  \begin{itemize}
    \item the subcategory $\mathbf{CMon}(\mathcal{C})$ of additive maps has a closed monoidal structure $(I, \otimes)$;
    \item the embedding $\iota: \mathbf{CMon}(\mathcal{C}) \to \mathcal{C}$ is a lax monoidal funtor with respect to the aforementioned structure of $\mathbf{CMon}(\mathcal{C})$ and the Cartesian structure of $\mathcal{C}$.
  \end{itemize}
\end{definition}

Then, we can define the category of lenses with backward passes additive in the second component.

\begin{definition}
  Let $\mathcal{C}$ be an ACCLAC with Cartesian structure is $(1, \times)$ and whose subcategory $\mathbf{CMon}(\mathcal{C})$ has monoidal structure $(I, \otimes)$. Then, we define
  \[\mathbf{Lens}_A(\mathcal{C}) = \mathbf{Optic}_{\left(\begin{smallmatrix}\times \\ \otimes\end{smallmatrix}\right)}^{\mathcal{C}(-,\iota(-))}.\]
\end{definition}

As argued in \cite{gavranovic2024fundamental}, the symbol $\mathbf{Lens}_A$ is justified because one such optic of type $\left(\begin{smallmatrix}X \\ X'\end{smallmatrix}\right) \to \left(\begin{smallmatrix}Y \\ Y'\end{smallmatrix}\right)$ can be concretely represented as a lens with forward pass $f : \mathcal{C}(X,Y)$ and backward pass $f' : \mathcal{C}(X \times Y',X')$, which is the approach we illustrate in this thesis.  Nevertheless, some potential expressivity is lost when passing from weighted optic composition to concrete lens composition. In particular, if we operated with optics, we would be able to implement backpropagation without resorting to gradient checkpointing, which is not possible if we use lenses (\cite{gavranovic2024fundamental}). 

The generalization mentioned above is possible because $\mathbf{Lens}_A$ is an endofunctor.

\begin{definition}
  We defined $\mathbf{CLACat}$ as the category whose objects are Cartesian left-additive categories and whose morphisms are Cartesian left-additive functors (see e.g.\cite{blute2006differential}).
\end{definition}

\begin{proposition}
  If $\mathcal{C} : \mathbf{CLACat}$, then $\mathbf{Lens}_A(\mathcal{C}) : \mathbf{CLACat}$.
\end{proposition}
\begin{proof}
  The Cartesian structure on $\mathbf{Lens}_A(\mathcal{C})$ is given by $\left(\begin{smallmatrix}X \\ X'\end{smallmatrix}\right) \times \left(\begin{smallmatrix}Y \\ Y'\end{smallmatrix}\right) = \left(\begin{smallmatrix}X\times Y \\ X' \times Y'\end{smallmatrix}\right)$ and by the initial object $\left(\begin{smallmatrix}1 \\ 1\end{smallmatrix}\right)$. The monoidal structure on each $\left(\begin{smallmatrix}X \\ X'\end{smallmatrix}\right)$ is given by the unit $0_{\left(\begin{smallmatrix}X \\ X'\end{smallmatrix}\right)} = \left(\begin{smallmatrix}0_A \\ !_{1 \times A'}\end{smallmatrix}\right): \left(\begin{smallmatrix}1 \\ 1\end{smallmatrix}\right) \to \left(\begin{smallmatrix}X \\ X'\end{smallmatrix}\right)$ and by the multiplication $+_{\left(\begin{smallmatrix}X \\ X'\end{smallmatrix}\right)} = \left(\begin{smallmatrix}+_A \\ \pi_2 \fatsemi \Delta_{A'}\end{smallmatrix}\right): \left(\begin{smallmatrix}X \times X \\ X' \times X'\end{smallmatrix}\right) \to \left(\begin{smallmatrix}X \\ X'\end{smallmatrix}\right)$.
\end{proof}

\begin{proposition}
  $\mathbf{Lens}_A: \mathbf{CLACat} \to \mathbf{CLACat}$ is a functor.
\end{proposition}
\begin{proof}
  Given a Cartesian left-additive functor $F: \mathcal{C} \to \mathcal{D}$, we can define $\mathbf{Lens}_A(F)$ as the functor that maps $\left(\begin{smallmatrix}X  \\ X' \end{smallmatrix}\right) \mapsto \left(\begin{smallmatrix} F(X) \\ F(X') \end{smallmatrix}\right)$ and maps $\left(\begin{smallmatrix}f  \\ f' \end{smallmatrix}\right) \mapsto \left(\begin{smallmatrix} F(f) \\ \underline{f}' \end{smallmatrix}\right)$, where $\underline{f}' = F(X) \times F(Y') \stackrel{\cong}{\longrightarrow} F(X \times Y') \stackrel{F(f')}{\longrightarrow}F(X')$. It can be shown that $\mathbf{Lens}_A(F)$ is also Cartesian left-additive.
\end{proof}

\begin{proposition}
  $\mathbf{Lens}_A$ has a copointed structure\footnote{An endofunctor $F: \mathcal{D} \to \mathcal{D}$ is copointed if it is endowed with a natural transformation $\epsilon: F \Rightarrow \mathrm{id}_{\mathcal{D}}$.}.
\end{proposition}
\begin{proof}
  It suffices to endow $\mathbf{Lens}_A$ with the natural transformation $\epsilon$ whose components are the forgetful functors $\epsilon_{\mathcal{C}}: \mathbf{Lens}_A(\mathcal{C}) \to \mathcal{C}$ which strip away the backward passes.
\end{proof}

Hence, \cite{gavranovic2024fundamental} defines generalized CRDCs as follows.

\begin{definition}[Generalized Cartesian reverse differential category]
  A generalized Cartesian reverse differential category is a coalgebra for the pointed endofunctor $\mathbf{Lens}_A$. 
\end{definition}

Explicitly, a colagebra for $\mathbf{Lens}_A$ is a pair $(\mathcal{C}, \mathbf{R}_{\mathcal{C}})$ such that $\mathcal{C}: \mathbf{CLACat}$ and $\mathbf{R}_{\mathcal{C}}: \mathcal{C} \to \mathbf{Lens}_A(\mathcal{C})$ satisfies $\mathbf{R}_{\mathcal{C}} \fatsemi \epsilon_{\mathcal{C}} = \mathrm{id}_{\mathcal{C}}$. The intuition behind such definition is that $\mathbf{R}_{\mathcal{C}}$ should map $f \mapsto \left(\begin{smallmatrix} f \\ \mathrm{R}[f] \end{smallmatrix}\right)$, where $\mathrm{R}$ is a generalized reverse derivative combinator. \cite{gavranovic2024fundamental} shows that ordinary CRDCs are generalized CRDCs under this definition of $\mathbf{R}_{\mathcal{C}}$. 

\subsection{Parametric lenses}

We conclude this section discussing the relation between the $\mathbf{Para}$ construction and the $\mathbf{Lens}_A$ endofunctor. \cite{gavranovic2024fundamental} and \cite{gavranovicposition} show that, under an appropriate definition, actegorical strong functors induce $2$-functors between parametric $2$-categories.

\begin{proposition}
  \label{prop: paraend}
  Suppose $(\mathcal{C}, \bullet)$ is a strict $\mathcal{M}$-actegory and $F: \mathcal{C} \to \mathcal{C}$ is an actegorical endofunctor with strength $\sigma$. Then, $F$ induces a  $2$-endofunctor $\mathbf{Para}(F): \mathbf{Para}_{\bullet}(\mathcal{C}) \to \mathbf{Para}_{\bullet}(\mathcal{C})$.
\end{proposition}
\begin{proof}
  Define $\mathbf{Para}(F)$ so that: 
  \begin{enumerate}
    \item $\mathbf{Para}(F)$ acts like $F$ on objects $A: \mathcal{C}$;
    \item $\mathbf{Para}(F)(f) = P \bullet F(A) \stackrel{\sigma_{P,A}}{\longrightarrow} F(P \bullet A) \stackrel{F(f)}{\longrightarrow} F(B)$ for all $(P,f):\mathbf{Para}_{\bullet}(\mathcal{C})(A,B)$;
    \item $\mathbf{Para}(F)$ leaves reparametrizations unchanged.
  \end{enumerate}
\end{proof}

As a consequence, it can be shown that, if  $(\mathcal{C}, \mathbf{R}_{\mathcal{C}})$ is a generalized CRDC,  $\mathbf{R}_{\mathcal{C}}$ induces a $2$-functor $\mathbf{Para}(\mathbf{R}_{\mathcal{C}}): \mathbf{Para}_{\times}(\mathcal{C}) \to \mathbf{Para}_{\bullet}(\mathbf{Lens}_A(\mathcal{C}))$, which takes a parametric map $f: P \times A \to B$ and augments is with its reverse derivative $\mathrm{R}[f]$, forming a parametric lens. Parametric lenses behave very similarly to lenses, but we provide a separate stand-alone definition (which we take from \cite{cruttwell2022categorical}) for the reader's convenience.

\begin{figure}[h]
  \begin{center}
    \includegraphics[width=0.5\textwidth]{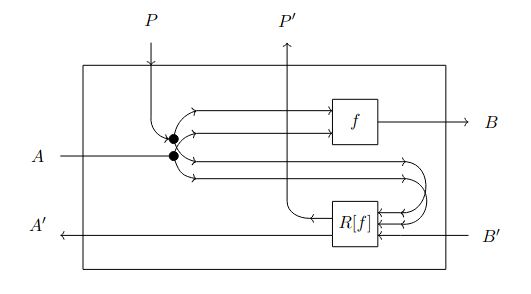}             
    \caption[Parametric lens]{String diagrams representing the inner workings of a parametric lens. (Images taken from \cite{cruttwell2022categorical}.)}
    \label{fig: paralenses}
  \end{center}
\end{figure}

\begin{definition}[Parametric lenses]
  The category of parametric lenses over a Cartesian category $(\mathcal{C},1,\times)$ is $\mathbf{Para}_{\bullet}(\mathbf{Lens}(\mathcal{C}))$, where $\bullet$ is the action on the lenses generated by the Cartesian structure of $\mathcal{C}$:
  \[\begin{pmatrix}
    P \\ P'
  \end{pmatrix} \bullet \begin{pmatrix}
    A \\ A'
  \end{pmatrix} = \begin{pmatrix}
    P \times A \\ P' \times A'
  \end{pmatrix}.\]
  Refer to \textit{Fig. \ref{fig: paralenses}} to see a string diagram that shows the inner workings of a parametric lens.
\end{definition}

\section{Supervised learning with parametric lenses}
\label{sec: opticbasedlearning}

In this section, we show how parametric lenses can be used to model supervised gradient-based learning (\cite{cruttwell2022categorical}, \cite{gavranovic2024fundamental}, \cite{shiebler2021category}). While lenses are not as general as weighted optics, it is shown in \cite{cruttwell2022categorical} that they are powerful enough for most purposes and that there is empirical evidence of their applicability and performance. The paper also discusses the use of parametric lenses in modeling unsupervised deep learning and deep dreaming, but we do not have enough space to discuss this topic.

\subsection{Model, loss, optimizer, learning rate}

Supervised gradient-based learning can be modeled using parametric lenses as follows:
\begin{enumerate}
  \item  we can design an \textit{architecture} $(P,\mathsf{Model})$ as a parametric morphism in $\mathbf{Para}_{\bullet}(\mathcal{C})$ for some generalized CRDC $(\mathcal{C},\mathbf{R}_{\mathcal{C}})$;
  \item we can use the functor $\mathbf{R}_{\mathcal{C}}$ to endow $(P,\mathsf{Model})$ with its reverse derivative $\mathrm{R}[(P,\mathsf{Model})]$, yielding a lens in $\mathbf{Para}_{\bullet}(\mathbf{Lens}_A(\mathcal{C}))$;
  \item we can use $2$-categorical machinery of $\mathbf{Para}_{\bullet}(\mathbf{Lens}_A(\mathcal{C}))$ to provide a \textit{loss function}, a \textit{learning rate}, and an \textit{optimizer}, which can be assembled onto $\mathbf{R}_{\mathcal{C}}(P,\mathsf{Model})$ to yield a supervised learning lens able to update parameters based on inputs and predictions;
  \item we can use copy maps from the Cartesian structure of $\mathcal{C}$ to create a learning iteration.
\end{enumerate}

The theory of parametric optics and differential categories does not offer explicit insight with respect to architecture design, so we will assume a good architecture has already been designed\footnote{We will come back on this in \textit{Chpt. 2}.}. Given an architecture $(P,\mathsf{Model})$, it can be embedded into $\mathbf{Para}_{\bullet}(\mathbf{Lens}_A(\mathcal{C}))$ as a lens $\left(\begin{smallmatrix} P \times A \\ P \times A \end{smallmatrix}\right) \to \left(\begin{smallmatrix} B \\ B \end{smallmatrix}\right)$ by breaking it up into its basic components (such as linear layers, convolutional layers, etc.), augmenting such components with their reverse derivatives, and the composing the resulting lenses. The backward pass of the composition is the reverse derivative of its forward pass because $\mathbf{R}_{\mathcal{C}}$ is a functor\footnote{As highlighted by \cite{shiebler2021category}, the diagram for the backward pass of the composition of two lenses looks exactly like the diagram describing the chain rule for reverse derivatives, which is what makes $\mathbf{R}_{\mathcal{C}}$ a well-defined functor.}. Many examples can be found in \cite{cruttwell2022categorical}.

\begin{figure}[h]
  \begin{center}
    \begin{subfigure}{0.3\textwidth}
      \includegraphics[width=\textwidth]{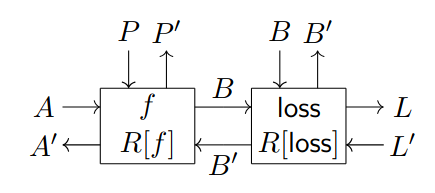}
      \caption{}
    \end{subfigure}
    \begin{subfigure}{0.4\textwidth}
      \includegraphics[width=\textwidth]{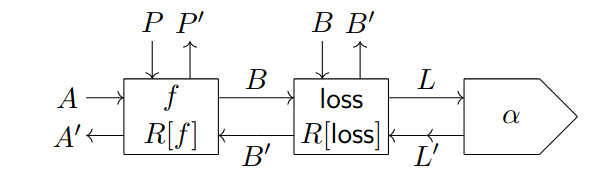}
      \caption{}
    \end{subfigure}
    \begin{subfigure}{0.5\textwidth}
      \includegraphics[width=\textwidth]{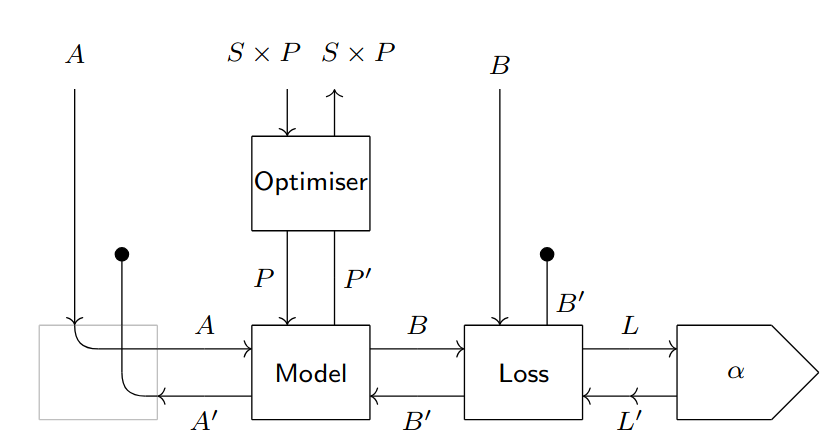}
      \caption{}
    \end{subfigure}                   
    \caption[Supervised learning with parametric lenses]{String diagrams representing (a) the composition of a model lens and a loss function lens (b) the composition of a model lens, a loss function lens, and a learning rate lens, (d) a supervised learning lens. (Images taken from \cite{cruttwell2022categorical}.)}
    \label{fig: suplearnlenses}
  \end{center}
\end{figure}

Updating the parameters based on data requires a loss function, an optimizer and a learning rate. Loss functions can be implemented as parametric lenses which take in predictions as input and labels as parameters. The output they produce can be considered the actual loss that needs to be differentiatied. Given a model parametric lens $\left(\begin{smallmatrix} \mathsf{Model} \\ \mathrm{R}[\mathsf{Model}] \end{smallmatrix}\right):\left(\begin{smallmatrix} P \times A \\ P \times A \end{smallmatrix}\right) \to \left(\begin{smallmatrix} B \\ B \end{smallmatrix}\right)$ and a loss parametric lens $\left(\begin{smallmatrix} \mathsf{Loss} \\ \mathrm{R}[\mathsf{Loss}] \end{smallmatrix}\right):\left(\begin{smallmatrix} B \times B\\ B \times B \end{smallmatrix}\right) \to \left(\begin{smallmatrix} L \\ L \end{smallmatrix}\right)$, the composition $\left(\begin{smallmatrix} \mathsf{Model} \\ \mathrm{R}[\mathsf{Model}] \end{smallmatrix}\right) \fatsemi \left(\begin{smallmatrix} \mathsf{Loss} \\ \mathrm{R}[\mathsf{Loss}] \end{smallmatrix}\right)$ takes in features as input and takes model parameters and labels as parameters. Then, this information is used to compute the loss associated with the predictions of the model. See \textit{Fig. \ref{fig: suplearnlenses} (a)} for the associated string diagram.

It can be helpful to think about dangling wires in the diagrams as open slots where other components can be plugged. For instance, the diagram of \textit{Fig. \ref{fig: suplearnlenses} (a)} has dangling wires labeled with $L$ on its right. We can use a learning rate lens $\alpha$ to link these wires and allow forward-propagating information to "change direction" and go backwards. $\alpha$ must have domain equal to $\left( \begin{smallmatrix} L \\ L \end{smallmatrix} \right)$ and codomain equal to $\left(\begin{smallmatrix} 1 \\ 1 \end{smallmatrix}\right)$, where $1$ is the terminal object of $\mathcal{C}$. For instance, if $\mathcal{C} = \mathbf{Smooth}$, $\alpha$ might just multiply the loss by some $\epsilon$, which is what machine learning practitioners would ordinarily call learning rate. \textit{Fig. \ref{fig: suplearnlenses} (b)} shows how a learning rate can be linked to the loss function and the model using post-composition.

The final element needed for the model $\mathsf{Model}$ in \textit{Fig. \ref{fig: suplearnlenses} (b)} to learn is an optimizer. It is shown in \cite{cruttwell2022categorical} that optimizers can be represented as reparametrisations in $\mathbf{Para}(\mathbf{Lens}(\mathcal{C}))$. More specifically, we might see an optimizer as a lens $\left( \begin{smallmatrix} P \\ P \end{smallmatrix} \right) \to \left( \begin{smallmatrix} Q \\ Q \end{smallmatrix} \right)$. In gradient descent, for example, $P = Q$ and the aforementioned lens is $\left( \begin{smallmatrix} 1_P \\ -_P \end{smallmatrix} \right)$. We can plug such reparametrisation on top of the model, we can redirect the input wires of the model to convert them into parameters, and we can plug useless wires with delete maps taken from the Cartesian structure of $\mathcal{C}$. We are then left with a $\left(\begin{smallmatrix} 1 \\ 1 \end{smallmatrix}\right) \to \left(\begin{smallmatrix} 1 \\ 1 \end{smallmatrix}\right)$ parametric lens with parameter space $\left(\begin{smallmatrix} A \\ 1 \end{smallmatrix}\right) \times \left(\begin{smallmatrix} P \\ P \end{smallmatrix}\right) \times \left(\begin{smallmatrix} B \\ 1 \end{smallmatrix}\right)$. This lens is pictured in \textit{Fig. \ref{fig: suplearnlenses} (c)}.The diagram shows how the machinery hidden by the $\mathbf{Para}(\mathbf{Lens}(\mathcal{C}))$ can take care of forward propagation, loss computation, backpropagation and parameter updating in a seamless fashion.

\subsection{Weight tying, batching, and the learning iteration}

\begin{figure}[h]
  \begin{center}
    \begin{subfigure}{0.3\textwidth}
      \includegraphics[width=\textwidth]{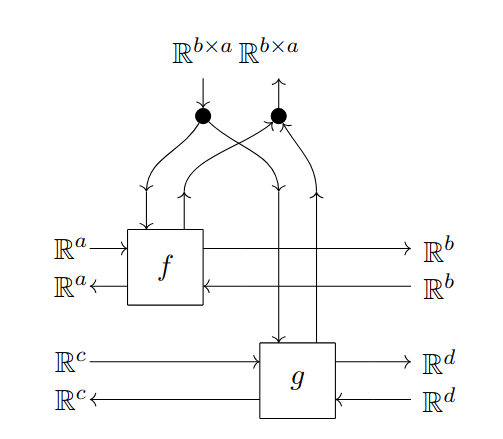}
      \caption{}
    \end{subfigure}
    \begin{subfigure}{0.3\textwidth}
      \includegraphics[width=\textwidth]{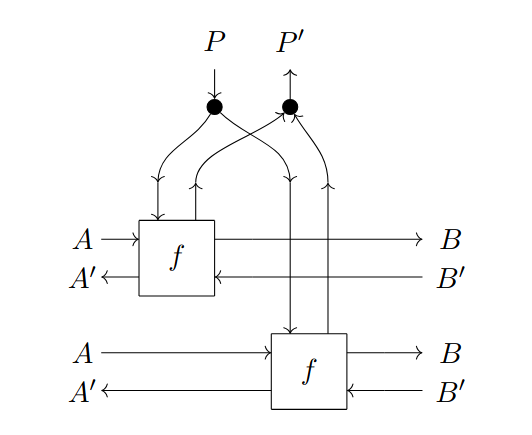}
      \caption{}
    \end{subfigure}              
    \caption[Weight tying and batching]{String diagrams representing (a) weight trying and (b) batching, both implemented using the copy maps of a Cartesian category. (Images taken from \cite{cruttwell2022categorical}.)}
    \label{fig: weighttying}
  \end{center}
\end{figure}

Both \cite{cruttwell2022categorical} and \cite{gavranovic2024fundamental} emphasize the essential role played by \textit{weight tying} in deep learning. Weight tying can be implemented within the parametric lens framework as a reparametrization that copies a single parameter to many parameter slots (see \textit{Fig. \ref{fig: weighttying}} (a)): given $(P \times P, f): \mathbf{Para}(\mathcal{C})(X,Y)$, we can define $(P, f^{\Delta_P}): \mathbf{Para}(\mathcal{C})(X,Y)$ so that
\[f^{\Delta_P}: P \times X \stackrel{\Delta_P \times X}{\longrightarrow} P \times P \times X \stackrel{f}{\longrightarrow} Y.\]
Copy maps can also be used for \textit{batching}: batching is implemented by instantiating $n$ different copies of our supervised learning lens (comprised of model, loss function, and learning rate) and tying the parameters to a unique value. Then, it suffices to feed the $n$ data points to the $n$ lenses, and we can optimize across a single parameter (see \textit{Fig. \ref{fig: weighttying}} (b)).

\cite{cruttwell2022categorical} introduces a possible representation for the whole learning iteration of a supervised learning model as a single map. The paper suggests extracting the backward pass of the lens in \textit{Fig. \ref{fig: suplearnlenses} (d)} and reframing it as a $P \to P$ parametric map with parameters $A \times B$. Since this is an endomap, it can be composed $n$ times with itself to obtain a $P \to P$ map, which is proposed as a model of the learning iteration. While this approach requires breaking lenses apart, it is markedly simple.

\subsection{Empirical evidence}

Empirical evidence for the effectiveness of the parametric lens framework discussed in this section can be found in \cite{cruttwell2022categorical}, where the authors implement a Python library for gradient-based learning rooted in these ideas. They use the library to develop a MNIST classifier, obtaining comparable accuracy to models developed using traditional tools. The Python implementation of components of learning as parametric lenses is elegant and mathematically principled, as it mirrors an abstract categorical structure. It is also insightful because it highlights possible generalizations, which manifest as simple modifications of existing lenses. 

This kind of success story foreshadows a future where popular machine learning libraries also follow elegant principled paradigms informed by category theory. Quoting  \cite{cruttwell2022categorical} directly, \dblquote{[the] proposed algebraic structures naturally guide programming practice}.

\section{Future directions and related work}

The parametric optic framework discussed in this chapter is very promising, but there is still a lot of work that needs to be done for it to reach its full potential. For instance, \cite{gavranovic2024fundamental} conjectures that weighted optics can be used in its full generality to model differentiation in cases which are not covered by lenses. For instance, lenses cannot model automatic differentiation algorithms that do not use gradient checkpointing, while weighted optics are conjectured to be able to do so. \cite{gavranovic2024fundamental} suggests investigating locally graded categories as potential replacements for actegories, and also suggests investigating the applications of parametric optics to meta-learning, that is deep learning where the optimizers themselves are learned. Moreover, \cite{cruttwell2022categorical} conjectures that some of the axioms of CRDC may be used to model higher order optimization algorithms. Finally, as suggested by \cite{cruttwell2022categorical}, future work might allow the parametric optic framework to encompass non-gradient based optimizers such as the ones used in probabilistic learning. See \cite{shiebler2021category} for more on this topic.

We conclude this chapter by discussing three other directions of machine learning research that are closely related to the framework of parametric optics.

\subsection{Learners}

One of the first compositional approaches to training neural networks in the literature can be found in the seminal paper \cite{fong2019backprop}, which spurred a lot of research in the field, including what is presented in \cite{gavranovic2024fundamental} and \cite{cruttwell2022categorical}. The authors introduce a category of \textit{learners}, objects which are meant to represent neural network components and behave similarly to parametric lenses. 

\begin{definition}[Category of learners]
  Let $A$ and $B$ be sets. A learner $A \to B$ is a tuple $(P, I, U, r)$ where $P$ is a set, and $I: P \times A \to B$, $U: P \times A \times B \to P$, and $r: P \times A \times B \to A$ are functions.
  $P$ is known as parameter space, $I$ as implement function, $U$ as update function, and $r$ as request function. Two learners $(P, I, U, r): A \to B$ and $(Q, J, V, s): B \to C$ compose forming $(P \times Q, I \ast J, U \ast V, r \ast s ): A \to C$, where
  \[(I \ast J)(p,q,a) = J(q,I(p,a)),\]
  \[(U \ast V)(p,q,a,c) = (U(p,a,s(q,I(p,a),c)), V(q,I(p,a),c)),\]
  \[(r \ast s)(p,q,a,c) = r(p,a,s(q,I(p,a),c)).\]
  Learners quotiented by an appropriate reparametrization relationship\footnote{As argued in \cite{fong2019backprop}, learners could be studied from a bicategorical point of view, where reparametrizations would just be $2$-cells. We could then use a connected component projection to compress $\mathbf{Learn}$ into a $1$-category $\mathbf{Learn}$, as it is done for $\mathbf{coPara}$ when defining weighted optics.} form a category $\mathbf{Learn}$.
\end{definition}

A learner represents an instance of supervised learning: the implement function takes a parameter and implements a function and the update function updates the parameters using a data from a dataset. The request function is necessary to implement backpropagation when optimizing a composition of learners. Suppose we select a learning rate $\epsilon$ and an error function $e: \mathbb{R}^2 \to \mathbb{R}$ such that $y \mapsto \frac{\partial e }{\partial x}(x_0, y)$ is invertible for all $x_0$. It is argued in \cite{fong2019backprop} that we can define a functor $L_{\epsilon, e}: \mathbf{Para}_{\times}(\mathbf{Smooth}) \to \mathbf{Learn}$ which takes a parametric map and yields an associated learner that implements gradient descent. 

We do not have the space to talk about learners at length, but we wish to draw a short comparison between parametric weighted optics (and, in particular, parametric lenses) and the approach of \cite{fong2019backprop}, given the relevant position held by the latter in machine learning literature. The similarities between learner-based learning and lens-based learning are evident: every learner $(P, I, U, r)$ looks like a parametric lens, where $I$ passes information forward, $r$ passes information backwards and $P$ is the parameter space. Moreover, the role of $L_{\epsilon, e}$ is very similar to the role played by $\mathbf{Para}(\mathbf{R}_{\mathcal{C}})$ in optic-based learning. Such similarities were even discussed in the original paper \cite{fong2019backprop} and have been researched at length: it has been proved in \cite{fong2019lenses} that learners can be functorially embedded in a special category of symmetric lenses (as opposed to the lenses of \textit{Def. \ref{def: lenses}}, which are asymmetric).

Despite the similarities, there is one fundamental difference between the lens-based approach and the learner-based approach: each learner carries its own optimizer, whereas optimization of lenses is usually carried out separately. Moreover, if we compare parametric weighted optics with learners, the latter clearly win in versatility, generality, and (at least from our point of view) conceptual clarity. It is argued in \cite{shiebler2021category} and \cite{cruttwell2022categorical} that the parametric lens framework largely subsumes the learner approach. More information regarding the comparison can also be found in \cite{gavranovic2024fundamental}.

\subsection{Exotic differential categories}

We have presented the parametric weighted optic approach of \cite{gavranovic2024fundamental} and \cite{cruttwell2022categorical} within the context of neural networks for the sake of simplicity, but the framework has been developed with generality in mind and applies to a much wider range of situations. For instance, we can easily replace $\mathbf{Smooth}$ with any other CRDC $\mathcal{C}$, yielding a full-feature compositional framework for gradient-based learning over $\mathcal{C}$.

Switching to a different CRDC is useful because different differential categories can lead to different learning outcomes, both in terms of accuracy of the model and in terms of computational costs (\cite{wilson2022categories}).
For instance, is argued in \cite{wilson2022categories} that polynomial circuits can be used to define and train intrinsically discrete machine learning models. Even \lq radical\rq environments such as Boolean circuits - where scalars reside in $\mathbb{Z}_2$ - seem to be conductive to machine learning under the right choice of architecture and optimizer (\cite{wilson2021reverse}). Using such exotic differential categories could be of great advantage because they might be able to better reflect the intrinsic computational limits of computer arithmetic, leading to more efficient learning (\cite{wilson2022categories}). 

\subsection{Functional reverse-mode automatic differentiation}

Finally, we wish to highlight the similarities between the formal theory of differential categories illustrated here and the work in \cite{elliott2018simple}. The paper describes the Haskell implementation of a purely functional automatic differentiation library, which is able to handle both forward mode and reverse mode automatic differentiation without resorting to the mutable computational graphs used by most current day libraries. 

Among the main insights of \cite{elliott2018simple}, it is stated that derivatives should not be treated as simple vectors, but as linear maps, or multilinear maps in the case of uncurried higher-order derivatives. Moreover, the author shows that differentiation can be made compositional by working on pairs $(f, Df)$, which behaved very similarly to lenses. As noted by \cite{shiebler2021category}, however, \cite{cruttwell2022categorical} and other lens-theoretical perspectives do not subsume the work in \cite{elliott2018simple} because of the latter's programming focus. See \cite{shiebler2021category} for more information regarding this comparison.

\clearpage{\pagestyle{empty}\cleardoublepage}


\chapter{From Classical Computer Science to Neural Networks}
\lhead[\fancyplain{}{\bfseries\thepage}]{\fancyplain{}{\bfseries\rightmark}}

Classical computer science focuses on discovering algorithms, that is ordered sequences of steps which operate in precisely set, idealized conditions and have strong guarantees of correctness due to their exact mathematical formulations. Neural networks, on the other hand, are able to work in messy, real-world conditions, but offer very so few guarantees of correctness that their performance is often described as \textit{unreasonably} good. Moreover, whereas algorithms generalize very well (most software engineers will only need a few dozen algorithms in their entire career), neural networks are often completely helpless when pitted against out of distribution inputs. Hence, algorithms and neural networks can be seen as complementary opposites (\cite{velivckovic2021neural}, \cite{velivckovic2022clrs}).

Recent attempts going under the label of \textit{neural algorithmic reasoning} (see \cite{velivckovic2021neural} for a very short introduction to the subject) have tried to get the best of both worlds by training neural networks to execute algorithms (see e.g \cite{ibarz2022generalist}). The CLRS benchmark (introduced by \cite{velivckovic2022clrs}) uses graphs to represent the computations associated with a few classical algorithms from the famous CLRS introductory textbook (\cite{cormen2022introduction}) so that graph neural networks (GNNs) can be trained to learn these algorithms. The benchmark has spurred a large amount of research in this direction, with very promising results.

More generally, linking machine learning to classical computer science might unlock interesting advances. For example, recovering neural networks as parametric versions of known algorithms might help classify existing architectures in a conceptually clear manner, and it might even help discover new neural network architectures by taking inspiration from well-researched classical notions. In this chapter, we illustrate two lines of inquiry which use category theory to build such a bridge: \textit{categorical deep learning} and an interesting categorical approach to \textit{algorithmic alignment}. Before treating such topics, we go on a short categorical tangent regarding (co)algebras and integral transforms.

\section{Categorical toolkit}

\subsection{(Co)algebras}
\label{subsec: algebras}

Categorical \textit{algebras} and \textit{coalgebras} are a formalization of the principles of induction and coinduction. Induction and coinduction are fundamental to computer science because they allow us to give precise definitions for many data structures and to formalize recursive and corecursive algorithms on such structures. We will touch on (co)algebras very briefly but we refer interested readers to \cite{jacobs1997tutorial} and \cite{wisbauer2008algebras} for further detail.

\begin{definition}[(Co)algebra over an endofunctor]
  \label{def: endalg}
  Let $F: \mathcal{C} \to \mathcal{C}$ be an endofunctor. An algebra over $F$ is a pair $(A,a)$ where $A: \mathcal{C}$ and $a:\mathcal{C}(F(A),A)$. A coalgebra is a pair $(A,a)$ where $A: \mathcal{C}$ and $a:\mathcal{C}(A,F(A))$. In both cases $A$ is known as carrier set and $a$ as structure map.
\end{definition}

(Co)algebras can also be defined on monads: the only difference between (co)algebras over an endofunctor and (co)algebras over a monad is that the latter also need to be compatible with the monad structure, that is, they have satisfy various coherence conditions (see \cite{gavranovicposition}). (Co)algebras over the same functor can be given a categorical structure by using the following notion of homomorphism.

\begin{figure}[h]
  \begin{center}
    \begin{subfigure}{0.25\textwidth}
      \includegraphics[width=\textwidth]{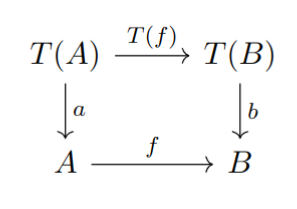}
      \caption{}
    \end{subfigure}
    \begin{subfigure}{0.25\textwidth}
      \includegraphics[width=\textwidth]{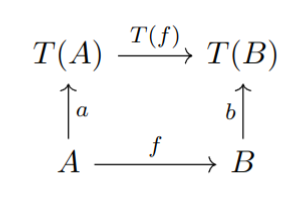}
      \caption{}
    \end{subfigure}   
    \begin{subfigure}{0.25\textwidth}
      \includegraphics[width=\textwidth]{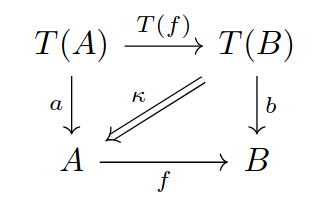}
      \caption{}
    \end{subfigure}            
    \caption[(Co)algebra homomorphisms]{Conditions under which $f:(A,a)\to(B,b)$ is (a) an algebra homomorphism, (b) a coalgebra homomorphism, (c) a lax algebra homomorphism. (Image (c) is taken from \cite{gavranovicposition}.)}
    \label{fig: alghom}
  \end{center}
\end{figure}

\begin{definition}[Homomorphisms of (co)algebras over an endofunctor]
  Let $(A,a)$ and $(B,b)$ be algebras over the same endofunctor $F: \mathcal{C} \to \mathcal{C}$. An algebra homomorphism $(A,a)\to (B,b)$ is a map $f: \mathcal{C}(A,B)$ such that the diagram in \textit{Fig. \ref{fig: alghom} (a)} is commutative.

  Now suppose $(A,a)$ and $(B,b)$ are coalgebras. A homomorphism between them is a map $f: \mathcal{C}(A,B)$ such that the diagram in \textit{Fig. \ref{fig: alghom} (b)} is commutative.
\end{definition}

The main intuition behind the notions of algebra and coalgebra is the following: the underlying functor defines a signature for the (co)algebraic structure; the structure of an algebra is a constructor that takes data from $F(A)$ and uses it to build data from $A$, whereas the structure of a coalgebra observes data from $A$ and produces an observation in the form of data from $F(A)$; (co)algebra homomorphism are arrows that preserve the underlying structure. Consider the following clarifying examples from \cite{gavranovicposition}.

\begin{remark}
  In the examples below we use polynomial and exponential expressions to define endofunctors over $\mathbf{Set}$. In this context, $X$ is the argument of the functor, $\times$ is the Cartesian product, $+$ is the disjoint union, $\langle f,g \rangle$ is the pairing induced by $\times$, $[f,g]$ is the pairing induced by $+$, and $B^A$ is the set of functions $A \to B$. The $\times$ operator is assumed to take precedence over the $+$ operator. Similarly, the exponential operator is assumed to take precedence over the $\times$ operator. 
\end{remark}

\begin{example}[Lists]
  \label{ex: lists}
  Let $A$ be a set. Consider the endofunctor $1 + A \times X: \mathbf{Set} \to \mathbf{Set}$. If $\mathsf{List}(A)$ is the set of $A$-labeled lists, $(\mathsf{List}(A), [\mathsf{Nil}, \mathsf{Cons}])$ is an algebra over $1 + A \times X$. Here, $\mathsf{Nil}: 1 \to {List}(A)$ is the map which takes the unique object of $1$ and returns the empty list, while $\mathsf{Cons}: A \times \mathsf{List}(A) \to \mathsf{List}(A)$ is the map which takes an element $a \in A$ and a list $l$ of elements of $A$ and returns the concatenated list $l \cup \{a\}$. The algebra $(\mathsf{List}(A), [\mathsf{Nil}, \mathsf{Cons}])$ describes lists in $\mathsf{List}(A)$ inductively as objects formed by concatenating elements of $A$ to other lists in $\mathsf{List}(A)$. The base case is the empty list.
\end{example}

\begin{example}[Mealy machines]
  \label{ex: mealy}
  Now consider two sets $I$ and $O$ of possible inputs and outputs, respectively. Consider the endofunctor $(O \times X)^I: \mathbf{Set} \to \mathbf{Set}$.  Define $\mathsf{Mealy}_{I,O}$ as the set of Mealy machines with inputs and outputs in $I$ and $O$, respectively. Now we can consider the coalgebra $(\mathsf{Mealy}_{I,O}, \mathsf{Next})$, where $\mathsf{Next}$ is the map that takes a Mealy machine $m \in \mathsf{Mealy}_{I,O}$ and yields a function which in turn, given $i \in I$, returns the output of $m$ at $i$ and a new machine $m'$. This is a coinductive description of Mealy machines.
\end{example}

\begin{remark}
  Notice how the description we have given of Mealy machines does not mention internal states at all. This is a recurring aspect of coinductive descriptions: as argued in \cite{jacobs1997tutorial}, coinduction is best interpreted as a process where an observer tracks the behavior of an object from the outside, with no access to its internal state. This is very useful in machine learning because the internal state of a learning model is often difficult to interpret.
\end{remark}

The link between (co)algebras and (co)induction does not stop at the definition level. The example below shows that an algebra homomorphism can model a recursive fold procedure. A similar corecursive unfold procedure can be defined by using a coalgebra homomorphism (see \cite{gavranovicposition} for further detail).

\begin{example}[List folds]
  \label{ex: listfold}
  Consider the algebra $(\mathsf{List}(A), [\mathsf{Nil}, \mathsf{Cons}])$ of lists from \textit{Ex. \ref{ex: lists}}, and consider a second algebra $(Z, [r_0,r_1])$ over the same functor. A homomorphism $f: \mathsf{List}(A) \to Z$ from the former into the latter must satisfy 
  \begin{align*}
    f(\mathsf{Nil}) &= r_0,\\
    f(\mathsf{Cons}(a,l)) &= r_1(a,f(l)). 
  \end{align*}
  Hence, $f$ is necessarily a fold over a list with recursive components $r_0$ and $r_1$. Incidentally, this proves that $f$ is unique, making $(\mathsf{List}(A), [\mathsf{Nil}, \mathsf{Cons}])$ an initial object in the category of algebras over the polynomial endofunctor $1 + A \times X$.
\end{example}

The notion of (co)algebra over a functor can be generalized to the sphere of $2$-categories, defining the notion of (co)algebra over a $2$-endofunctor. The basic concepts stay the same but the commutativity of the diagrams definining (co)algebra homomorphisms is relaxed into lax-commutativity. A square diagram of $1$-cells is lax-commutative if there exists a $2$-cell that carries the top-right composition of the diagram onto its left-bottom composition, as in \textit{Fig. \ref{fig: alghom} (c)}. Once again, we refer to \cite{gavranovicposition} for further information.

\subsection{Integral transform}

\begin{remark}
  In accordance with the notation of \cite{dudzik2022graph} and \cite{dudzik2024asynchronous}, we use $[A,B]$ to represent the set of $A \to B$ functions, where $A$ and $B$ are sets.
\end{remark}

Suppose $(R, \oplus, \otimes)$ is a commutative semiring. An \textit{integral transform} is a transformation that carries a function in $[W,R]$ to a function in $[Z,R]$ following a precise chain of steps. Integral transforms\footnote{The label integral transform refers to the fact that similar ideas can be used to write categorical definitions for familiar analytical integral transforms (\cite{golem_ph_utexas_edu}). A similar construct is also used in physics (\cite{eastwood1980cohomology}).} have been introduced by \cite{dudzik2022graph} to provide a single formalism able to describe both dynamic programming and GNNs. Integral transforms can be encoded as \textit{polynomial spans}.

\begin{figure}[h]
  \begin{center}
    \begin{subfigure}{0.21\textwidth}
      \includegraphics[width=\textwidth]{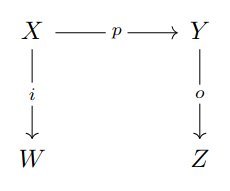}
      \caption{}
    \end{subfigure}
    \begin{subfigure}{0.28\textwidth}
      \includegraphics[width=\textwidth]{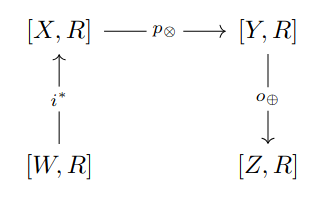}
      \caption{}
    \end{subfigure}            
    \caption[Integral transform]{(a) A polynomial span and (b) the associated integral transform. (Images taken from \cite{dudzik2022graph}.)}
    \label{fig: inttransform}
  \end{center}
\end{figure}

\begin{definition}[Polynomial span]
  A polynomial span is a triplet $(i: X \to W,p: X \to Y,o: Y \to Z)$ of morphisms in $\mathbf{FinSet}$ (that is, the category of finite sets and functions). $i$ is known as input, $p$ as process, $o$ as ouput. $W$ is known as input set, $X$ as argument set, $Y$ as message set, $Z$ as ouput set. We also ask that the fibers of $p$ have total orderings\footnote{Neither we nor \cite{dudzik2022graph} use this requirement but, as stated in the original paper, the requirement is useful to support functions with non-commuting arguments.}. The polynomial span $(i,p,o)$ can be graphically represented as the diagram in \textit{Fig. \ref{fig: inttransform} (a)}.
\end{definition}

\begin{definition}[Integral transform]
  Let $(R, \oplus, \otimes)$ be a commutative semiring. Let $(i: X \to W,p: X \to Y,o: Y \to Z)$ be a polynomial span. The associated integral transform is the triplet $(i^*: [W,R] \to [X,R], p_{\otimes}: [X,R] \to [Y,R], o_{\oplus}: [Y,R] \to [Z,R])$, where:
  \begin{enumerate}
    \item $i^*$ is the pullback mapping $f \mapsto i \fatsemi f$;
    \item $p_{\otimes}$ is the argument pushforward mapping 
    \[p_{\otimes}(a)(u) = \bigotimes_{e \in p^{-1}(u)}a(e);\]
    \item $o_{\otimes}$ is the message pushforward mapping 
    \[o_{\oplus}(m)(v) = \bigoplus_{e \in o^{-1}(v)}m(e).\]
  \end{enumerate}
  The integral transform $(i^*, p_{\otimes}, o_{\oplus})$ can be represented by the diagram in \textit{Fig. \ref{fig: inttransform} (b)}.
\end{definition}

\begin{remark}
  Whereas defining $i^*$ is quite straight-forward, defining $p_{\otimes}$ and $o_{\otimes}$ is more difficult because the arrows $p$ and $o$ point in the wrong direction, which implies that the underlying functions must be inverted before considering the associated pullbacks. However, inverting non-invertible functions yields functions into the powersets of the original domains. Moreover, if we want to preserve the multiplicity of arguments and messages, we have to construct inverses that go into the sets of multisets over the original domains. Hence why we need $\otimes$ and $\oplus$ to aggregate results over such multisets.
  The significance of these steps will be clarified later on in this chapter.
\end{remark}

\section{Categorical deep learning}

The optic-based framework we presented in the last chapter provides a structured general-purpose compositional framework for gradient-based learning, but its great versatility has a price: optics are unable to guide the architectural design of our models. It has been shown times and times again that a better architecture makes as much of a difference in machine learning as an algorithm with a better asymptotic cost does in classical computer science. 
Therefore, finding a principled mathematical framework able to guide such architectural choices is of paramount importance. In this section, we discuss a categorical approach to this problem known as \textit{categorical deep learning} (CDL). To understand the origin and motivations behind this approach, we also briefly touch upon its main precursor: \textit{geometric deep learning} (GDL).

\subsection{From GDL to CDL}

GDL (see e.g. \cite{bronstein2021geometric}) is one of the most significant approaches to the problem of architecture design. Not unlike the \textit{Erlangen Programme}, discussed in the introduction, GDL taxonomizes architectures based on the notion of symmetry. In particular, GDL considers architectures that implement \textit{equivariance} constraints with respect to group actions.

\begin{definition}[Group action equivariance and invariance]
  \label{def: equivariance}
  Let $G$ be a group and let $(S, \cdot)$ and $(T, \ast)$ be $G$-actions. A function $f: S \to T$ is equivariant with respect to the aforementioned actions if $f(g \cdot s) = g \ast f(s)$ for all $s \in \mathcal{S}$ and for all $g \in \mathcal{G}$. We say that $f$ is invariant if $\ast$ is the trivial action on $T$, and thus $f(g \cdot s) = f(s)$ for all $s$ and $g$.
\end{definition}

The GDL framework is very general and is powerful enough to derive many fundamental neural network architectures in a principled fashion.
For instance, GDL recovers convolutional neural networks from equivariance with respect to translations (actions of translation groups) and recovers graph neural networks from equivariance with respect to permutations (actions of permutation groups). However, GDL has also its limitations: first and foremost, many interesting transformations are not invertible and cannot even be approximated by group actions (\cite{gavranovicposition}). Hence, a generalization of GDL able to work outside group theory is desirable. Since category theory can be seen as a generalization of the \textit{Erlangen Programme}, it makes sense to generalize the geometric approach using category theory: \cite{gavranovicposition} achieves this by replacing the group-theoretical notion of equivariant map with the categorical notion of (co)algebra homomorphism. The authors call their approach CDL.

\begin{remark}
  At the moment, to the best of our knowledge,  \cite{gavranovicposition} is the only publicly available paper that discusses the ideas of CDL. 
\end{remark}

The main insight of CDL is that group actions can be represented as algebras over group action monads, and that maps that are equivariant with respect to these actions are homomorphisms between these algebras. Hence, GDL can be generalized by taking into consideration (co)algebras over other monads and endofunctors. According to \cite{gavranovicposition}, this yields a "theory of all architectures". The field is too young to know whether this prophecy will actually be fulfilled, but the results obtained by \cite{gavranovicposition} already look very promising.

The following proposition and the subsequent example show how exactly CDL subsumes GDL.

\begin{proposition}
  Let $(G,e,\cdot)$ be a group. The endofunctor $G \times X: \mathbf{Set} \to \mathbf{Set}$ can be given a monad structure using the natural transformations $\eta$, with components $\eta_S: s \mapsto (e,s)$, and $\mu$, with components $(g,h,s) \mapsto (g \times h, s)$. 
  The monad $(G \times -, \eta, \mu)$ can serve as a signature for $G$-actions. The actions themselves can be recovered by considering algebras $(S,\ast)$ for the monad, and, given two actions $(S,\ast)$ and $(T,\star)$, an associated equivariant map $f: S \to T$ is a $(S,\ast) \to (T,\star)$ monad algebra homomorphism.
\end{proposition}
\begin{proof}
  It suffices to compare the equations that define group actions and group action invariance with the commutative diagrams in \textit{Fig. \ref{fig: alghom}}.
\end{proof}

\begin{example}[Linear equivariant layer]
  \label{ex: translationaction}
  Consider a carrier set $S = \mathbb{R}^{\mathbb{Z}_2}$, which can be seen as a pair of pixels. Consider the translation action $(i \ast s)(j) = s(i-j)$ of $G = \mathbb{Z}_2$ on $S$, which can be seen as swapping the pixels.
  We want to find a linear map $f: S \to S$ which is equivariant with respect to the action. Imposing the equivariance constraints as equations on the entries of the matricial representation $W_f \in \mathbb{R}^{2 \times 2}$ of the map, we can prove that $f$ is equivariant if and only if $W_f$ is symmetric (\cite{gavranovicposition}).
\end{example}

\subsection{(Co)inductive definitions for RNNs}

As seen in \textit{Ex. \ref{ex: translationaction}}, the formalism of CDL subsumes the formalism of GDL, but the difference between the two is not a simple matter of notation: CDL offers a fresh new perspective and builds a novel bridge between classical computer science and machine learning. The most significant piece of novel contribution delineated in \cite{gavranovicposition} is the use of (co)algebras and (co)algebra homomorphisms over parametric categories to (co)inductively define recurrent neural networks (RNNs) and recursive neural networks (TreeRNNs). (Co)algebras are used to define cells, whereas the associated homomorphisms provide the weight-sharing mechanics used to unroll them. Let us build on \textit{Ex. \ref{ex: lists}} and \textit{Ex. \ref{ex: listfold}}, as is done in \cite{gavranovicposition}.

\begin{figure}[h]
  \begin{center}
    \begin{subfigure}{0.3\textwidth}
      \includegraphics[width=\textwidth]{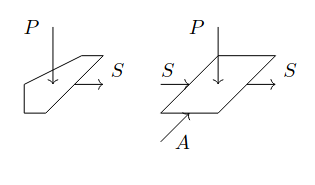}
      \caption{}
    \end{subfigure}
    \begin{subfigure}{0.48\textwidth}
      \includegraphics[width=\textwidth]{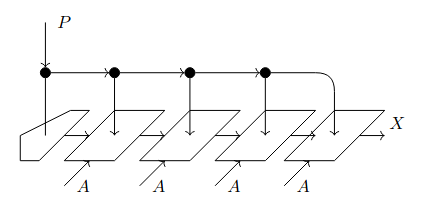}
      \caption{}
    \end{subfigure}            
    \caption[Folding RNN]{(a) Two kinds of folding RNN cell representing the initial state of the network and a generic state update based on input. (b) A folding RNN unrolled using weight tying. (Images taken from \cite{gavranovicposition}.)}
    \label{fig: foldingrnn}
  \end{center}
\end{figure}

\begin{example}[Folding recurrent neural network cell]
  \label{ex: frnncell}
  Consider the endofunctor $1 + A \times X: \mathbf{Set} \to \mathbf{Set}$ from \textit{Ex. \ref{ex: lists}}. Consider the Cartesian action of $\mathbf{Set}$ on itself and associate the following actegorical strength to the functor: $\sigma_{P,X}(p,\mathrm{inl}) = \mathrm{inl}$ and  $\sigma_{P,A}(p,\mathrm{inr}(x,x')) = \mathrm{inr}((p,x), (p,x'))$. Now that the functor is actegorical strong, we can use \textit{Prop. \ref{prop: paraend}} to construct an endofunctor $\mathbf{Para}(1 + A \times X): \mathbf{Para}_{\bullet}(\mathbf{Set}) \to \mathbf{Para}_{\bullet}(\mathbf{Set})$. Consider an algebra $(S,(P,\mathsf{Cell}))$ for this functor. Via the isomorphism $P \times (1 + A \times X) \cong P + P \times A \times X$, we deduce that $\mathsf{Cell} = [\mathsf{Cell}_0, \mathsf{Cell}_1]$, where $\mathsf{Cell}_0: P \to S$ and  $\mathsf{Cell}_1: P \times A \times S \to S$. We can interpret $\mathsf{Cell}_0$ and $\mathsf{Cell}_1$ as folding recurrent neural network cells: $\mathsf{Cell}_0$ provides the initial state based on its parameter and $\mathsf{Cell}_1$ takes in the old state, a parameter, and an input, which are then used to return a new state (\textit{Fig. \ref{fig: foldingrnn} (a)}).
\end{example}

\begin{figure}[h]
  \begin{center}
    \includegraphics[width=0.6\textwidth]{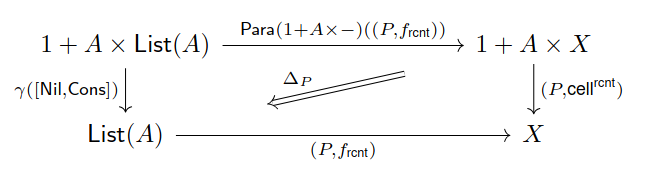}     
    \caption[Lax-commutative diagram for the unrolling of a folding RNN]{ Lax-commutative diagram that makes $(P,f): (\mathsf{List}(A), [\mathsf{Nil}, \mathsf{Cons}]) \to (S,(P,\mathsf{Cell}))$ an algebra homomorphisms. (Image taken from \cite{gavranovicposition}.)}
    \label{fig: foldingrnncommdiag}
  \end{center}
\end{figure}

\begin{example}[Unrolling of a folding recurrent neural network]
  \label{ex: frnnunroll}
  Use \textit{Prop. \ref{prop: embeddingth}} to embed the list algebra $(\mathsf{List}(A), [\mathsf{Nil}, \mathsf{Cons}])$ from \textit{Ex. \ref{ex: lists}} as an algebra over the endofunctor $\mathbf{Para}(1 + A \times X)$ define in \textit{Def. \ref{ex: frnncell}}. Now consider an algebra homomorphism $(P,f): (\mathsf{List}(A), [\mathsf{Nil}, \mathsf{Cons}]) \to (S,(P,\mathsf{Cell}))$. Since we are working with algebras over a $2$-endofunctor, we also need to specify a $2$-cell that makes the homomorphism diagram (\textit{Fig. \ref{fig: alghom} (c)}) lax-commutative. Using the weight-tying reparametrization $\Delta_P$ yields the lax commutative diagram in \textit{Fig. \ref{fig: foldingrnncommdiag}}, which uniquely identifies $f$ as the fold function which takes a list of inputs in $A$ and unrolls a folding recurrent neural network that reads such inputs. The weight-tying reparametrization makes sure that each cell of the unrolled network uses the same parameters (see \textit{Fig. \ref{fig: foldingrnn} (b)} for a graphical representation). 
\end{example}

\begin{figure}[h]
  \begin{center}
    \includegraphics[width=\textwidth]{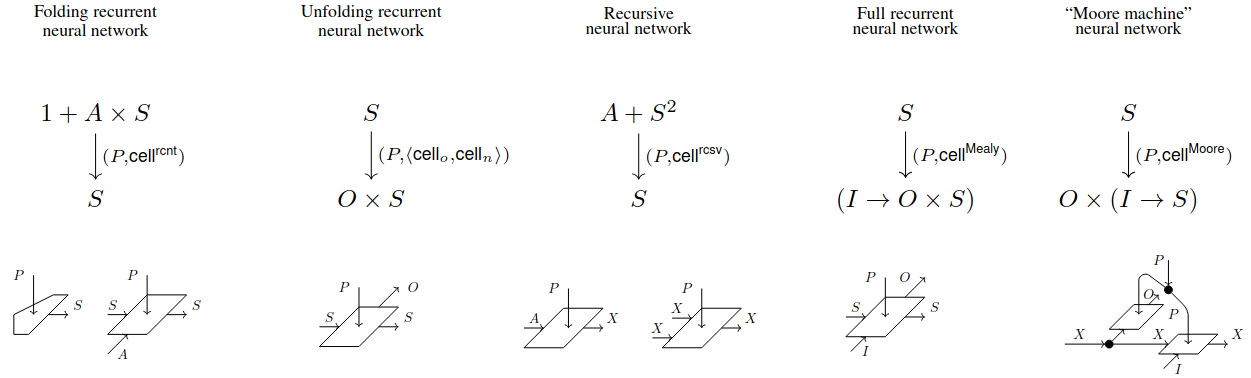}     
    \caption[Neural network cells]{Various kinds of neural network cells and the endofunctors used in their (co)algebraic definitions. (Image taken from \cite{gavranovicposition}.)}
    \label{fig: cells}
  \end{center}
\end{figure}

The construction in  \textit{Ex. \ref{ex: frnncell}} and \textit{Ex. \ref{ex: frnnunroll}} constitutes a precise mathematical link between the classical data structure of lists and the machine learning construct of folding RNNs. Similarly, \cite{gavranovicposition} recovers recursive neural networks (TreeRNNs) by building upon classical binary trees and, even more interestingly, complete RNNs are recovered from the coalgebra of \textit{Ex. \ref{ex: mealy}}, which reveals an interesting link between RNNs and Mealy machines. This begs the question: if Mealy machines generalize to recurrent neural networks, what do Moore machines generalize to? It is argued in the paper that they generalize to a variant of RNN where different cells (which share the same weights) are used for state update and output production. Hopefully, more work in this direction will lead to new neural network architectures inspired from other classical concepts. \textit{Fig. \ref{fig: cells}} shows various kinds of neural network cells and the endofunctors used in their (co)algebraic definitions.

\begin{remark}[CDL and optic-based learning]
  In all the examples discussed above, the (co)algebra homomorphisms in question return parametric maps $(P,\mathsf{model})$, which we can interpret as untrained neural networks. We can feed these maps into the $\mathbf{R}_{\mathcal{C}}$ functor associated with a generalized Cartesian reverse differential category\footnote{The examples illustrated in this section have been developed in $\mathbf{Set}$, but we see no reason why they couldn't be specialized to an appropriate CRDC.} to augment them with their reverse derivative. The framework of parametric lenses described in 
  \textit{Sec. \ref{sec: opticbasedlearning}} can then be used to train these networks. CDL and optic-based learning are thus compatible and even complementary.
\end{remark}

\section{Algorithmic alignment: GNNs and dynamic programming}

One of them main tenets of neural algorithmic reasoning is \textit{algorithmic alignment} (\cite{xu2019can}), that is, the presence of structural similarities between the subroutines of a particular algorithm and the architecture of the neural network selected to learn such algorithm.
Since \cite{xu2019can} has shown that dynamic programming algorithms align very well with message passing GNNs, and since dynamic programming encompasses a wide variety of techniques used in various domains, these GNNs  are at the forefront of neural algorithmic reasoning research (\cite{dudzik2022graph}). However, the exact link between GNNs and dynamic programming has yet to be fully formalized. In this section we present the work of \cite{dudzik2022graph}, which attempts to derive such a formalization, and the work in \cite{dudzik2024asynchronous}, which studies conditions under which message passing GNNs are invariant with respect to various form of asynchrony\footnote{While much of the work described in \cite{dudzik2024asynchronous} does not fall under the umbrella of applied category theory, we still mention it because on its close link with the work of \cite{dudzik2022graph} and with the idea of algorithmic alignment. Hopefully, future work will explore the intersection between this work and category theory.}, which is argued to improve algorithmic alignment in some cases.

\subsection{Integral transforms for GNNs and dynamic programming}

The main link between dynamic programming and GNNs is that dynamic programming itself can be interpreted from a graph-theoretical point of view. Dynamic programming breaks up problems into subproblems recursively until trivial base cases are reached. We can thus consider the graph with nodes corresponding to subproblems and edges $(y,x)$ corresponding to the relationships \lq $y$ is a subproblem of $x$\rq. Then, the solutions of the subproblems are recursively recombined to solve the original problem. This dynamic is very similar to message passing: the simpler cases are solved first, and their solutions are passed as messages along the edges so that they can be used to solve more complex cases. More precisely, we can implement a dynamic programming algorithm as a GNN on this subproblem graph, where the feature vector $h_x^{k}$ associated with a node $x$ at the $k$-th message passing iteration represents the state of the solution of the subproblem $x$ at the $k$-th iteration of the algorithm (\cite{xu2019can}). Despite this striking resemblance, rigorously formulating the link between the architecture of GNN and the structure an associated dynamic programming algorithm is not easy, the main obstacle being the difference in data type handled by the two mathematical processes: dynamic programming usually deals with tropical objects such as the semiring $(\mathbb{N} \cup \{\infty\}, \min, +)$, while GNNs usually deal with linear algebra over $\mathbb{R}$ (\cite{dudzik2022graph}).

\cite{dudzik2022graph} proposes the formalism of integral transforms as the common structure behind both message passing GNNs and dynamic programming. While a full formal proof is not given, the idea is illustrated by showing that both the Bellman-Ford algorithm and a message passing GNN can be expressed with the help of integral transforms. The difference in data type is overcome by using the weakest common hypothesis: that the data and associated operations form a semiring.

\subsubsection{Bellman-Ford algorithm}

The Bellman-Ford (BF) algorithm is one of the most popular dynamic programming algorithms and is used to find the shortest paths between a single starting node and every other node in a weighted graph $G = (V, E)$. Since we can see every node of the graph as a subproblem, and since we can see the associated edges as subproblem relationships, the BF algorithm is a very good candidate for a GNN implementation. The algorithm operates within the tropical min-plus semiring $(R = \mathbb{N} \cup \{\infty\}, \oplus = \min, \otimes = +)$, and the data can be provided as a tuple $(d, b, w)$ of three functions into $R$. Here, $d: V \to R$ stores the current best distances of the nodes, $b: V \to R$ stores the weights of the nodes, and $w: E \to R$ stores the weights of the edges. $d$ is initialized as the function that maps the initial node to $0$ and every other node to $\infty$. The values of $d$ are updated at each step of the algorithm according to the following formula, where $\mathcal{N}_u = \{v \text{ s. t. } (v,u) \in E\}$ represent the one-hop neighborhood of a node $u$:
\[d_u \longleftarrow  \min\left(d_u, \min_{v \in \mathcal{N}_u}(d_v) + w_{v,u}\right).\]

\begin{figure}[h]
  \begin{center}
    \begin{subfigure}{0.45\textwidth}
      \includegraphics[width=\textwidth]{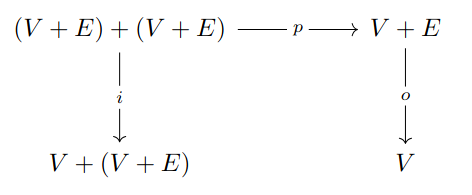}
      \caption{}
    \end{subfigure}
    \begin{subfigure}{0.4\textwidth}
      \includegraphics[width=\textwidth]{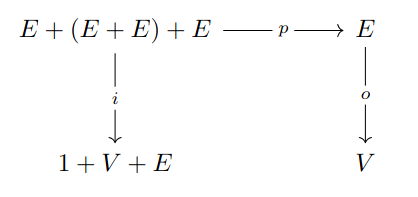}
      \caption{}
    \end{subfigure}            
    \caption[Bellman-Ford algorithm and message passing GNN]{Polynomial spans used to by \cite{dudzik2022graph} to model (a) the Bellman-Ford algorithm and (b) a message passing GNN. (Images taken from \cite{dudzik2022graph}.)}
    \label{fig: polyspans}
  \end{center}
\end{figure}

\cite{dudzik2022graph} propose the integral transform encoded by the polynomial span in \textit{Fig. \ref{fig: polyspans} (a)} as the supporting structure of the BF algorithm. The functions $i$, $p$, and $o$ are defined as follows:
\begin{enumerate}
  \item $i: (V + E) + (V + E) \to V + (V + E)$ acts as the identity on the first $V$, it maps the edges of the first $E$ to their sources, and it acts as the identity on the second $V + E$ pair;
  \item $p: (V + E) + (V + E) \to V + E$ just collapses the two copies of $V + E$;
  \item $o$ acts as the target function on the $E$ and as the identity on $V$.
\end{enumerate}
It is argued in the paper that the whole integral transform acts as step of the algorithm, carrying the data in $(d, b, w)$ to the updated function $d$. Let's examine each step: the input pullback extracts the distances of the sources of every edge; the argument pushforward computes the lengths of the one-hop extensions of the known shortest paths (the weight of each node is treated as the weight of a self-edge in this case); finally, the message pushforward selects the shortest paths to each node among the ones studied by the argument pushforward. 
Hence, the simple polynomial span in \textit{Fig. \ref{fig: polyspans} (a)} successfully encode the whole BP algorithm without any information loss or ad hoc choice. 

\subsubsection{Message passing neural network}

Consider the message passing GNN architecture described by the following equations (\cite{pmlr-v70-gilmer17a}):
\begin{align*}
  m_v^{t+1} &= \sum_{w \in \mathcal{N}_v}M_t(h_v^t,h_w^t,e_{v,w}),\\
  h_v^{t+1} &= U_t(h_v^t,m_v^{t+1}),\\
\end{align*}
where $t$ represents the time step, and $M_t$ and $U_t$ are learned differentiable functions.\cite{dudzik2022graph} argues that this GNN layer can be implemented as the integral transform associated with the polynomial span of \textit{Fig. \ref{fig: polyspans} (b)}, with an extra MLP. Here,
\begin{enumerate}
  \item $i$ sends the first $E$ to $1$,  acts as the source function on the second $E$, acts  and target function on the third $E$, and acts as the identity on the fourth $E$;
  \item $p$ collapses four $E$'s into one;
  \item $o$ acts as the target function.
\end{enumerate}
In the associated integral transform, $i^*$ gathers graph features, node features, and edge features; $p_{\otimes}$ projects such features on the edges, the MLP combines them; finally, $o_{\oplus}$ sends them to the right target.

Although not a perfect representation of the message passing architecture (due to the extra MLP), the polynomial span in \textit{Fig. \ref{fig: polyspans} (b)} can be used to inform the design of new architectures which are obtained by simple manipulations of the arrows or objects in the diagram. For instance, \cite{dudzik2022graph} uses the integral transform formalism to investigate possible performance improvements on CLRS benchmark tasks (\cite{velivckovic2022clrs}). The authors consider messages that reduce over intermediate nodes, and they show that these architectures lead to better average performance on these tasks, which is likely a result of better algorithmic alignment.

\subsection{Asynchronous algorithmic alignment}

The customary assumption behind the message passing GNN architecture requires that all messages are generated, sent, and received at the same time. We call this kind of GNN synchronous. \cite{dudzik2024asynchronous} derives conditions under which synchronous GNNs are invariant under a hypothetical asynchronous execution. This is relevant because, as stated in the paper, in many dynamic programming tasks modeled by graphs, only small parts of the aforementioned graphs are changed at each step. A synchronously executed GNN that is trained on these tasks must learn the identity function many times over, which leads to brittleness and wasted computational resources. On the other hand, an asynchronously executed GNN would be more aligned with these algorithms and thus achieve a better performance. The work in \cite{dudzik2024asynchronous} aims to reproduce these performance improvements on synchronous GNNs by imposing asynchrony invariance constraints. 

\begin{figure}[h]
  \begin{center}
    \begin{subfigure}{0.4\textwidth}
      \includegraphics[width=\textwidth]{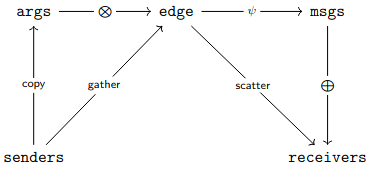}
      \caption{}
    \end{subfigure}
    \hfill
    \begin{subfigure}{0.45\textwidth}
      \includegraphics[width=\textwidth]{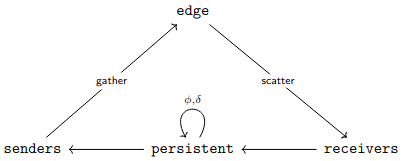}
      \caption{}
    \end{subfigure}            
    \caption[Persistent and transient features in message passing]{Diagrams describing the message passing GNN architecture discussed in \cite{dudzik2024asynchronous}. (Images taken from \cite{dudzik2024asynchronous}.)}
    \label{fig: messagepassing}
  \end{center}
\end{figure}

The authors of \cite{dudzik2024asynchronous} revise the model explored in \cite{dudzik2022graph} so that it includes a message function $\psi$ that generates messages based on gathered arguments (see \textit{Fig. \ref{fig: messagepassing} (a)} for the update diagram). Moreover, the authors argue that it is best to consider GNNs where every graph component that has a persistent state is elevated to the status of node, whereas transient computations are carried out along edges. The resulting GNN can be described by the diagram in \textit{Fig. \ref{fig: messagepassing} (b)}, where $\phi$ is the transit function that updates the persistent state of each node, and $\delta$ is the function that computes the arguments needed to generate the next messages.

It is argued in \cite{dudzik2024asynchronous} that invariance under asynchrony can be modeled by giving both arguments and messages monoidal structures. For instance, let $(M, 1, \cdot)$ be the message monoid and let $(A, 0, +)$ be the argument monoid. Then, if $S$ is the set of persistent states, state update and argument generation can be modeled as a function $M \times S \to S \times A$ which maps $(m,s) \mapsto (m \bullet s, \delta_m(s))$. Invariance under asynchronous message aggregation is obtained by defining $\bullet$ as a monoidal action of $M$ on $S$. However, \cite{dudzik2024asynchronous} shows that this is meaningful if and only if the argument generation function $\delta$ is compatible with the unitality and associativity equations of the action. This can only happen if $\delta: m \mapsto \delta_m$ is a $1$-cocycle.

\begin{definition}[$1$-cocycle]
  A map $\delta: M \to [S,A]$ is a $1$-cocycle if and only if the following are satisfied:
  \begin{enumerate}
    \item $\delta_1(s) = 0$ for all $s \in S$;
    \item $\delta_{n \cdot m}(s) = \delta_{n}(m \cdot s) + \delta_m(s)$ for all $s \in S$.
  \end{enumerate} 
  \label{def: cocycle}
\end{definition}

\begin{proposition}
  The state update function $\delta$ described above is asynchronous with respect to message passing if and only if it is a $1$-cocycle.
\end{proposition}

\cite{dudzik2024asynchronous} also proves the following.

\begin{proposition}
  Under the hypotheses described above, a single-input message function $\psi$ supports asynchronous invocation if and only if $\psi$ is a homomorphism of monoids.
\end{proposition}

We will not describe the whole formalism of \cite{dudzik2024asynchronous}, but we will show (without proof) its implications on GNN architecture design. 

\begin{example}
  Consider the message passing GNN architecture:
  \[x_u' = \phi\left(x_u, \bigoplus_{v \in \mathcal{N}_u}\psi(x_{u},x_{v})\right),\]
  where $\oplus$ is a message aggregator. The authors of \cite{dudzik2024asynchronous} derive conditions under which this architecture is invariant under asynchronies in message aggregation, node update, and argument generation: the GNN is trivially invariant under asynchronous message aggregation if messages $(M,0,\oplus)$ are given a commutative monoidal structure; invariance under asynchronies in node updates is obtained by selecting an update function $\phi$ which satisfies the associative law $\phi(s,m \oplus n) = \phi(\phi(s,m),n)$ for all $m,n \in M$ and for all $s \in S$; finally, invariance under argument generation is obtained if $\delta$ satisfies the $1$-cocycle equations (\textit{Def. \ref{def: cocycle}}). 
  These conditions are all satisfied if $\oplus$ is commutative, $M=A=S$ and $\delta_m(s) = \phi(m,s) = m \cdot s = m \oplus s$.
\end{example}

\section{Future directions and related work}

In this section we provide a brief introduction to the theory of differentiable causal computations and the theory of sheaf neural networks. These two lines of work are adjacent to the main theme of this chapter - relating classical computer science to modern machine learning - and they highlight possible directions for future research into categorical deep learning and the application of integral transforms to neural networks.

\subsection{Differentiable causal computations}

A trained RNN can be seen as a casual function according to the following definition (\cite{sprunger2019differentiable}).

\begin{definition}[Causal function]
  Let $A$ and $B$ be sets. A function $f: A^{\mathbb{N}} \to B^{\mathbb{N}}$ is causal if and only if, for all sequences $\mathbf{x}, \mathbf{y} \in A^{\mathbb{N}}$ and for all $n \in \mathbb{N}$, if $x_m = y_m$ for all $m = 0,1,\dots,n$, then $f(x_m) = f(y_m)$ for all $m = 0,1,\dots,n$.
\end{definition}

\cite{sprunger2019differentiable} studies the \textit{differential properties of causal computations}, offering valuable insight into the formal properties of RNNs. The paper focuses on sequences $\{f_k\}_{k=0,1,\dots}$ of functions $f_k: S_k \times X_k \to S_{k+1} \times Y_k$ which represent computations executed in discrete time $k$, where, at each tick $k$ of the clock, $f_k$ takes an input $x_k$ and the current state $s_k$, and uses this data to compute an output $y_k$ and a new state $s_{k+1}$. In symbols, $f_k(s_k,x_k) = (s_{k+1},y_k)$. Such sequences are given a nice compositional structure using the formalism of double categories.

\begin{figure}[h]
  \begin{center}
    \begin{subfigure}[c]{0.3\textwidth}
      \includegraphics[width=\textwidth]{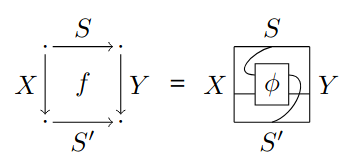}
      \caption{}
    \end{subfigure}
    \begin{subfigure}[c]{0.45\textwidth}
      \includegraphics[width=\textwidth]{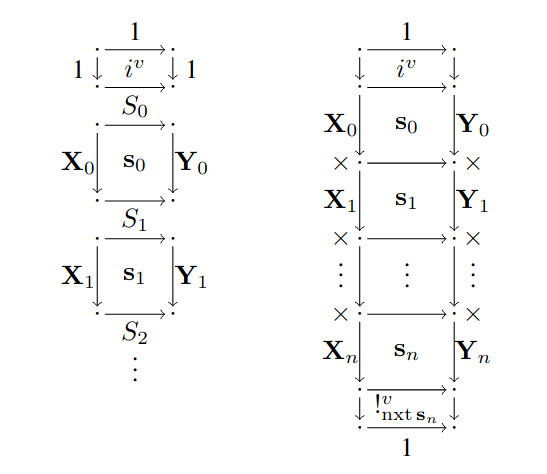}
      \caption{}
    \end{subfigure}            
    \caption[Tiles and stateful morphisms]{(a) String diagrams for a tile in $\mathbf{Dbl}(\mathcal{C})$. (b) String diagrams for a stateful morphism in $\mathbf{St}(\mathcal{C})$ and one of its truncation. (Images taken from \cite{sprunger2019differentiable}.)}
    \label{fig: tiles}
  \end{center}
\end{figure}

\begin{definition}[Category of tiles]
  Let $\mathcal{C}$ be a Cartesian category. Define $\mathbf{Dbl}(\mathcal{C})$ as the double category with the following data:
  \begin{enumerate}
    \item there is only one $0$-cell, which we represent with the $\cdot$ symbol;
    \item the horizontal and vertical $1$-cells are the objects of $\mathcal{C}$;
    \item a $2$-cell (tile) with horizontal source $S$, horizontal target $S'$, vertical source $X$, and vertical target $Y$ is a morphism $f: S \times X \to S' \times Y$ which we represent with the symbol $f:X \overset{S}{\underset{S'}\longrightarrow} Y$.
  \end{enumerate}
  It is handy to also represent $2$-cells $f$ as the tile string diagrams in \textit{Fig. \ref{fig: tiles} (a)}. The horizontal and vertical composition laws for $2$-cells are consistent with the tile diagrams. Refer to \cite{sprunger2019differentiable} for more information.
\end{definition}

\begin{definition}[Category of stateful morphism sequences]
  Let $\mathcal{C}$ be a Cartesian category. Define $\mathbf{St}(\mathcal{C})$ as the category with the following data:
  \begin{enumerate}
    \item the objects of $\mathbf{St}(\mathcal{C})$ are sequences $\mathbf{X} = \{X_k\}_{k = 0,1,\dots}$ of objects of $\mathcal{C}$;
    \item the morphisms $\mathbf{X} \to \mathbf{Y}$ are pairs $(\mathbf{f}, i)$, where $\mathbf{f} = \{f_k\}_{k = 0,1,\dots}$ is a sequence of tiles in $\mathbf{Dbl}(\mathcal{C})$ such that $f_k: X_k \overset{S_k}{\underset{S_{k+1}}\longrightarrow} Y_k$, for some sequence $\mathbf{S}$ of states, and $i: 1 \to S_0$ selects an initial state. 
  \end{enumerate}
  The morphisms of $\mathbf{St}(\mathcal{C})$ are known as stateful morphisms sequences and are represented using string diagrams as in \textit{Fig. \ref{fig: tiles} (b)}.
\end{definition}

Stateful morphisms sequences can be easily truncated and unrolled as one would expect, and it is proved in \cite{sprunger2019differentiable} that there is a bijection between stateful morphism sequences in $\mathbf{St}(\mathbf{Set})([A],[B])$ and causal functions $A^{\mathbb{N}} \to B^{\mathbb{N}}$ (here $[A]$ is the constant sequence of objects $\{A,A,A,\dots\}$). More generally, given any $\mathbf{St}(\mathcal{C})$, we can restrict our attention to constant sequences $[A]$ and stateful sequences of morphisms in the form $([f],i)$, where $f:X \overset{S}{\underset{S}\longrightarrow} Y$ is a tile in $\mathbf{Dbl}(\mathcal{C})$. This yields a subcategory $\mathbf{St}_0(\mathcal{C})$ whose morphisms can be thought of as Mealy machines that take in an input and produce an output based on an internal state which is updated after every computation. The new state is fed back to the machine after the computation, so that a new computation can take place. This is represented by the diagram in \textit{Fig. \ref{fig: tiles} (b)}.

\begin{figure}[h]
  \begin{center}
    \begin{subfigure}{0.3\textwidth}
      \includegraphics[width=\textwidth]{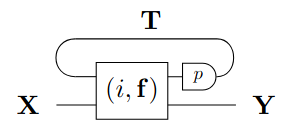}
      \caption{}
    \end{subfigure}
    \begin{subfigure}{0.45\textwidth}
      \includegraphics[width=\textwidth]{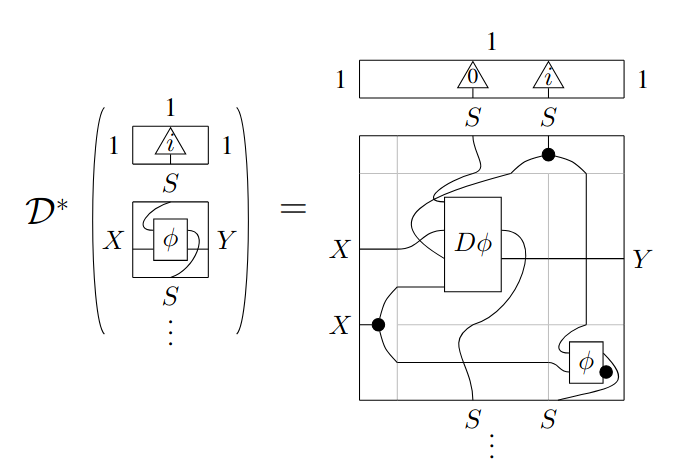}
      \caption{}
    \end{subfigure}
    \\[1cm]
    \begin{subfigure}{0.5\textwidth}
      \includegraphics[width=\textwidth]{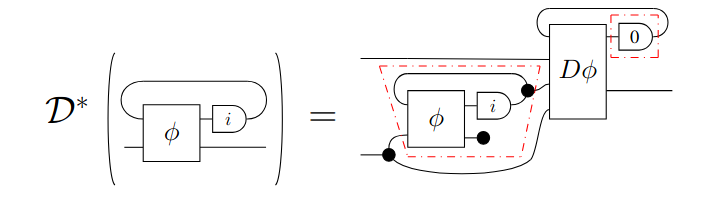}
      \caption{}
    \end{subfigure}           
    \caption[Delayed trace and derivatives in $\mathbf{St}_0(\mathcal{C})$]{String diagrams (a) representing a delayed trace in $\mathbf{St}(\mathcal{C})$ as a delayed feedback loop, (b) representing the derivative of a morphism in $\mathbf{St}_0(\mathcal{C})$, (c) representing the derivative of a delayed trace in $\mathbf{St}_0(\mathcal{C})$. (Images taken from \cite{sprunger2019differentiable}.)}
    \label{fig: differentiationanddelayedtrace}
  \end{center}
\end{figure}

The authors of \cite{sprunger2019differentiable} go on to define a delayed trace operator, which provides a rigorous formalization for feedback loops such as the one in \textit{Fig. \ref{fig: differentiationanddelayedtrace} (a)}. As stated in the paper, the delayed trace operator is closely related to the more popular trace operator (\cite{joyal1996traced}) and shares many of the same properties. 
Finally, the authors of \cite{sprunger2019differentiable} show how both $\mathbf{St}(\mathcal{C})$ and $\mathbf{St}_0(\mathcal{C})$ can be given the structure of a CDC (\textit{Def. \ref{def: cdc}}), as long as $\mathcal{C}$ is itself a CDC. This differential structure is conceptually clear, rigorously defined, and compatible with the dealyed trace operator. We do not have space to describe the details of these definitions, but we report the relevant string diagrams in \textit{Fig. \ref{fig: differentiationanddelayedtrace} (b),(c)}.

The work in \cite{sprunger2019differentiable} provides a theoretical foundation for the technique of backpropagation through time (BPTT), which consists in computing the gradient of the $k$-th unrolling of an RNN in place of the gradient of the RNN at discrete time $k$. Despite the alleged ad hoc nature of BPTT, \cite{sprunger2019differentiable} proves that the technique does not just \dblquote{involve differentiation} but is an actual \dblquote{form of differentiation} that can be reasoned about in the formalism of CDCs. Nevertheless, as stated in the paper, the differential operator of $\mathbf{St}(\mathcal{C})$ does not compute explicit gradients, and deriving the latter from the former would be computationally infeasible when there are millions of parameters. 

It is interesting to compare the approach of \cite{sprunger2019differentiable} with the framework of categorical deep learning: both CDL and the work in \cite{sprunger2019differentiable} synthetically describe RNN architectures, but, while CDL focuses on weight sharing mechanics and the (co)inductive nature of the definition, \cite{sprunger2019differentiable} focuses on the differential properties of these architectures. However, neither categorical framework deals with the problems that come up when computing gradients of unrolled RNNs, such as the presence of vanishing or exploding gradients (see e.g. \cite{hanin2018neural}). 

\subsection{Sheaf neural networks}

The theory of \textit{sheaf neural networks} (\cite{hansen2020sheaf}, \cite{bodnar2022neural}, \cite{zaghen2024nonlinear}), or SNNs, is informed by both topology and category theory, and aims to improve the GNN architecture by endowing graphs with cellular sheaf structures. In particular, SNNs are designed to solve two main issues that are encountered when training GNNs: oversmoothing, which is the tendency of deep GNNs to spread information too far in the graph to be able to effectively classify nodes, and the poor performance characteristic of GNNs when applied on heterophilic input graphs, i.e., input graphs where the nodes features are diverse in structure and attributes.

\begin{definition}[Cellular sheaf]
  A cellular sheaf $\mathcal{F}$ associated with a graph $G = (E,V)$ consists of the following data:
  \begin{enumerate}
    \item a vector space $\mathcal{F}(v)$ for every node $v \in V$;
    \item a vector space $\mathcal{F}(e)$ for every edge $e \in E$;
    \item a linear map $\mathcal{F}_{v \trianglelefteq e}: \mathcal{F}(v) \to \mathcal{F}(e)$ for each incident node-edge pair $v \trianglelefteq e$.
  \end{enumerate}
  The vector spaces associated to nodes and edges are known as stalks. The linear maps associated to incident node-edge pairs are known as restriction maps. The direct sum $C^0(G, \mathcal{F})$ of all node stalks is known as space of $0$-cochains, and the direct sum $C^1(G, \mathcal{F})$ of all edge stalks is known as space of $1$-cochains.
\end{definition}

As stated in \cite{zaghen2024nonlinear}, the  node stalks assigned by $\mathcal{F}$ serve as spaces for node features, while the restriction maps allow the data that resides on adjacent nodes to interact on edge stalks. Given a cellular sheaf $\mathcal{F}$, we can define a coboundary map $\delta$ which measures the amount \lq disagreement\rq\footnote{There is a close link between SNNs and the theory of opinion dynamics. See \cite{zaghen2024nonlinear} for further information.} between nodes. The coboundary map can then be used to define a sheaf Laplacian which can be used to propagate information in the graph (\cite{hansen2020sheaf}).

\begin{definition}[Coboundary map]
  Let $\mathcal{F}$ be a cellular sheaf on a directed graph $G = (E,V)$. The coboundary map associated with $\mathcal{F}$ is the linear map $\delta: C^0(G, \mathcal{F}) \to C^1(G, \mathcal{F})$ that maps $\delta(\mathbf{x})_{e} = \mathcal{F}_{v \trianglelefteq e}(x_v) - \mathcal{F}_{u \trianglelefteq e}(x_u)$ for each edge $e: u \to v$.
\end{definition}

\begin{definition}[Sheaf Laplacian]
  Let $\mathcal{F}$ be a cellular sheaf on a directed graph $G = (E,V)$ and let $\delta$ be the associated coboundary map. The sheaf Laplaciant associated with $\mathcal{F}$ is the linear map $L_{\mathcal{F}} = \delta^T \circ \delta: C^0(G, \mathcal{F}) \to C^0(G, \mathcal{F})$. The normalized sheaf Laplacian associated with the sheaf is the linear map $\Delta_{\mathcal{F}} = D^{-\frac{1}{2}} \circ L_{\mathcal{F}} \circ D^{\frac{1}{2}}$, where $D$ is the diagonal of $L_{\mathcal{F}}$.
\end{definition}

\begin{remark}
  The coboundary map and the sheaf Laplacian associated with a cellular sheaf $\mathcal{F}$ are generalizations of the more commonly known incidence matrix and Laplacian associated to a graph (see e.g. \cite{ward2022practical}).
\end{remark}

There are many kinds of SNN architectures (\cite{hansen2020sheaf}, \cite{bodnar2022neural}, \cite{zaghen2024nonlinear}). Due to space constraints, we only give a short description of the first one to appear in the literature: the Hansen-Gebhart SNN proposed by \cite{hansen2020sheaf}, as described by \cite{zaghen2024nonlinear}.

\begin{definition}[Sheaf neural network]
  \label{def: sheafnn}
  Suppose $G = (E,V)$ is a directed graph and $\mathcal{F}$ is a cellular sheaf on it. Suppose the stalks of $\mathcal{F}$ are all equal to $\mathbb{R}^{d \times f}$, where $d$ is the dimension of each feature vector and $f$ is the number of channels. Then, $C^0(G, \mathcal{F})$ is isomorphic to $\mathbb{R}^{nd \times f}$, where $n$ is the number of nodes, and its elements can be represented as matrices $\mathbf{X}$. The sheaf neural network proposed by \cite{hansen2020sheaf} uses the following transition function to update this features:
  \[\mathbf{Y} = \sigma((I_{nd} - \Delta_{\mathcal{F}})(I_{n} \otimes W_1)\mathbf{X}W_2),\]
  where $\sigma$ is a non-linearity, $I$ refers to identity matrices, $\Delta_{\mathcal{F}}$ is the normalized sheaf Laplacian, $\otimes$ is the Kronecker product, and, finally, $W_1 \in \mathbb{R}^{d \times d}$ and $W_2 \in \mathbb{R}^{f \times f}$ are weight matrices.
\end{definition}

\begin{remark}
  The values of $f$, $n$, $d$, and the restriction maps are all hyperparameters.
  Choosing $W_2 \in \mathbb{R}^{f_1 \times f_2}$ allows the SNN layer described above to change the number of features from $f_1$ to $f_2$.  
\end{remark}

As observed by \cite{bodnar2022neural}, the SNN architecture proposed by \cite{hansen2020sheaf} can be seen as a discretization of the differential equation 
\[\dot{\mathbf{X}}(t) = -\Delta_{\mathcal{F}}\mathbf{X}(t),\]
which is known as sheaf diffusion equation and is analogous to the heat diffusion equation used in graph convolutional networks (\cite{bodnar2022neural}). Studying the time limit of the sheaf diffusion equation yields important results about the diffusion of information through the graph after repeated application of the transformation in \textit{Def. \ref{def: sheafnn}}. In particular, \cite{bodnar2022neural} argues that, in the time limit, node feature tend to values that \lq agree\rq on the edges. Hence, \dblquote{sheaf diffusion can be seen a synchronization process over the graph}. \cite{bodnar2022neural} goes on to study the discriminative power of different classes of cellular sheaves and proposes strategies to learn the restriction maps themselves. \cite{zaghen2024nonlinear} extends the work of \cite{bodnar2022neural} by analyzing non-linear sheaf Laplacians and the associated sheaf diffusion process.

It is important to notice that SNNs can be considered a strict generalization of GNNs  since the latter are nothing but instances of the former where the sheaf structure is trivial (\cite{bodnar2022neural}). Thus, although we are not aware of any work applying SNNs to neural algorithmic reasoning, given the success enjoyed by GNNs in this area of research, we hypothesize that SNNs are be even more effective at executing algorithms. 
To the best of our knowledge, no one has ever explicitly described SNN architectures using integral transforms either. However, a passing remark in \cite{dudzik2024asynchronous} hints that the message passing dynamic of GNNs is very similar to sheaf diffusion and thus such a generalization should be all but impossible. Hopefully, future research will shed light on these conjectures.

\clearpage{\pagestyle{empty}\cleardoublepage}


\chapter{Functor Learning}
\lhead[\fancyplain{}{\bfseries\thepage}]{\fancyplain{}{\bfseries\rightmark}}

All categorical machine learning frameworks examined in the previous chapters represent machine learning models, both trained and untrained, as morphisms in some category. Morphisms capture the core idea of compositionality and are thus a very good choice in many contexts; nevertheless, there are also cases where a simple morphism is unable to interface with the structure that one might want preserved. For instance, different datasets might be linked using morphisms in an appropriate category (see e.g. \cite{spivak2012functorial}, \cite{gavranovic2019compositional}) or sets containing machine learning data could be given a categorical structure, where elements are objects and morphisms capture relations between such objects (see e.g. \cite{lambek1999type}). In these and other cases, \textit{learning functors} instead of morphisms is advantageous as it allows us to preserve the aforementioned structure during the learning process. In this chapter we examine various approaches that testify to the usefulness of this insight: we will see how functors can be used to separate different layers of abstraction in the machine learning process (\cite{gavranovic2019compositional}), to embed data in vector spaces (\cite{sheshmani2021categorical}, \cite{coecke2010mathematical},\cite{lewis2019compositionality}), to carry out unsupervised translation (\cite{sheshmani2021categorical}), and to impose equivariance constraints and pool data effectively (\cite{chytas2024poolingimagedatasetsmultiple}).
As we will illustrate, functors can be learned by gradient descent, just like morphisms, as long as we have appropriate parametrizations (\cite{gavranovic2019compositional}) or specialized objective functions (\cite{sheshmani2021categorical}, \cite{chytas2024poolingimagedatasetsmultiple}). 

\section{Using functors to separate layers of abstraction}

The author of \cite{gavranovic2019compositional} takes inspiration from the field of categorical data migration (\cite{spivak2012functorial}) to create a categorical framework for deep learning that separates the development of a machine learning model into a number of key steps. The different steps concern different levels of abstraction and are linked by functors.

\subsection{Schemas, architectures, models, and concepts}

\begin{figure}[h]
  \begin{center}
    \includegraphics[width=0.2\textwidth]{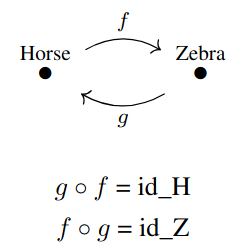}     
    \caption[CycleGAN schema]{Schema for the cycleGAN architecture. (Image taken from \cite{gavranovic2019compositional}.)}
    \label{fig: cyclegan}
  \end{center}
\end{figure}

The first step in the learning pipeline proposed by \cite{gavranovic2019compositional} is to write down the bare-bones structure of the model in question. This can be done by using a directed multigraph $G$, where nodes represent data and edges represent neural networks interacting with such data. Constraints can be added at this level in the form of a set $\mathcal{X}$ of equations that identify parallel paths (see e.g. \textit{Fig. \ref{fig: cyclegan}}). 

\begin{definition}[Model schema]
  The schema of a model represented by a multigraph $G$ is the freely generated category $\mathbf{Free}(G)$.
\end{definition}

The schema of a model does not contain any data nor does it do any computation, but it encodes the bare-bones structure of such model. If ${\sim}$ is the path congruence relation induced by the equations in $\mathcal{X}$, we can take the module $\mathbf{Free}(G)/{\sim}$ to impose the constraints represented by $\mathcal{X}$. Depending on the context, the word schema will refer either to $\mathbf{Free}(G)$ or to $\mathbf{Free}(G)/{\sim}$. As we will discuss later, each equation in $\mathcal{X}$ can be associated with a specific loss function and these losses can be used to teach models to abide by ${\sim}$.

Given a schema $\mathbf{Free}(G)$, we can choose an architecture for the model, that is, we can assign to each node a Euclidean space and to each morphism a parametric map. This procedure yields an untrained neural network.

\begin{definition}[Model architecture]
  Let $\mathrm{Free}(G)$ be a model schema. An architecture for such schema is a functor $\mathsf{Arch}: \mathrm{Free}(G) \to \mathbf{Para}_{\times}(\mathbf{Smooth})$ (see \textit{Def. \ref{def: para}} and see \textit{Ex. \ref{ex: smooth}}).
\end{definition}

$\mathsf{Arch}$ maps objects to Euclidean spaces. These might be interpreted as the spaces data will live in, but it is wiser to put data outside the $\mathbf{Para}$ machinery and in the simpler $\mathbf{Set}$ category, as this allows for better compartmentalisation. Thus, \cite{gavranovic2019compositional} also defines an embedding functor, which agrees with $\mathsf{Arch}$ on objects but exists independently of it.

\begin{definition}[Model embedding]
  Let $\mathrm{Free}(G)$ be a model schema and let $\mathsf{Arch}$ be a chosen architecture. An embedding for such schema is a functor $E: |\mathrm{Free}(G)| \to \mathbf{Set}$ which agrees with $\mathsf{Arch}$ on objects\footnote{The reason why the domain $E$ is the discretized schema $|\mathrm{Free}(G)|$ instead of the original schema $\mathrm{Free}(G)$ is clarified in \textit{Rem. \ref{rem: discretized}}.}.
\end{definition}

Now that we have a model, we must find a way to assign specific values to the parameters, so that the model can be optimized by gradient descent. Consider the function $\mathfrak{p}: (P,f) \mapsto P$, which takes the parameter space out of a parametric map in $\mathbf{Para}_{\times}(\mathbf{Smooth})$. We can use it to define the function 
\[\mathfrak{P}: \mathsf{Arch} \mapsto \prod_{f : \mathsf{Gen}_{\mathbf{Free}(G)}}\mathfrak{p}(\mathsf{Arch}(f)),\]
where $\mathsf{Gen}_{\mathbf{Free}(G)}$ is the set of generating morphisms of the free category on the multigraph $G$. The function $\mathfrak{P}$ takes and architecture and returns the parameter space. Given $\mathfrak{P}$, we can define the notion of parameter specification function.

\begin{definition}[Parameter specification function]
  Let $\mathrm{Free}(G)$ be a model schema and let $\mathsf{Arch}$ be a chosen architecture. A parameter specification function is a function $\mathsf{PSpec}$ which maps a pair $(\mathsf{Arch}, p)$ - comprised of an architecture $\mathsf{Arch}$ and some $p \in \mathfrak{P}(\mathsf{Arch})$ - to a functor $\mathsf{Model}_p: \mathbf{Free}(G) \to \mathbf{Smooth}$. The functor $\mathsf{Model}_p$ takes the model schema and returns its implementation according to $\mathsf{Arch}$, partially applying $p_f$ to each $\mathsf{Arch}(f)$, so that we obtain an actual smooth map. 
\end{definition}

\begin{figure}[h]
  \begin{center}
    \includegraphics[width=0.35\textwidth]{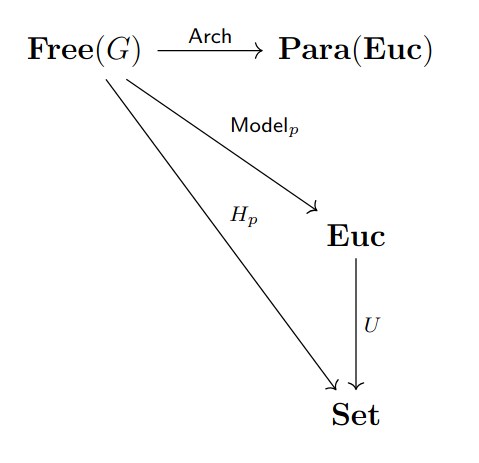}     
    \caption[Layers of abstraction]{Functors are used to link different layers of abstraction in the framework of \cite{gavranovic2019compositional}. (Image taken from \cite{gavranovic2019compositional}.)}
    \label{fig: layers}
  \end{center}
\end{figure}

The functor $\mathsf{Model}_p$ takes a schema and implements it replacing nodes with Euclidean spaces and arrows with appropriate smooth functions. Although we need to choose an architecture and specific values for its parameter in order to define $\mathsf{Model}_p$, the latter directly acts on the schema and does not pass through the architecture. See \textit{Fig. \ref{fig: layers}} for a srting diagram that depicts the relationship between $\mathsf{Arch}$ and $\mathsf{Model}_p$.

Now, if we hope to train the model we have defined, we will need a dataset. \cite{gavranovic2019compositional} suggests that a dataset should be represented as a subfunctor of the model embedding functor.

\begin{definition}[Dataset]
  Let $E$ be a model embedding. Then, a dataset is a subfunctor $D_E: |\mathbf{Free}(G)| \to \mathbf{Set}$ which maps every object $A$ of the discretized free category $|\mathbf{Free}(G)|$ to a finite subset $D_E(A) \subseteq E(A)$.
\end{definition}

\begin{remark}
  \label{rem: discretized}
  The reason why \cite{gavranovic2019compositional} defines $E$ and $D_E$ on discretized categories is because it often happens in practical machine learning that the available data is not paired. In these cases, it would be meaningless to provide an action on morphisms because they would end up being incomplete maps.
\end{remark}

Given a node $A$ of $G$, we have associated to $A$ a Euclidean space $E(A)$ and a dataset $D_E(A)$. A dataset may be considered a collection of instances of something more specific than just vectors; for instance, if we have a finite dataset of pictures of horses, we are clearly interested in the concept of horse, that is in the set of all possible pictures of horses, which is much larger than our dataset but still much smaller than the vector space used to host such pictures. It makes thus sense to define another set $\mathfrak{C}(A)$ representing such concept: this set will satisfy the inclusion relations $D_E(A) \subseteq \mathfrak{C}(A) \subseteq E(A)$. Moreover, since concepts are assumed to be complete, we can extend $\mathfrak{C}$ to a functor.

\begin{definition}[Concept functor]
  Given a schema $\mathbf{Free}(G)/{\sim}$, an embedding $E$ and a dataset $D_E$, a concept associated with this information is a functor $\mathfrak{C}: \mathbf{Free}(G)/{\sim} \to \mathbf{Set}$ such that, if $I: |\mathbf{Free}(G)| \to \mathbf{Free}(G)/{\sim}$ is the inclusion functor, $D_E \subseteq I \fatsemi \mathfrak{C} \subseteq E$.
\end{definition}

As \cite{gavranovic2019compositional} states, $\mathfrak{C}$ is an idealization, but it is a useful idealization as it represent the goal of the optimization process: given a dataset $D_E: |\mathbf{Free}(G)| \to \mathbf{Set}$, we wish to learn the concept functor $\mathfrak{C}: \mathbf{Free}(G)/{\sim} \to \mathbf{Set}$. More concretely, we want to train a model $\mathsf{Model}_p$ which is as close of an approximation of $\mathfrak{C}$ as possible\footnote{Notice that $\mathsf{Model}_p$ and $\mathfrak{C}$ have different domains and codomains. The difference of codomains is not an issue: we can just forget the differential structure of the codomain of $\mathsf{Model}_p$, which yields a functor $H_p$ with image in $\mathbf{Set}$ (see \textit{Fig. \ref{fig: layers}}). The difference of domains, on the other hand, is by design as $\mathsf{Model}_p$ can only approximate the constraints imposed by $\sim$. This is not an issue in practice as we are only interested in the performance of the image of the trained model $\mathsf{Model}_p$.}. Total achievement of such goal is clearly impossible as, even in the simplest of cases (such as linear regression on synthetic linearly generated data), finite arithmetics and the finite nature of the learning iteration prevent us from obtaining a perfect copy of the generating function.
Nevertheless, it is often possible to design an optimization process which makes the model converge towards the ideal goal.

Now that we know what the optimization goal is, we can define tasks. The task formalism brings together what has been defined in this section in an integrated fashion.

\begin{definition}[Task]
  Let $G$ be a directed multigraph, let ${\sim}$ be a congruence relation on $\mathbf{Free}(G)$ and let $D_E: |\mathbf{Free}(G)| \to \mathbf{Set}$ be a dataset. Then, we call the triple $(G,{\sim},D_E)$ a task.
\end{definition}

Once we are assigned a machine learning task $(G,{\sim},D_E)$, we have to choose an architecture, an embedding and a concept compatible with the given multigraph, equations and dataset. Then, we specify a random initial parameter with an appropriate parameter specification function. Now we can choose an optimizer, but we must be careful to design an appropriate loss function. The loss function should incorporate both an architecture specific loss and a path equivalence loss. The former penalizes wrong predictions while the latter penalizes violations of the constraints embodied by ${\sim}$.

\begin{definition}[Path equivalence loss]
  Let $(G,{\sim},D_E)$ be a task. Let $\mathsf{Model}_p$ be an associated model. Then, if $f \sim g: A \to B$ in $G$, we define the path equivalence loss associated with $f$, $g$ and $\mathsf{Model}_p$ as 
  \[\mathcal{L}_{\sim}^{f,g} = \mathbb{E}_{a \sim D_E(A)}[\|\mathsf{Model}_p(f)(a) - \mathsf{Model}_p(g)(a)\|].\]  
\end{definition}

\begin{definition}[Total loss]
  Let $(G,{\sim},D_E)$ be a task. Let $\mathsf{Arch}$ be an associated architecture, let $\mathsf{Model}_p$ be an associated model, and let $\mathcal{L}'$ be an architecture specific loss. Suppose $\gamma$ is a non-negative hyperparameter. Then, we define the total loss associated with the task, the architecture, the model, and the hyperparameter as
  \begin{equation}
    \label{eq: sumoflosses}
    \mathcal{L} = \mathcal{L}' + \gamma\sum_{f \sim g}\mathcal{L}_{\sim}^{f,g}.
  \end{equation}
\end{definition}

We can now proceed as usual, computing the loss on the dataset for a number of epochs and updating the parameter $p$ each time.

It is important to notice that, while the learning iteration employed by \cite{gavranovic2019compositional} is nothing new, the functor approach is actually novel, in that the usual optimization process is used to explore a functor space instead of a simple morphism space. This point of view offers two main advantages: on one hand, it separates different layers of abstraction are separated, which provides much needed conceptual clarity; on the other hand, it offers an explicit treatment of constraints, which are often only described implicitly and hidden away in the architecture of the model or in the loss function. 

\subsection{Datasets influence the semantics of tasks}

\cite{gavranovic2019compositional} centers its investigation around the powerful cycleGAN architecture (\cite{zhu2017unpaired}), which is described using an appropriate task consisting of the cycleGAN schema (see \textit{Fig. \ref{fig: cyclegan}}), the cycle consistency equations, and a cycleGAN dataset, that is, a dataset with two sets $A$ and $B$ containing data which is essentially isomorphic, such as pictures of horses and zebras. This description is particularly insightful because it can be used to prove a very important point: the choice of dataset influences the semantics of the learned task in non-trivial ways. In other words, changing the dataset functor can result in semantically different tasks even if the same schema and equations are retained. 

For instance, combining the cycleGAN schema with the cycleGAN equations and a cycleGAN dataset yields a task whose semantics can be described as \textit{learn maps that turn horses into zebras and vice versa}. Now consider replacing the cycleGAN dataset with a new dataset consisting of two sets $A$ and $B$, where $A$ contains pictures depicting two elements $X$ and $Y$ together, and $B$ contains separate images of $X$ and $Y$. The resulting task has very different semantics: \textit{learn how to separate $X$ from $Y$}. \cite{gavranovic2019compositional} follows this paradigm and shows how the CelebA dataset can be used to train a neural network able to remove glasses from pictures of faces, and one able to insert them.

This example is especially relevant to the present discussion because it shows how important the categorical structure of $\mathbf{Set}$ can be to machine learning. We can interpret pairs \textit{(face, glasses)} as elements of the Cartesian product of the set of faces and the set of glasses. On the other hand, the set of pictures of faces with glasses can instead be considered another categorical product of the aforementioned sets. Since categorical products are unique, we know that there must be a unique isomorphism between the two products. The task at hand can thus be interpreted as \textit{find the canonical isomorphism}. \cite{gavranovic2019compositional} labels this task as \textit{product task}.

\section{Categorical representation learning}

We now discuss frameworks where the categorical structure of data is not only used to integrate data distributed among different sets (as in \cite{gavranovic2019compositional}) but also to model relationships within the data. Functorially preserving and exploiting this structure can lead to marked improvements in effectiveness and efficiency of the models in question  because the structure of data is often closely linked with its semantics (see e.g. \cite{sheshmani2021categorical}), especially in the field of natural language processing, or NLP (see e.g. \cite{coecke2010mathematical}, \cite{lewis2019compositionality}). 

The field of \textit{categorical representation learning} (\cite{coecke2010mathematical}, \cite{lewis2019compositionality}, \cite{sheshmani2021categorical}) aims to learn vectorial representations endowed with a categorical structure mirroring the one of the original data, so that models trained on these representations can use the structure efficiently. This is often a symbol-to-vector transformation, which poses conceptual and computational challenges (see e.g. \cite{coecke2010mathematical}). We will examine two examples of categorical representation learning: the approach of \cite{sheshmani2021categorical}, which uses the obtained representations to carry out (partially) unsupervised translation, and the approaches of \cite{coecke2010mathematical} and \cite{lewis2019compositionality}, which develop a compositional distributional model of meaning for NLP.

\subsection{Unsupervised functorial translation}

We now illustrate the approach of \cite{sheshmani2021categorical} showing how it is applied to the same example described in the original paper: unsupervised translation of the names of chemical elements from English to Chinese. The authors are tasked with converting between two identical chemical compound datasets, one labeled in English and the other in Chinese. Both datasets can be given a categorical structure by considering elements as objects and chemical bonds as morphisms. Suppose $\mathcal{C}$ and $\mathcal{D}$ are the resulting categories. The authors of \cite{sheshmani2021categorical} leverage the shared structure by training a model to learn a translation functor $\mathcal{F}: \mathcal{C} \to \mathcal{D}$. To this aim, they functorially embed these categories into the vector space category defined below, which contains representations as objects.

\begin{definition}[Vector space category]
  \label{def: vecspacecat}
  Let $n \in \mathbb{N}$. Let $\mathcal{R}$ be the category whose objects are the vectors in $\mathbb{R}^n$ and such that, for all $u,v :\mathcal{R}$, 
  \[\mathcal{R}(u, v) = \{M \in \mathbb{R}^{n \times n} \text{ s.t. } v = Mu\}.\]
  Composition is ordinary matrix multiplication and the identity on $v$ is $\mathrm{id}_v = \frac{vv^T}{|v|^2}$.
\end{definition}

The embedding of $\mathcal{C}$ into $\mathcal{R}$ is a $\mathcal{C} \to \mathcal{R}$ functor which maps each object $a$ to a vector $v_a$ and each morphism $f$ to a matrix $M_f$. The actual mapping can be learned with a neural network consisting of two separate embedding layers: one mapping objects to vectors and one mapping relations to matrices. The authors of \cite{sheshmani2021categorical} train the embedding layers using co-occurrence statistics and negative sampling to make sure that the embedded morphisms actually represent the same relations as the original morphisms. Here, the authors of \cite{sheshmani2021categorical} take on a distributional point of view, positing that co-occurrence must encode such relations. In the authors' words, \dblquote{co-occurrence does not happen for no reason}. The training strategy used in the paper is the following: given two embedded words $a$ and $b$, model the probability of co-occurrence as $P(a \to b) = \mathsf{sigmoid}(z(a \to b))$, where the logit $z(a \to b)$ is defined as
\[z(a \to b) = F\left(\bigoplus_f v_a^TM_fv_b \right).\]
Here, $F$ is non-linear and $\oplus_f$ represents concatenation (or more sophisticated forms of aggregation) over all morphisms in $\mathcal{C}$. Taking on a probabilistic perspective allows us to represent non-strict relations, which is often necessary in machine learning. From this point of view, the likelihood of $f: a \to b$ is proportional to the degree of alignment between $v_b$ and $M_fv_a$ are closely aligned. The latter can be computed as $v_a^TM_fv_b$, and the non-linearity $F$ reads out these measurements into a single value. Now, the actual co-occurrence probability $p(a,b)$ of two objects $a,b : \mathcal{C}$ can be approximated directly from the dataset. Given a negative sampling distribution $p_N$ on objects unrelated to $a$, we can implement the negative sampling objective
\[\mathcal{L} = \mathbb{E}_{(a,b) \sim p(a,b)}\left(\log P(a \to b) + \mathbb{E}_{b' \sim p(b')}\log (1-P(a \to b'))\right).\]
The embedding network can then be trained by maximizing this objective function. The embedding of $\mathcal{D}$ into $\mathcal{R}$ can be defined and trained in the same way.

\begin{remark}
  Strictly speaking, morphisms in category should have a specific domain and a specific codomain but, in the case of the vector space category of \textit{Def. \ref{def: vecspacecat}}, a single matrix $M$ can be in many hom-sets of $\mathcal{R}$. This is mirrored by $\mathcal{C}$ and $\mathcal{D}$, where morphisms correspond to classes of chemical bond and the same morphism can link different objects. From this point of view, morphisms in $\mathcal{R}$, in $\mathcal{C}$, and in $\mathcal{D}$ behave more similarly to relations on sets than to morphisms in categories.  Although this line of thinking does not strictly adhere to the definition of category, it helps our intuition and greatly reduces the size of the model. We will see a possible formal solution to this problem later on in this section, when we discuss the approach of \cite{coecke2010mathematical} and \cite{lewis2019compositionality}.
\end{remark}

The procedure described above can be been applied to both $\mathcal{C}$ and $\mathcal{D}$ to obtain meaningful vectorial representations of both datasets. An English to Chinese translation functor $\mathcal{F}: \mathcal{C} \to \mathcal{D}$ can be similarly embedded as an $\mathcal{R} \to \mathcal{R}$ functor by precomposition with the $\mathcal{C} \to \mathcal{R}$ embedding and postcomposition with the inverse of the $\mathcal{D} \to \mathcal{R}$ embedding. Such functor must equate chemical bonds of the same kind, for example, if $f$ is a covalent bond so is $\mathcal{F}(f)$. It is argued in \cite{sheshmani2021categorical} that this functor can be represented by a matrix $V_\mathcal{F}$ so that $v_{\mathcal{F}(a)} =  V_\mathcal{F}v_a$ and $M_{\mathcal{F}(f)} = V_\mathcal{F}M_f$. This representation is only meaningful if (i) $V_{\mathcal{F}}M_f = M_{\mathcal{F}}V_{\mathcal{F}}$ for all $f$, (ii) $V_{\mathcal{F}}M_{\mathrm{id}_a} = M_{\mathrm{id}_{\mathcal{F}(a)}}$ for all $a$, and (iii) $V_{\mathcal{F}}M_{f \circ g} = V_{\mathcal{F}}M_fV_{\mathcal{F}}M_g$ for all $f,g$. This is not true for all choices of  $V_\mathcal{F}$ but, if we choose every $v_a$ to be a unit vector, and if we constrain $V_{\mathcal{F}}$ to be orthogonal, (ii) and (iii) are trivially satisfied. Requirement (i), on the other hand, can be learned by minimizing the following structure loss:
\[\mathcal{L}_{\mathrm{struc}} = \sum_{f}\|V_{\mathcal{F}}M_f - M_{\mathcal{F}(f)}V_{\mathcal{F}}\|^2.\]
As the authors of \cite{sheshmani2021categorical} remark, this loss is universal, in the sense that it does not depend on any specific object, but acts on the morphisms themselves. While this approach is very elegant and does indeed return a functor, it might not produce the functor we expect because $V_{\mathcal{F}}$ is not unique if the $M_f$ happen to be singular. Thus, it is better to integrate the structure loss with a second alignment loss that introduces some supervision to the unsupervised translation task. For instance, if the value of $\mathcal{F}(a)$ is known for a set $A$ for objects, we can define
\[\mathcal{L}_{\mathrm{align}} = \sum_{a \in A}\|V_{\mathcal{F}}v_a - v_{\mathcal{F}(a)}\|.\]
Then, the total loss can be written as a weighted sum 
\begin{equation}
  \label{eq: sumoflosses2}
  \mathcal{L} = \mathcal{L}_{\mathrm{align}} + \lambda \mathcal{L}_{\mathrm{struct}},
\end{equation}
where $\lambda$ is a hyperparameter that regulates the relative importance of the two losses.

\begin{remark}
  It is interesting to compare the total loss in \textit{Eq. \ref{eq: sumoflosses2}} to the one in \textit{Eq. \ref{eq: sumoflosses}}. In both cases, the total loss is obtained as a linear combination of a model specific loss and a second more abstract loss. Both abstract losses enforce equality between morphisms but the abstract loss in \textit{Eq. \ref{eq: sumoflosses2}} does not enforce functoriality because $\mathsf{Model}_p$ is already a functor by definition in the framework of \cite{gavranovic2019compositional}. This highlights the main difference between the approach of \cite{gavranovic2019compositional} and the approach of \cite{sheshmani2021categorical}: the latter requires functoriality to be learned from the structure of data, while the former takes functoriality as a given.
\end{remark}

According to \cite{sheshmani2021categorical}, the categorical approach described above can also be strengthened by endowing the categories $\mathcal{C}$, $\mathcal{D}$, and $\mathcal{R}$ with monoidal products that make combining objects into higher-level structures possible. In particular, the monoidal structure $\otimes'$ of $\mathcal{R}$ can be given as
\begin{align*}
  v_a \otimes' v_b &= \Theta(v_a \otimes v_b),\\
  M_f \otimes' M_f &= \Theta(M_f \otimes M_g),\\
\end{align*}
where $\otimes$ is the Kronecker tensor product and $\Theta$ is a learned operator that sends the products back into the original spaces. This monoidal structure can then be used to mine categorical structure on multiple levels: for instance, in the running example of unsupervised translation of chemicals, it can be used to derive vectorial representations for functional groups or even whole compounds. If the data has a non-obvious high-level structure, \cite{sheshmani2021categorical} suggests a bootstrap approach where different possible links are randomly tested and the stronger ones are selected.

It is shown in \cite{sheshmani2021categorical} that the categorical representation learning framework described in the paper and the associated functor learning architecture for unsupervised translation can be successfully implemented. Benchmark tests against traditional sequence-to-sequence models show that the functorial paradigm leads to marked improvements in efficiency of learning. In particular, the authors compare the functorial model described in the paper with a GRU cell model of similar performance, noting that the former needs 17 times more parameters than the latter to learn to translate element names with a similar accuracy. The authors also compare their approach with the multi-head attention mechanism of \cite{vaswani2017attention}, arguing that the matrices $M_f$ are essentially equivalent to the products $Q_f^TK_f$, where $Q_f$ is a query matrix and $K_f$ is a key matrix. Keeping $M_f$ united emphasizes the important role of functoriality and provides intuition concerning the nature of the vectorial representations of objects: in fact, it makes sense to interpret each $M_f$ as a metric that distorts the space $\mathbb{R}^n$ and makes two vectors $v_a$ and $v_b$ closer if and only if there is a high likelihood that $a \stackrel{f}{\longrightarrow} b$.

\subsection{Compositional distributional model for NLP}

There are two general approaches to natural language processing (NLP): the \textit{compositional} (also known as \textit{distributional}) approach and the \textit{contextual} approach. The former aims to exploit the symbolic structure of grammar to understand text, while the latter posits that many aspects of text can be understood from distributional properties of the words (\cite{amigo2022information}). While compositionality has been very popular with theoreticians due to its elegance and its abstract structure, contextuality has been successfully applied by practitioners, yielding remarkably effective NLP models that learn the meaning of words and utterances by embedding them in vector spaces based on contextual information. The fact that both compositionality and contextuality are successful in their domains hints that the two approaches should be seen as complementary and that it would be desirable to integrate them both in a single \textit{compositional distributional framework}\footnote{Cognitive science faces a similar challenge in integrating the competing connectionists and symbolic models of the human mind. This is the central theme of \cite{smolensky2006harmonic}, whose approach inspired \cite{clark2007combining} (discussed in this section). The important role of language in cognition suggests that advances in the development of a compositional distributional model of language may yield advances in cognitive science and vice versa.} (see e.g. \cite{amigo2022information}, \cite{maruyama2019compositionality}). 

There have been numerous attempts to develop such a framework in the last few decades. For instance, \cite{clark2007combining} proposes to represent both text and the structural roles of its constituents as vectors, which can then be combined using sums and tensor products. We will focus on the approach of \cite{coecke2010mathematical}, whose authors take inspiration from the ideas in \cite{clark2007combining} and propose a framework that uses the language of compact closed categories to unite symbolic grammar rules with concrete representation vectors. Grammar is represented as a pregroup category, whereas vectorial representations are learned on vector spaces which are then organized within $\mathbf{FVect}$, the category of finite dimensional real vector spaces\footnote{We work under the same assumption as \cite{coecke2010mathematical}, namely, that each vector space in $\mathbf{FVect}$ is endowed with a scalar product $\langle-,-\rangle$.}. We start by defining pregroup categories.

\begin{definition}[Pregroup]
  A pregroup is a monoidal poset category $P$ where each object $a$ has a left adjoint $a^l$ and a right adjoint $a^r$. The adjoints must be such that the following are true:
  \begin{align*}
    a^la \leq &1 \leq aa^l,\\
    aa^r \leq &1 \leq a^ra,
  \end{align*}
  where $1$ is the monoidal unit of $P$.
\end{definition}

The use of pregroups to model grammatical structures dates back to \cite{lambek1999type}. The key idea is that we can define a grammar category as the free pregroup $P$ generated by an initial set of basic grammatical roles. Then, we can represent various parts of speech using basic objects, their adjoints, and their monoidal products. The poset structure of the pregroup serves as a system of reduction rules which can be used to prove if a sentence is grammatical. Consider the following example (taken from \cite{lewis2019compositionality}).

\begin{figure}[h]
  \begin{center}
    \begin{subfigure}{0.32\textwidth}
      \includegraphics[width=\textwidth]{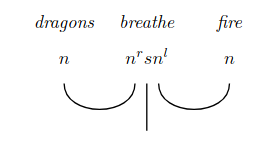}
      \caption{}
    \end{subfigure}
    \begin{subfigure}{0.47\textwidth}
      \includegraphics[width=\textwidth]{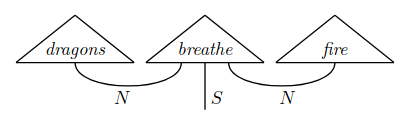}
      \caption{}
    \end{subfigure}         
    \caption[Pregroup representation of grammar]{Two diagrammatic representations of the grammatical structure of the sentence \dblquote{Dragons breathe fire.} (Images taken from \cite{lewis2019compositionality}.)}
    \label{fig: pregroup}
  \end{center}
\end{figure}

\begin{example}
  Suppose $n$ represents the role of a noun and $s$ represents the role of a sentence. Then, an English transitive verb can be represented by the product $n^rsn^l$ in the free pregroup $P$ generated by $s$ and $n$. Thus, the grammatical structure of the sentence \dblquote{Dragons breathe fire.} is
  \begin{equation}\
    \label{eq: grammreduction}
    n(n^rsn^l)n = nn^rsn^ln = (nn^r)s(n^ln) \leq 1s1 \leq s.
  \end{equation}
  The pregroup axioms show that the structure of  \dblquote{Dragons breathe fire.} can be reduced to the sentence type $s$. This is proof that the sentence is grammatical. More complex patter can be modeled by adding more grammatical types or by combining types in other ways. These kinds of computations have a nice representation in the form of diagrams such as the ones of \textit{Fig. \ref{fig: pregroup}}.
\end{example}

The pregroup grammar framework is useful to analyze the basic structure of language but gives no access to the actual meaning of words and sentences. The compositional distributional model of \cite{coecke2010mathematical} solves this problem by building a functor $\Phi: P \to  \mathbf{FVect}$. Given a certain pregroup grammar category $P$, for every grammatical type $x: \mathcal{P}$, we can associate a vector space $X:\mathbf{FVect}$ containing vectorial representations of utterances of type $x$. $\Phi$ can be defined so that it preserves the pregroup structure of $P$ thanks to the following proposition (\cite{coecke2010mathematical}).

\begin{proposition}
  Consider the category $\mathbf{FVect}$ of finite-dimensional real vector spaces. Suppose $\otimes$ is the tensor product of vector spaces. Then, $(\mathbf{FVect}, \mathbb{R}, \otimes)$ is a monoidal category.  
  For all $V: \mathbf{FVect}$, define $V^l = V^r = V$ and consider the (co)unit linear maps:
  \begin{align*}
    \eta_V^l = \eta_V^r: \mathbb{R} \to V \otimes V, \,\,\, &1 \mapsto \sum_i e_i \otimes e_i,\\
    \epsilon_V^l = \epsilon_V^r: V \otimes V \to \mathbb{R}, \,\,\, &\sum_{i,j} c_{ij} e_i \otimes e_j \mapsto \sum_{i,j} c_{ij} \langle e_i, e_j \rangle,
  \end{align*}
  where $e_1, \dots, e_n$ is a basis for $V$. Then, $(\mathbf{FVect}, \mathbb{R}, \otimes, (-)^l, (-)^r, \eta^l, \eta^r, \epsilon^l, \epsilon^r)$ is a compact closed category.
\end{proposition}

Now, pregroups have a trivial compact-closed structure as well. Thus, the functor $\Phi$ can be defined so that it preserves the shared compact-closed structure just by mapping the poset relatioins of $P$ to the appropriate linear maps of $\mathbf{FVect}$. In practice, $\Phi$ can be used to take a formal reduction in $P$ and implement it in $\mathbf{FVect}$ as a series of operations on vectorial representations. The goal of this process is to take vectorial representations of single words in a sentence and obtain the vectorial representation of the meaning of the sentence. The compositional distributional model is thus compositional because it is build around the compositional structure of grammar and distributional because it allows us to encode distributional information in vectorial representations of words and sentences.

\begin{example}
  The reduction of \textit{Eq. \ref{eq: grammreduction}} can be seen as a morphism $nn^rsn^ln \to s$. The image of this morphism through $\Phi$ is a linear map 
  $N \otimes N \otimes S \otimes N \otimes N \to S.$
\end{example}

\begin{remark}
  Differently from \cite{sheshmani2021categorical}, \cite{coecke2010mathematical} does not prescribe a specific method for learning the representations of individual words (some possible strategies are listed in the original paper), but it specifies a strategy to combine existing representations to generate representations of more complex utterances up to the level of whole sentences. It is interesting to notice that both papers suggest using tensor products in order to combine representations of lower-level structures into representations of higher-level structures.
\end{remark}

String diagrams such as the ones in \textit{Fig. \ref{fig: pregroup}} can be written down as binary trees thanks to the associativity of the monoidal products of $P$ and $\mathbf{FVect}$. 
\cite{lewis2019compositionality} proposes to exploit this fact and augment the compositional distributional model of \cite{coecke2010mathematical} by implementing the (co)unit maps of $\mathbf{FVect}$ as one or more recursive neural network (TreeRNN) cells. The paper argues that the use of TreeRNN cells should make the model easier to train and more robust. To our knowledge, such proposal has not yet been implemented, although TreeRNNs have already been applied to NLP many times (see e.g. \cite{ahmed2019improving}).

\section{Equivariant neural networks as functors}

\cite{chytas2024poolingimagedatasetsmultiple} presents a categorical framework that describes neural network layers as functors and uses their functorial nature to impose invariance and equivariance constraints. In particular, equivariance constraints are used to effectively transport unary operators from the dataset into a latent embedding layer. It is shown that this enables the creation of models that account for shift and imbalance in covariates when training on pooled medical image datasets. In this section, we describe the framework of \cite{chytas2024poolingimagedatasetsmultiple} and we draw a short comparison with the work of \cite{pfrommer2024transport}, which also deals with the transport of algebraic structure into embedding layers.

\subsection{An equivariant classifier to diagnose Alzheimer's disease}

The authors of \cite{chytas2024poolingimagedatasetsmultiple} consider a data category $\mathcal{S}$ whose objects $s$ are data points and whose morphisms $f: s_1 \to s_2$ represent differences in covariates. An example considered in the paper is the following: suppose the objects $s$ are comprised of brain scans and associated information concerning patient age and other covariates. The goal is to develop a model trained to diagnose Alzheimer's disease from the scans. An example of morphism in such a data category is $f_x: s_1 \to s_2$, which indicates a difference of $x$ years in age: $s_2.\mathrm{age} = s_1.\mathrm{age} + x$. Notice that this kind of operator is clearly endowed with an algebraic structure as $f_x \fatsemi f_y = f_{x+y}$. Since we are dealing with a classification task, the dataset at hand has labels. It is important not to include the labels in the data category, else any classifier model would just read such labels instead of learning to predict them. Use the notation $\mathbf{y}_s$ to represent the label associated to $s$. 

Now, if we consider another category $\mathcal{T}$, we can use a functor $F: \mathcal{S} \to \mathcal{T}$ to project $\mathcal{S}$ onto $\mathcal{T}$. Learning this functor instead of a simple map between objects is advantageous because the functoriality axioms imply that $F$ automatically satisfies equivariance constraints. This is trivial: if $g: s_1 \to s_2$, then
\begin{equation}
  \label{eq: categorical_equivariance}
  F(g): F(s_1) \to F(s_2).
\end{equation}
Invariance with respect to $g$ is not much harder to define: it suffices to impose $F(s_1) = F(s_2)$ and $F(g) = \mathrm{id}_{F(s_1)}$.

\begin{remark}
  \textit{Eq. \ref{eq: categorical_equivariance}} is a categorical generalization of the more usual group-theoretical notion of invariance, as defined in \textit{Def. \ref{def: equivariance}}. To be precise, \textit{Def. \ref{def: equivariance}} is equivalent to \textit{Eq. \ref{eq: categorical_equivariance}} if $\mathcal{S}$, $\mathcal{T}$ are Borel spaces and $g$, $F(g)$ are group actions, as highlighted in \cite{chytas2024poolingimagedatasetsmultiple}.
\end{remark}

\begin{figure}[h]
  \begin{center}
    \includegraphics[width=0.35\textwidth]{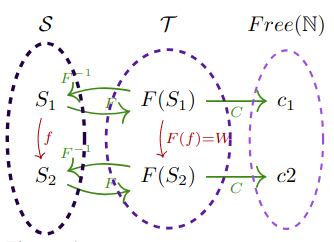}     
    \caption[Classifier]{Classifier model proposed by \cite{chytas2024poolingimagedatasetsmultiple} (Image taken from \cite{chytas2024poolingimagedatasetsmultiple}.)}
    \label{fig: classifier}
  \end{center}
\end{figure}

The authors of \cite{chytas2024poolingimagedatasetsmultiple} solve the Alzheimer's disease diagnosis task using a classifier that actively exploits the equivariant nature of functors. The proposed architecture consists of two modules: an invertible functor $F: \mathcal{S} \to \mathcal{T}$ that embeds the data into a latent space, and a functor $C: \mathcal{T} \to \mathbf{Free}(\mathbb{N})$ that does the actual classification (see \textit{Fig. \ref{fig: classifier}} for a diagrammatic representation). Here, $\mathcal{T}$ acts as a latent space and the equivariance of $F$ forces the representations in $\mathcal{T}$ to be robust with respect to covariate shifts and imbalances, so that the actual classification operated by $\mathcal{C}$ can be more effective at diagnosing the disease. The whole model can be compactly represented as $F \fatsemi C: \mathcal{S} \to \mathbf{Free}(\mathbb{N})$. Although the inverse of $F$ does not appear in the formula, its existence is needed to ensure that the latent space accurately represents the data.

The classifier described above is implemented using similar tools as the ones employed by \cite{sheshmani2021categorical}. The latent space $\mathcal{T}$ can be defined as a vector space category as in \textit{Def. \ref{def: vecspacecat}}. The functor $F$ is realized as the encoder of an appropriate autoencoder, and the decoder portion of this autoencoder acts as the inverse $F^{-1}$. The maps $f: \mathcal{S}(s_1,s_2)$ can be embedded as matrices as described in the previous section. In particular, \cite{chytas2024poolingimagedatasetsmultiple}  suggests using orthogonal matrices $F(f) = W_f$ so that the resulting transformations can be efficiently inverted.
Finally, $C$ can be implemented as a MLP classifier.

Now that the architecture has been specifies in terms of neural networks, the resulting model can be trained using gradient descent and a linear combination of three separate losses: $\mathcal{L} = \gamma_1\mathcal{L}_r + \gamma_2\mathcal{L}_p + \gamma_3\mathcal{L}_s$, where $\gamma_1$, $\gamma_2$, and $\gamma_3$ are hyperparameters. Here, $\mathcal{L}_r$ is a reconstruction loss, which makes sure that $F$ is invertible and that its inverse accurately reconstructs the original data; $\mathcal{L}_p$ is a prediction loss, which makes sure that $F \fatsemi C$ accurately predicts the labels of the data; $\mathcal{L}_s$ a structure loss, which makes sure that $F$ acts as a functor and not just a map. In formulas,
\[\mathcal{L}_r = \sum_{s : \mathcal{S}}\|s-(F^{-1} \circ F)(s)\|^2_2,\]
\[\mathcal{L}_p = \sum_{s : \mathcal{S}}\mathsf{crossentropy}(\mathbf{y}_s,(C \circ F)(s)),\]
\[\mathcal{L}_s = \sum_{\begin{smallmatrix}s_1, s_2 : \mathcal{S} \\ f: s_1 \to s_2 \end{smallmatrix}}\|W_fF(s_1)-F(s_2)\|^2_2.\]

\begin{remark}
  Notice that the amount of equivariance constraints imposed can vary with no substantial changes to the proposed architecture: it suffices to add or remove terms from $\mathcal{L}_s$. This is in stark contrast with approaches such as \cite{lokhande2022equivariance}, where handling multiple covariates requires a complicate multi-stage model. 
\end{remark}

\begin{figure}[h]
  \begin{center}
    \includegraphics[width=0.7\textwidth]{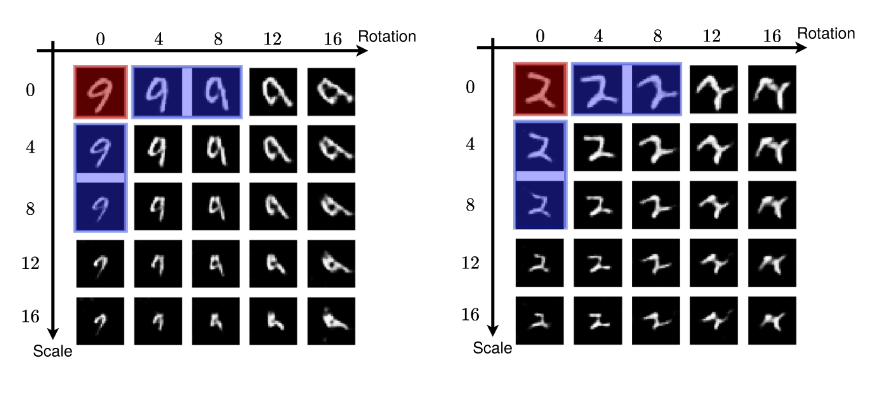}     
    \caption[Scaling and rotations of MNIST images]{The authors of \cite{chytas2024poolingimagedatasetsmultiple} applied rotations and scaling on MNIST images using the operators learned in the latent space. Despite being learned separately, the two transformations were successfully combined. (Image taken from \cite{chytas2024poolingimagedatasetsmultiple}.)}
    \label{fig: rotationsscaling}
  \end{center}
\end{figure}

The authors of \cite{chytas2024poolingimagedatasetsmultiple} test the validity of the proposed approach with two interesting experiments: a proof of concept trained on the MNIST dataset, and a working classifier trained on the ADNI brain imaging dataset. The proposed MNIST model implements equivariance with respect to increments, rotations, and zooming. It is shown in the paper that representing the associated morphisms with orthogonal matrices allows such morphisms to be inverted and combined in the latent space. A subsequent application of $F^{-1}$ shows the results of the aforementioned manipulation in human-understandable form. Such results are indeed very promising: the authors are able to combine rotations and scaling successfully, even though the network was only trained to apply them separately (see \textit{Fig. \ref{fig: rotationsscaling}}) for an example. The ADNI classifier model also shows promising results which are on par with state-of-the-art models that do not use categorical tools. The comparison takes place according to accuracy of prediction, maximum mean discrepancy, and adversarial validation.

\subsection{Transporting algebraic structure into embedding spaces}

Although stated in terms of equivariance conditions, the work of \cite{chytas2024poolingimagedatasetsmultiple} can be seen from an algebraic point of view as a framework that allows to transport unary operators from a space of data to an embedding space. This line of thinking has also been explored by \cite{pfrommer2024transport}. While \cite{pfrommer2024transport} deos not use category theory, the authors themselves state that their approach could be easily categorified using the language of categorical algebras (\textit{Def. \ref{def: endalg}}). The main innovation of \cite{pfrommer2024transport} is the use of \textit{mirrored algebras} to fill the gaps between original data and the embeddings.

\begin{figure}[h]
  \begin{center}
    \includegraphics[width=1\textwidth]{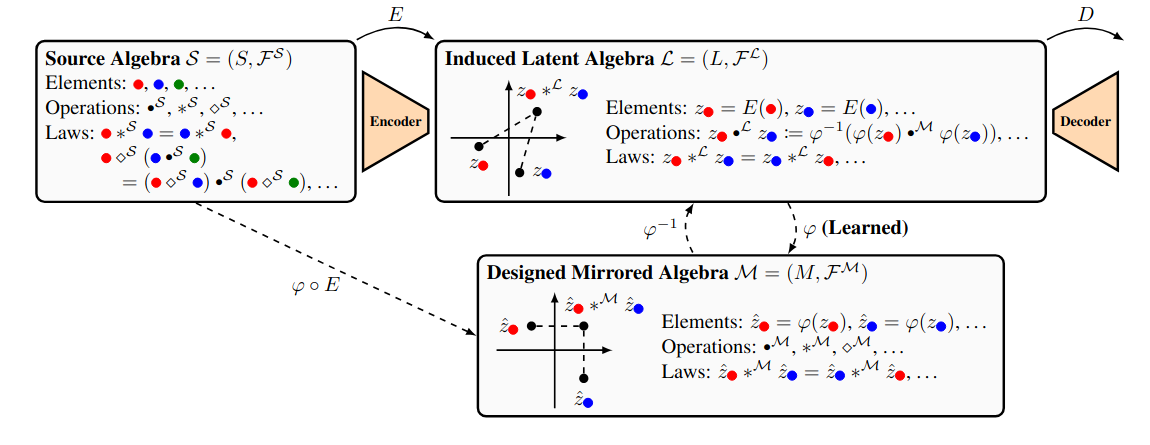}     
    \caption[Transport of algebraic structure]{Method proposed by \cite{pfrommer2024transport} to transport algebraic structure. (Image taken from \cite{pfrommer2024transport}.)}
    \label{fig: algtransport}
  \end{center}
\end{figure}

Consider a pre-trained autoencoder consisting of an encoder $E: \mathcal{S} \to \mathcal{L}$ and a decoder $D: \mathcal{L} \to \mathcal{S}$, where $\mathcal{S}$ is a data space endowed with an algebraic structure, such as a monoidal product $\bullet^{\mathcal{S}}$, and where $\mathcal{L}=\mathbb{R}^n$ is a latent space. Suppose we want to transport this algebraic structure to $\mathbb{R}^n$ through $E$, that is, we want to find a monoidal product $\bullet^{\mathcal{L}}$ over $\mathcal{L}$ so that $E$ acts as a $(\mathcal{S},\bullet^{\mathcal{S}}) \to (\mathcal{L},\bullet^{\mathcal{L}})$ homomorphism. While finding a product $\bullet^{\mathcal{L}}$ that makes $(\mathcal{S},\bullet^{\mathcal{S}})$ and $(\mathcal{L},\bullet^{\mathcal{L}})$ homomorphic might be easy, there is no guarantee that $E$ is an actual homomorphism with respect to this choice of algebraic structure. \cite{chytas2024poolingimagedatasetsmultiple} solves this problem for unary operators by learning appropriate orthogonal matrices, but it is not clear how this approach could be scaled to operators with arities greater than $1$.  \cite{pfrommer2024transport} suggests an alternative approach consisting in learning a bijection $\phi$ between the latent space and a copy $\mathcal{M} = \mathbb{R}^n$ of this space. Endow $\mathcal{M} = \mathbb{R}^n$ with a product $\bullet^{\mathcal{M}}$ which enjoyes similar properties as $\bullet^{\mathcal{S}}$ (see \textit{Fig. \ref{fig: algtransport}}). We will say that $(\mathcal{M},\bullet^{\mathcal{M}})$ is the mirrored algebra of $(\mathcal{S},\bullet^{\mathcal{S}})$. Now train $\phi$ so that $E$ is $(\mathcal{S},\bullet^{\mathcal{S}}) \to (\mathcal{L},\bullet^{\mathcal{L}})$ homomorphism, where $\bullet^{\mathcal{L}}$ is defined by
\[l_1 \bullet^{\mathcal{L}} l_2 = \phi^{-1}(\phi(l_1) \bullet^{\mathcal{M}} \phi(l_2)).\]

As stated in \cite{pfrommer2024transport}, the bijection $\phi$ can be implemented as an invertible neural network and can be trained using a loss in the form
\[\mathcal{L} = \frac{1}{N}\sum_{i=1}^N d(D(p_i^{\mathcal{L}}(E(s_1),\dots,E(s_{n_i}))),p_i^{\mathcal{S}}(s_1, \dots, s_{n_i}))\]
where the $p_i^{\mathcal{S}}$ are algebraic expressions sampled from the set of expressions over $(\mathcal{S},\bullet^{\mathcal{S}})$, the $s_j$ are sampled from $\mathcal{S}$, $p_i^{\mathcal{L}}(l_1,\dots,l_{n_i})$ is defined as $\phi^{-1}(p_i^{\mathcal{M}}(\phi(l_1),\dots,\phi(l_{n_i})))$, $p_i^{\mathcal{M}}$ is the translation of $p_i^{\mathcal{S}}$ in terms of the algebraic structure of $(\mathcal{M},\bullet^{\mathcal{M}})$, and $d$ is an appropriate distance function.

We only presented a sketch of the work in \cite{pfrommer2024transport}: the original paper presents the framework in terms of any general algebraic structure, deals with laws that regulate the interaction of different operations, proves theoretical guarantees, and explores the limitations of the proposed approach. It is shown in the paper that it is not always possible to find a mirrored algebra whose operations satisfy the same laws of the source algebra, which means that it is not always possible to transport this algebraic structure in its entirely. Nevertheless, the authors carry out experiments that prove that there are tangible advantages even in transporting a subset of the algebraic laws. The experiments support the following conjecture: \dblquote{learned latent space operations will achieve higher performance if they are constructed to satisfy the laws of the underlying source algebra} (\cite{pfrommer2024transport}). 

\clearpage{\pagestyle{empty}\cleardoublepage}


\chapter{Detailed Representations of Neural Networks}
\lhead[\fancyplain{}{\bfseries\thepage}]{\fancyplain{}{\bfseries\rightmark}}

\begin{figure}[h]
  \begin{center}
    \includegraphics[width=0.5\textwidth]{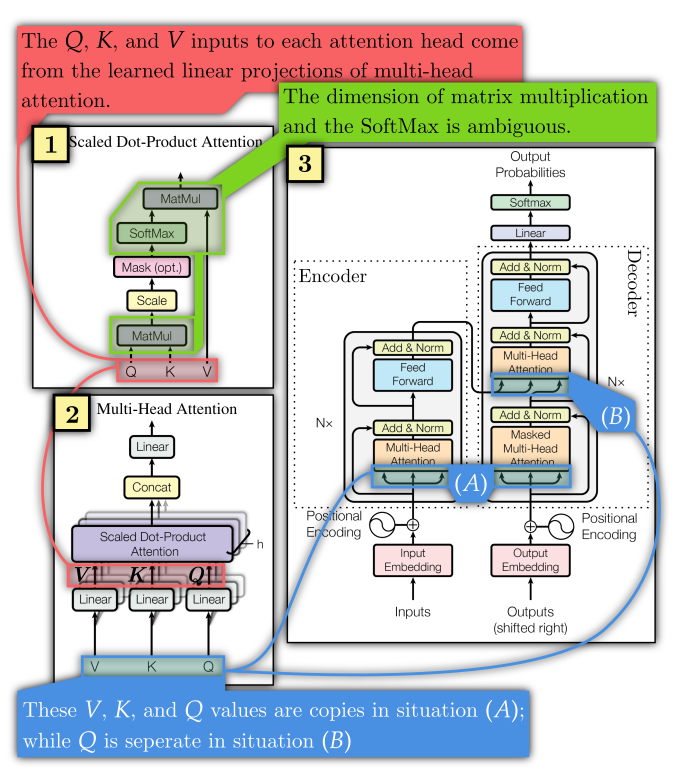}     
    \caption[Diagram of a transformer]{Schematics for a transformer as they were presented in the original paper \cite{vaswani2017attention}. The annotation are from the author of \cite{abbott2024neural}. (Image taken from \cite{abbott2023robust}.)}
    \label{fig: transformer}
  \end{center}
\end{figure}

One of the main concerns arising from machine learning research is low reproducibility. This is not a simple issue and investigation by many authors (see e.g. \cite{raff2019step}, \cite{pineau2021improving}, \cite{gundersen2022sources}) has suggested that many factors might be at play (e.g., bad incentives, unsatisfying experiment design, incorrect evaluation practices, and so on). Source code availability is a recurring theme in these discussions: for instance \cite{gundersen2022sources} observes that \dblquote{code describes implementation details perfectly and is required for outcome reproducibility for experiments of some complexity}. Similarly, \cite{pineau2021improving} describes efforts of the NeurIPS19 conference to improve reproducibility by requiring authors to provide source code. The paper justifies this requirement by speculating that the reproducibility issues affecting machine learning research might be partly caused by a reporting problem within the community, as machine learning papers often do not report all the details needed to reimplement the models they describe. For instance, it is argued that hyperparameters are often missing, together descriptions of entire components of the models at hand and descriptions of the pipeline and external tools employed.

While making code and datasets as available as possible can certainly be considered a good practice, it is argued in \cite{abbott2024neural}, \cite{khatri2024anatomy}, and \cite{phuong2022formal} that over-reliance on code is detrimental to research. In fact, while it is tautologically true that code perfectly describes implementation, code is hardly effective at capturing the essential structure of a machine learning model. Even high-level code (like Python or R code) tends to be excessively verbose and drowns important details in a sea of instructions that have little to do with the theory at play. Hence, the question remains: how should machine learning researches represent their models? This question is extremely relevant in deep learning, where the size and complexity of models is ever increasing. For instance, even the ubiquitous transformer introduced by \cite{vaswani2017attention} is a complex piece of machinery with many components, and it is very challenging to provide a satisfying description.

Many authors describe their models through equations, but this becomes more and more difficult as the size of the models grows. This is especially true especially if complex tensor operations need to be carried out as ordinary tensor notation is not very readable (\cite{xu2023graph}). Interestingly, the number of equations contained in a paper has been found to be negatively correlated to reproducibility (\cite{raff2019step}). An alternative approach that is becoming increasingly common is the use of specially crafted diagrams. For instance, \textit{Fig. \ref{fig: transformer}} shows the transformer blueprint as provided in the original paper \cite{vaswani2017attention} (the annotations are not from the original authors). \textit{Ad hoc} diagrams are intuitive but, as argued in \cite{abbott2024neural} and \cite{khatri2024anatomy}, they suffer from a lack of systematicity and do not usually have significant mathematical properties.
Moreover, important information is often omitted from the diagrams and the reader has to carefully parse the text to (hopefully) find or deduce it. For instance, the diagrams in \textit{Fig. \ref{fig: transformer}} do not provide any information concerning the size of the matrices $Q$, $K$, and $V$, nor is it clear across which dimension the softmax should be computed.

The \textit{ad hoc} diagrams often employed by machine learning researchers should be compared with the monoidal string diagrams that are almost standardized across applied category theory (see e.g. \cite{bonchi2022string}). We have already seen many such diagrams in the previous chapters, especially when describing optics (see e.g. \textit{Fig. \ref{fig: lenses}}). On top of being standardized, monoidal string diagrams have precise categorical semantics and can thus be used to write down rigorous mathematical proofs and computations. Given these good properties, it is reasonable to ask whether the systematic use of monoidal string diagrams to represent machine learning models might be helpful. Regrettably monoidal string diagrams are also very inefficient at representing the details of tensor operations. The reader can appreciate this from \textit{Fig. \ref{fig: suplearnlenses}}, where dimensionality is all but hidden. This is the price that needs to be paid for generality, as monoidal string diagrams can represent mathematical objects that go well-beyond tensors. Nevertheless, there have been attempts by \cite{abbott2023robust} and by \cite{khatri2024anatomy} to adapt the box-wire notation of monoidal string diagrams so that it can perfectly represent a neural network architecture without leaving out important details. Category theory can then be used to provide rigorous semantics to these diagrams. Hopefully, such diagrams will be able to convey the information that is necessary for implementation and theoretical analysis, leading to improved insight and better reproducibility.

In this chapter, we discuss the two aforementioned categorical diagrammatic approaches and we compare them with other non-categoric and even non-graphical approaches to the problem of representing a neural network. The papers we discuss measure the validity of their approaches against the problem of representing the transformer architecture, which is both well-known and highly non-trivial. We will reproduce some of the diagrams in questions to give an idea of how the approaches compare. Regrettably, we lack the necessary space to fully describe the categorical semantics of the diagrams in question, but the reader can find complete accounts of each approach in the respective papers.

\section{Neural circuit diagrams}

\cite{abbott2023robust} proposes to compensate for the shortcomings of monoidal string diagrams by introducing a novel kind of string diagrams: neural circuits diagrams. neural circuits diagrams are a specialization of a broader class of powerful string diagrams known as functor string diagrams, developed in order to represent objects, morphisms, functors, natural transformations, and products in a single diagram. In this section, we provide a brief description of both classes of diagrams, and we show how neural circuits diagrams can be used to describe the transformer architecture.

\subsection{Functor string diagrams}

Despite their unquestionable usefulness, monoidal string diagrams have are fundamentally limited by their inability to represent functors and natural transformations. \cite{marsden2014category} and \cite{nakahira2023diagrammatic} solve this issue by relying on colors and symbols, but this only shifts the problem as their diagrams are unable to effectively represent products (\cite{abbott2024functor}). Functor string diagrams follow a different approach: they represent both objects and functors as wires and both morphisms and natural transformations as boxes\footnote{In functor string diagrams, the actual box is often not drawn for the sake of simplicity. In these cases, the box is replaced by a single letter, but we still use the word \dblquote{box} for lack of better terminology.}. These diagrams are built on two fundamental principles, which are already observed in traditional string diagrams. Following these principles ensures that the diagrammatic syntax stays consistent with the underlying categorical structure as it becomes more expressive. 

\begin{principle}[Vertical section decomposition]
  A string diagram can be divided into vertical columns. A single vertical columns must either contain only objects or only morphisms. Object columns and morphism columns must alternate each other.
\end{principle}

\begin{principle}[Equivalent expression]
  Every newly introduced piece of graphical notation must be expressible using the notation already present in a compatible manner.
\end{principle}

\begin{figure}[h]
  \begin{center}
    \begin{subfigure}{0.39\textwidth}
      \includegraphics[width=\textwidth]{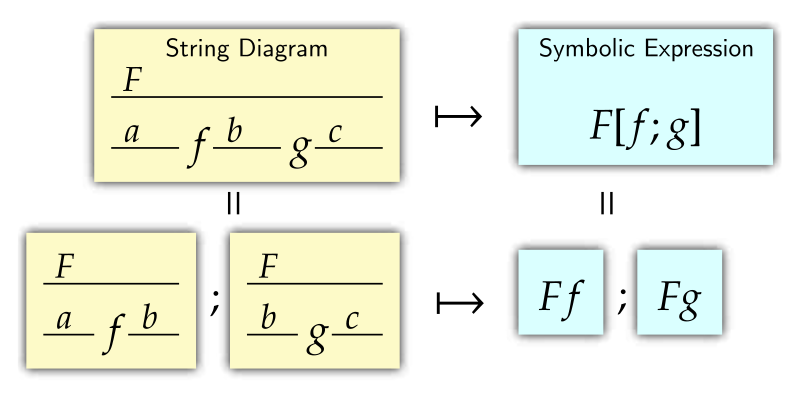}
      \caption{}
    \end{subfigure}
    \begin{subfigure}{0.6\textwidth}
      \includegraphics[width=\textwidth]{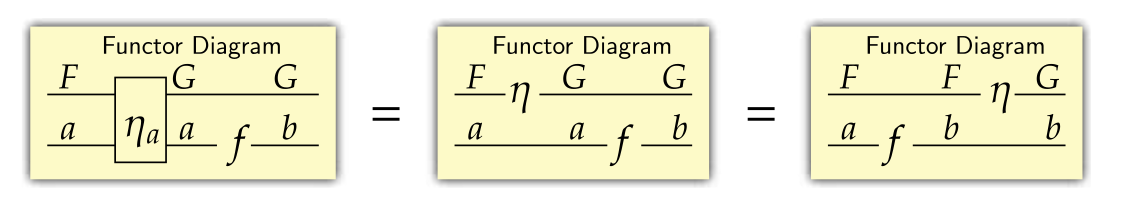}
      \caption{}
    \end{subfigure}
    \begin{subfigure}{0.5\textwidth}
      \includegraphics[width=\textwidth]{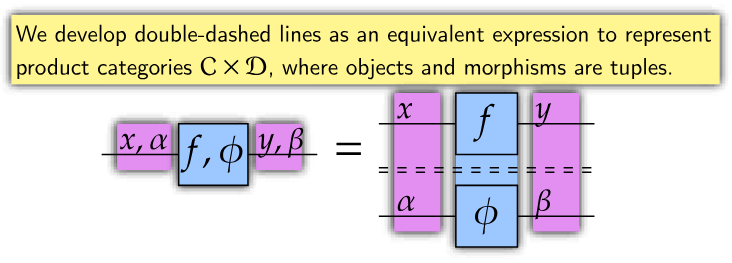}
      \caption{}
    \end{subfigure}
    \begin{subfigure}{0.6\textwidth}
      \includegraphics[width=\textwidth]{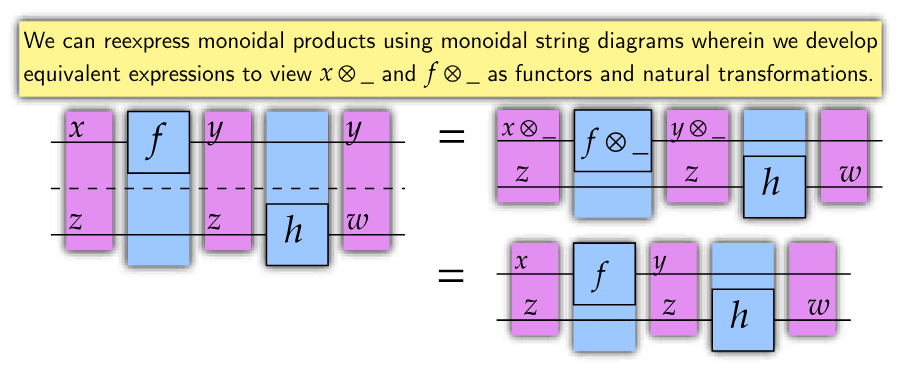}
      \caption{}
    \end{subfigure}         
    \caption[Functor string diagrams]{Functor string diagrams associated with (a) a functor, (b) a natural transformation, (c) a product bifunctor, (d) a monoidal product. (Images and annotations taken from \cite{abbott2024functor}.)}
    \label{fig: functorstringdiagrams}
  \end{center}
\end{figure}

In accordance with these principles, \cite{abbott2023robust} introduces notation that represents a functors as wires that lie above objects wires and natural transformations as boxes that lie above morphism boxes, as in \textit{Fig. \ref{fig: functorstringdiagrams} (a), (b)}. In standard string diagrams, a wire running through a morphism column is interpreted as an identity morphism. Similarly, in the notation of \cite{abbott2023robust}, a wire running where a natural transformation is supposed to be is interpreted as an identical natural transformation. This diagrammatic notation is already effective enough to write a graphical proof of the Yoneda lemma, and can be easily adapted to represent product bifunctors: it suffices to stack two rows (one per category envolved) and to seprate them using a double dashed line, as in \textit{Fig. \ref{fig: functorstringdiagrams} (c)}. Finally, monoidal products can be represented by using a single dashed separation line or no line at all (as in \textit{Fig. \ref{fig: functorstringdiagrams} (d)}). These pieces of diagrammatic notation are also in perfect compliance with the vertical section decompositionl principle and the equivalent expression principle.

\subsection{Neural circuit diagrams}

Neural circuit diagrams are a specialization of functor string diagrams designed to represent precise schematics for deep neural network architectures. The idea is to capture all information relevant to the implementation and analysis of the aforementioned architecture without using any specific programming framework. In this sense, neural circuits diagrams can be compared to pseudocode. Although neural circuit diagrams can be learned and applied without any reference to category theory (this is the perspective adopted in \cite{abbott2024neural}), the fact that these diagrams are just an instance of the wider class of functor string diagrams is not just a curiosity, but proof that they have rigorous categorical semantics. In other words, every well-formed neural circuit diagrams corresponds to a well-defined neural network, and neural circuit diagrams can be composed intuitively.

Differently from the string diagrams seen in e.g. \cite{gavranovic2024fundamental} or \cite{wilson2022categories}, neural circuit diagrams explicitly keep track of dimensionality and indexing. This allows them to represent complex architectures that involve convolutions, residuality, and so on. Crucially, neural string diagrams provide convenient notation to represent broadcasting, which is of fundamental importance in deep learning but is rarely adequately represented by ad-hoc diagrams. 

\begin{figure}[h]
  \begin{center}
    \begin{subfigure}{0.7\textwidth}
      \includegraphics[width=\textwidth]{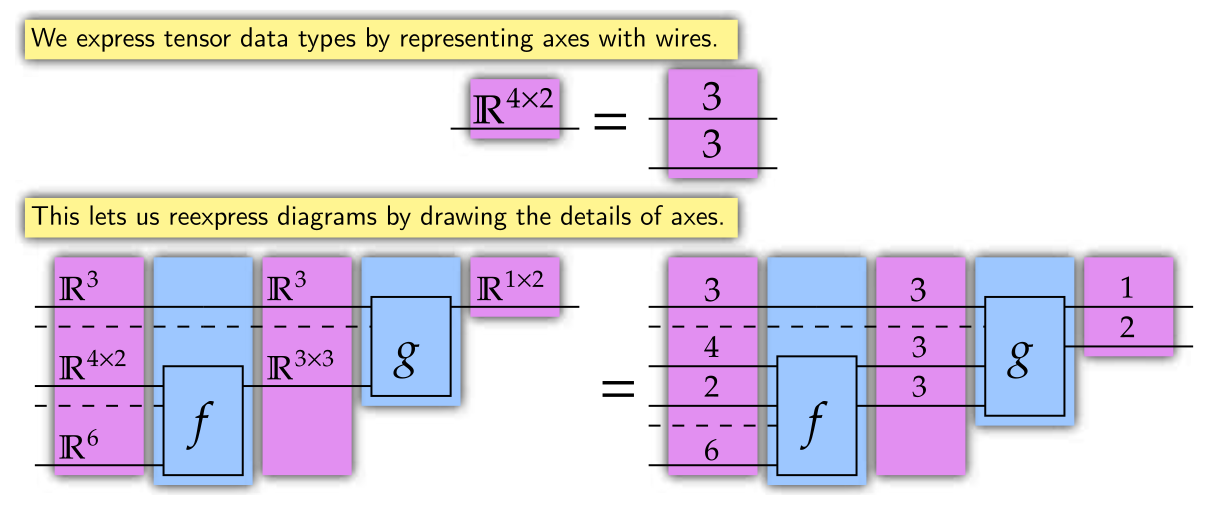}
      \caption{}
    \end{subfigure}
    \begin{subfigure}{0.7\textwidth}
      \includegraphics[width=\textwidth]{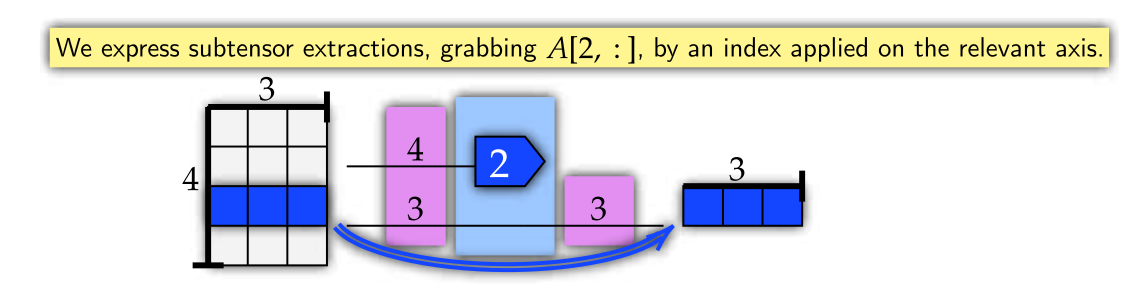}
      \caption{}
    \end{subfigure}
    \begin{subfigure}{0.7\textwidth}
      \includegraphics[width=\textwidth]{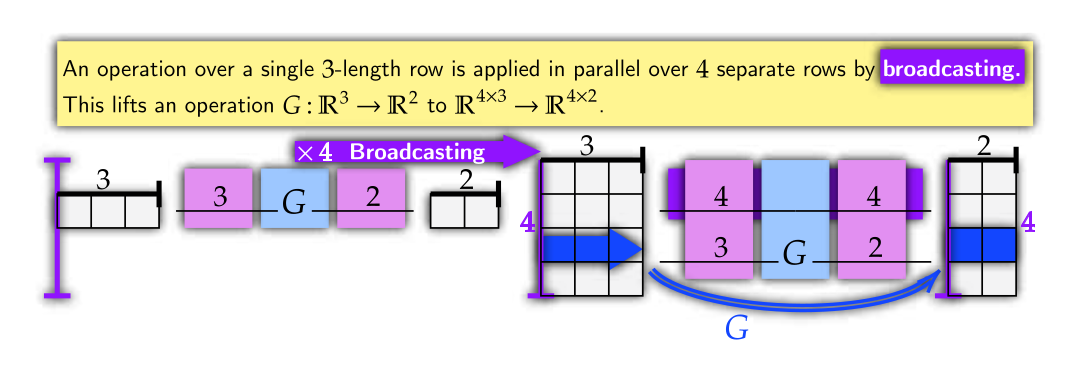}
      \caption{}
    \end{subfigure}      
    \caption[Neural circuit diagrams]{Neural circuit diagram formalisms for (a) representation of tensors, (b) indexing of tensors, (c) broadcasting. (Images and annotations taken from \cite{abbott2024neural}.)}
    \label{fig: neuralstringdiagrams}
  \end{center}
\end{figure}

\begin{figure}[h]
  \begin{center}
    \includegraphics[width=0.7\textwidth]{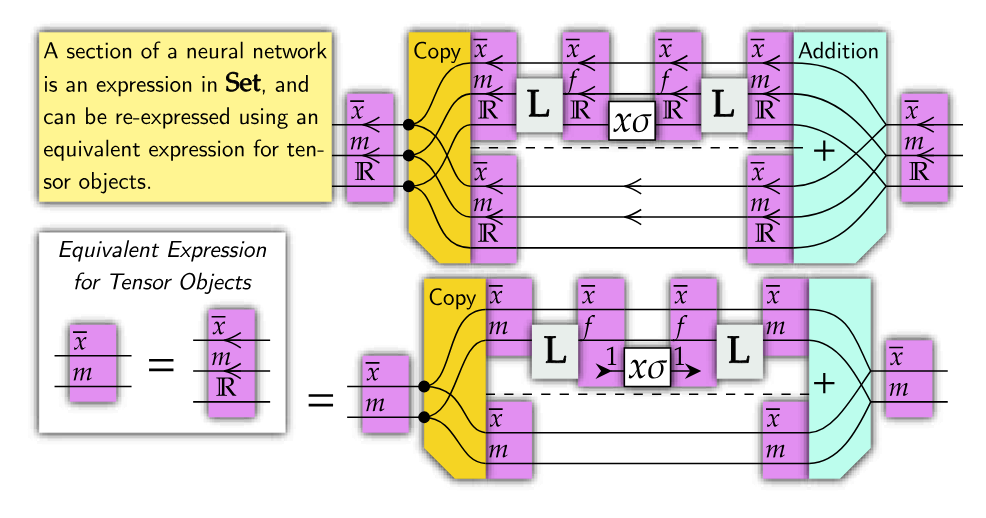}     
    \caption[From functor string diagrams to neural circuit diagrams]{From functor string diagrams to neural circuit diagrams. (Image taken from \cite{abbott2024functor}.)}
    \label{fig: fromfsdtoncd}
  \end{center}
\end{figure}

Neural circuit diagrams represent tensor axes as parallel wires and decorate each wire with the associated number of dimensions (see \textit{Fig. \ref{fig: neuralstringdiagrams} (d)}). Indexing is implemented by assigning a specific value to the wire representing the axis in question (see \textit{Fig. \ref{fig: neuralstringdiagrams} (b)}). Finally, broadcasting is represented by running a wire that represents the additional axis over the morphism that is to be broadcasted (see \textit{Fig. \ref{fig: neuralstringdiagrams} (c)}). Mathematically, tensors are constructed as hom-functors over the real line and broadcasting is just the action of these functors on morphisms (\textit{Fig. \ref{fig: fromfsdtoncd}}). This is all representable using functor string diagrams: neural circuit diagrams are just the result of omitting unnecessary detail from the latter. See \cite{abbott2024neural} for more information regarding neural circuit diagrams and see \cite{abbott2023robust} for more information regarding the relationship between neural circuit diagrams and functor string diagrams.

\begin{figure}[h]
  \begin{center}
    \includegraphics[width=0.85\textwidth]{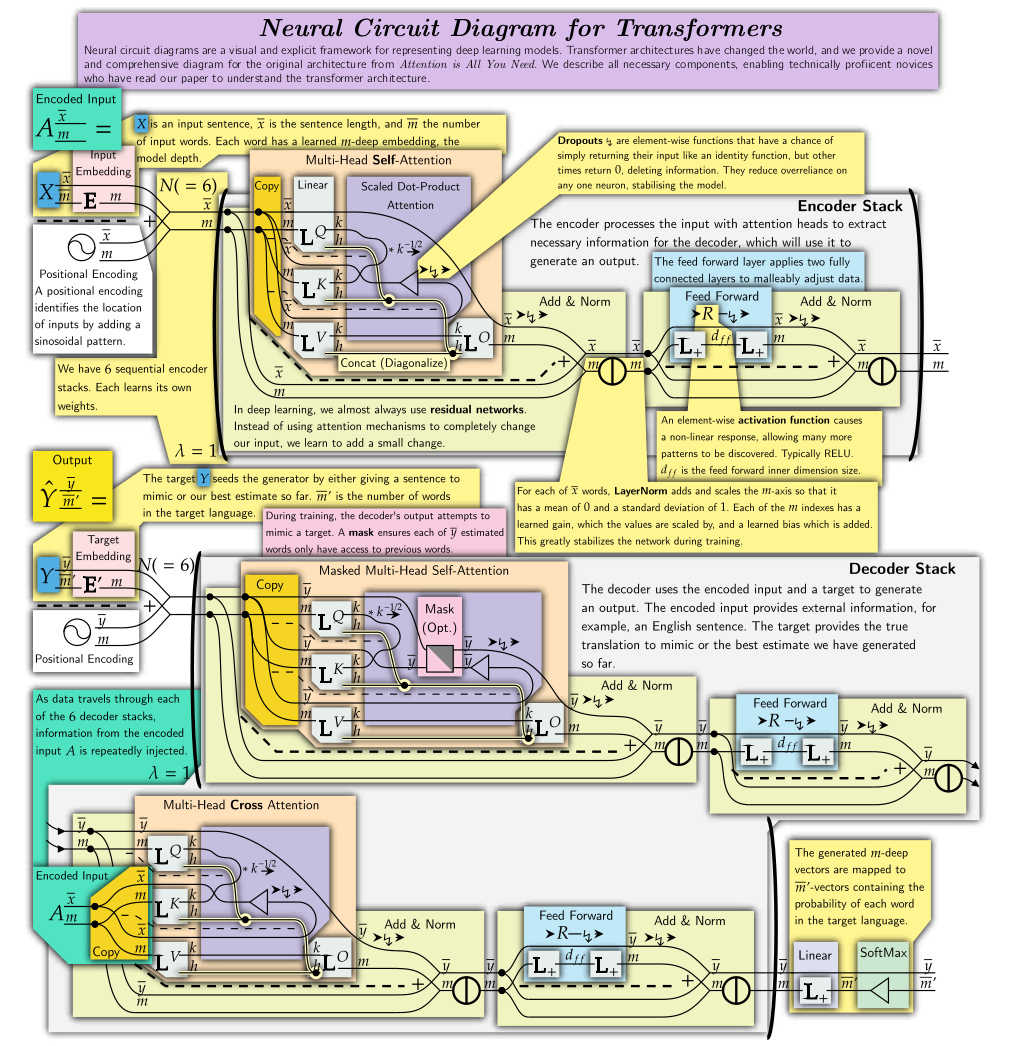}     
    \caption[Neural circuit diagram of a transformer]{Neural circuit diagram of a transformer. (Image taken from \cite{abbott2023robust}.)}
    \label{fig: transformerncd}
  \end{center}
\end{figure}

\cite{abbott2024neural} proves the usefulness of neural circuit diagrams by drawing schematics (\textit{Fig. \ref{fig: transformerncd}}) for the powerful transformer architecture proposed by \cite{vaswani2017attention}. Comparing these schematics with the ones presented in the original paper (\textit{Fig. \ref{fig: transformer}}) shows the wealth of detail that is easily encoded by the former but is completely absent from the latter. Neural circuit diagrams for additional architectures can be found in \cite{abbott2024neural}.

\section{Diagrams with universal approximators}

\cite{khatri2024anatomy} also develops a string-diagrammatic language capable of giving detailed unambiguous descriptions of neural network architectures. It is only natural to compare the approach of \cite{khatri2024anatomy} with the neural circuit diagrams of \cite{abbott2023robust} discussed in the previous section. Despite their unquestionable similarity, the two approaches differ on one key point: while neural circuit diagrams leave little space for abstraction and only allow equational rewrites, \cite{khatri2024anatomy} develops a semantics based on the notion of universal approximator that allows to abstract away details and compare different architectures based on shared abstractions. In this section, we give a brief overview of the approach of \cite{khatri2024anatomy} and we discuss interesting experimental results presented by the paper. We do not discuss the precise categorical semantics of the diagrams introduced \cite{khatri2024anatomy} because we lack the space necessary to do so, but the reader can find a full account of them in the appendices of the original paper. 

\subsection{The diagrammatic approach of \cite{khatri2024anatomy}}

Neural circuit diagrams effectively describe neural network architectures, but the aboundance of detail they convey also means that it can be challenging to extract the fundamental structure of an architecture from its complete schematics.
\cite{khatri2024anatomy} attacks this problem by drawing different diagrams for different levels of abstraction so that complete schematics (like the ones encoded in neural circuit diagrams) become only the end of a spectrum that begins with a purely abstract black box. The same fundamental grammar underpins all these diagrams, and appropriate rewrites can be used to seamlessly move between layers of abstraction. 

\begin{figure}[h]
  \begin{center}
    \includegraphics[width=\textwidth]{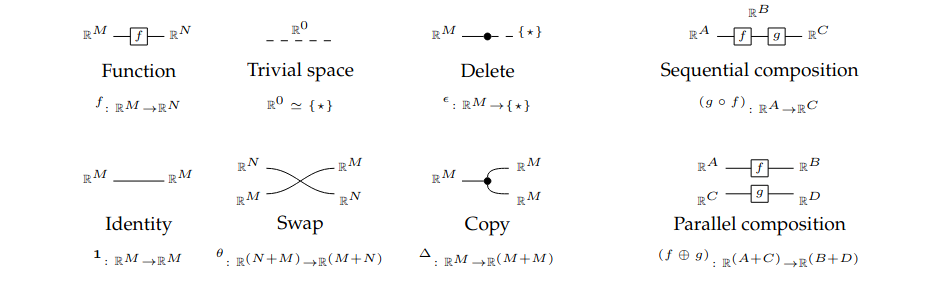}
    \caption[Diagrammatic syntax of \cite{khatri2024anatomy}]{Basic elements of the diagrammatic syntax proposed by \cite{khatri2024anatomy}. (Image taken from \cite{khatri2024anatomy}.)}
    \label{fig: khatribasicdiagrams}
  \end{center}
\end{figure}

Let us begin by discussing the most concrete end of the aforementioned specturm. \cite{khatri2024anatomy} describes computations in the Cartesian category $\mathbf{CartSp}$ of Euclidean spaces and continuous functions, and follows the basic conventions of Cartesian monoidal string diagrams. A sketch of the language can be found in \textit{Fig. \ref{fig: khatribasicdiagrams}}. Unlike neural circuit diagrams, the diagrams of \cite{khatri2024anatomy} represent broadcasting using \textit{SIMD boxes (Same Instruction applied to Multiple Data)}, (\textit{Fig. \ref{fig: khatribasicdiagrams} (b)}), which are also used to represent tensor contractions (\textit{Fig. \ref{fig: khatribasicdiagrams} (c)}). Finally, syntactic sugar is offered to represent reshaping operations (\textit{Fig. \ref{fig: khatribasicdiagrams} (c)}).

\begin{figure}[h]
  \begin{center}
    \begin{subfigure}{0.7\textwidth}
      \includegraphics[width=\textwidth]{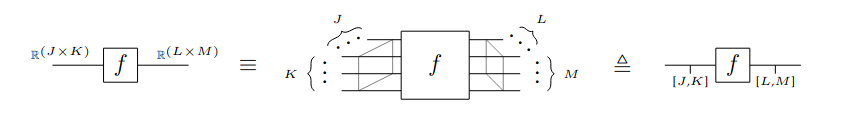}
      \caption{}
    \end{subfigure}  
    \begin{subfigure}{0.9\textwidth}
      \includegraphics[width=\textwidth]{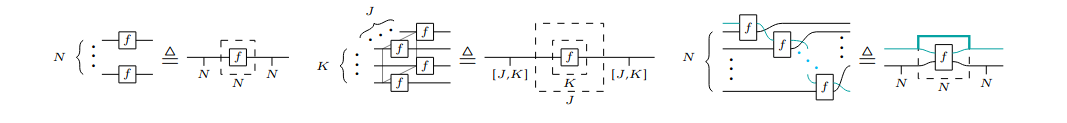}
      \caption{}
    \end{subfigure}    
    \begin{subfigure}{0.9\textwidth}
      \includegraphics[width=\textwidth]{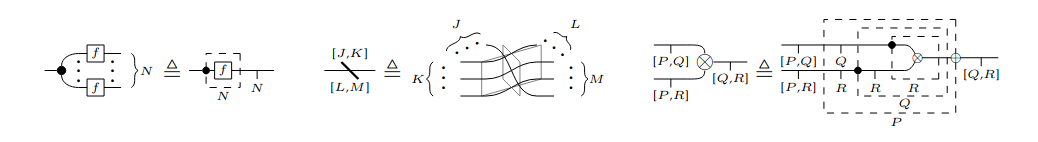}
      \caption{}
    \end{subfigure}    
    \caption[Dimensionality, SIMD boxes, and syntactic sugar]{Syntax proposed by \cite{khatri2024anatomy} for (a) dimensionality, (b) SIMD boxes, (c) copys, reshape operators, and contractions. (Images taken from \cite{khatri2024anatomy}.)}
    \label{fig: khatridiagrams}
  \end{center}
\end{figure}

\begin{figure}[h]
  \begin{center}
    \includegraphics[width=\textwidth]{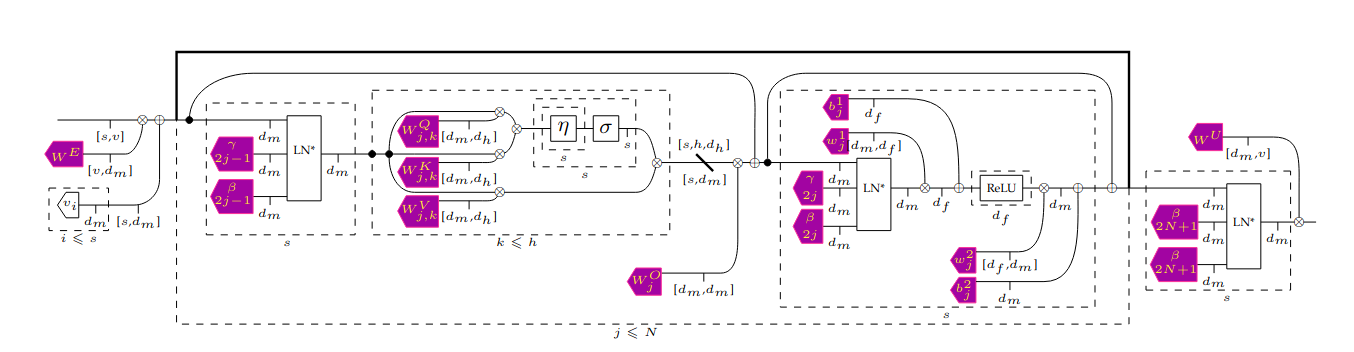}
    \caption[Transformer blueprint according to \cite{khatri2024anatomy}]{Blueprint of the encoder part of a transformer according to \cite{khatri2024anatomy}. (Image taken from \cite{khatri2024anatomy}.)}
    \label{fig: khatritransformer}
  \end{center}
\end{figure}

The reader can compare \textit{Fig. \ref{fig: transformerncd}} against \textit{Fig. \ref{fig: khatritransformer}} to see how both neural circuit diagrams and the approach of \cite{khatri2024anatomy} can be used to describe the transformer architecture in detail. While both formalisms shine in accuracy of representation, they also both suffer from low readibility: one might ask whether adopting such systems is even worth the trouble given that simpler diagrams such as the one in \textit{Fig. \ref{fig: transformer}} seem to be better at communicating basic ideas despite their lack of detail. \cite{abbott2023robust} compensates for this by annotating its diagrams and by encapsulating portions of the architectures at hand into dedicated boxes. On the other hand, \cite{khatri2024anatomy} proposes to abstract away detail by replacing portions of the graph representing specific implementations with filled boxes that represent universal approximators with the same inputs and outputs.

\begin{figure}[h]
  \begin{center}
    \includegraphics[width=0.5\textwidth]{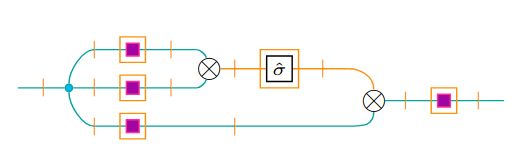}
    \caption[Abstract blueprint for self-attention]{Abstract blueprint for the self-attention mechanism for the transformer architecture of \cite{vaswani2017attention}. (Image taken from \cite{khatri2024anatomy}.)}
    \label{fig: khatriattention}
  \end{center}
\end{figure}

\begin{definition}[Universal approximators]
  A universal approximator for a class $\mathcal{C}$ of functions is a neural network architecture able to approximate any element of $\mathcal{C}$ to any desired precision.
\end{definition}

Many well known results concern the existence and characterization of such approximators (see e.g. \cite{cybenko1989approximation}). \cite{khatri2024anatomy} uses universal approximators to rigorously represent \dblquote{typed holes} that need to be filled with an appropriate operator. Replacing detailed structures with these holes allows us to climb the ladder of abstraction. This can be combined with the omission of dimensional information that is not necessary to understrand the architecture at hand, but only to implement it. \cite{khatri2024anatomy} showcases this process in the diagram of \textit{Fig. \ref{fig: khatriattention}}, which highlights the fundamental structure of the attention mechanism proposed by \cite{vaswani2017attention}. Notice how colors are used to distringuish between dimensions that are fixed as hyperparameters and dimensions that vary with the length of the sequence that is being processed.

\subsection{Comparing attention mechanisms}

\begin{figure}[h]
  \begin{center}
    \includegraphics[width=0.7\textwidth]{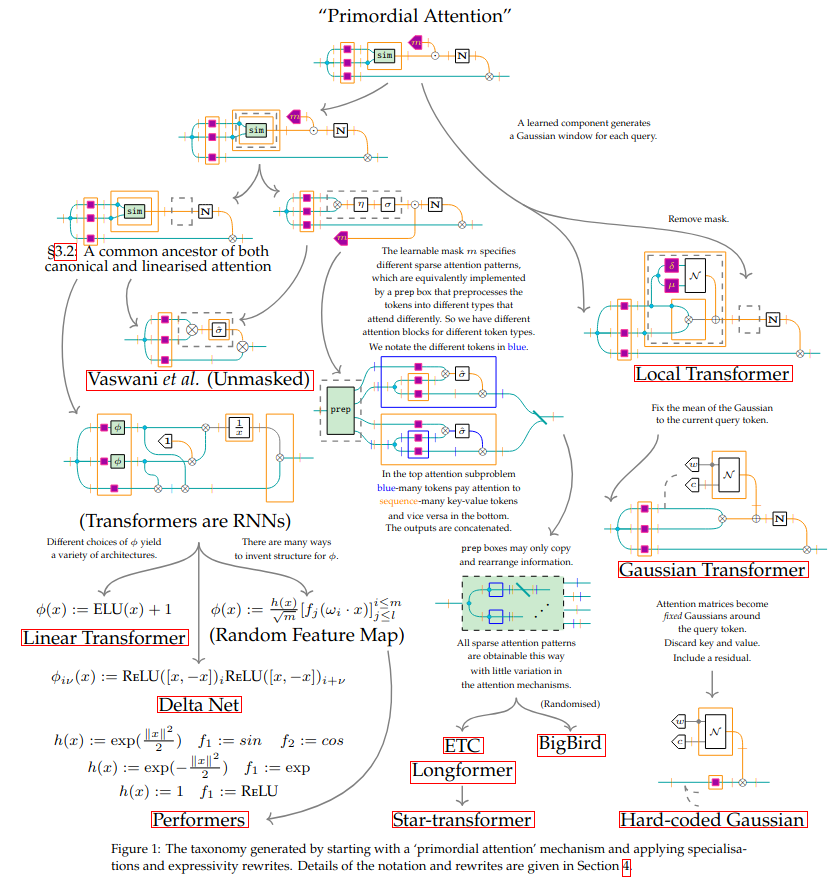}
    \caption[Tree of attention]{Evolutionary tree of attention mechanisms according to \cite{khatri2024anatomy}. Child nodes are obtained by applying rewrites to parent nodes. (Image taken from \cite{khatri2024anatomy}.)}
    \label{fig: attentiontree}
  \end{center}
\end{figure}

The universal approximator abstraction can also be used to compare different architectures. Two architectures that only differ at a point can be considered as descendents of a commmon ancestor architecture, and this common link allows us to compare them. For instance, \cite{khatri2024anatomy} derives all known attention mechanisms as descendents of a prototype \dblquote{primordial attention}, forming a sort of evolutionary tree of attention (see \textit{Fig. \ref{fig: attentiontree}}). The process of filling in a universal approximator with a more specialized structure is called by \cite{khatri2024anatomy} an \dblquote{expressivity reduction}, as it reduces number of architectures represented by the given schematics. Expressivity reductions can be formalized as string diagrammatic rewrites, which implies that the categorical semantics of these diagrams have a central role in the comparative analysis of architectures.

Just like biological taxonomy classifies species based on their evolutionary history, we can use the tree in \textit{Fig. \ref{fig: attentiontree}} as a blueprint for the classification of attention mechanisms. This undertaking is analogous to the one of \cite{gavranovicposition}, where recurrent neural network cells are classified based on the classical data structures and automata they are related to. Both taxonomies serve to systematize the scattered panorama of machine learning research, but may also inspire new architectures. The authors of \cite{khatri2024anatomy} pursue this line of thinking and explore a space of possible attention mechanisms freely constructed from five generators (represented as abstract diagrams). This large space is reduced in size by rewriting sections that would lead to overparameterization. This results in a selection of eleven possible attention mechanisms, which the authors of \cite{khatri2024anatomy} go on to test: the mechanisms are embedded in a decoder-only transformer and trained on a Penn-Treebank task. Interestingly, the 11 mechanisms (among which there are known architectures such as the one proposed by \cite{vaswani2017attention}) all perform comparably well, with only one having a slightly better performance than the others. The authors entertain the conjecture that the aforementioned results might be a sign that the choice of attention mechanism matters little, as long as data is exchanged between tokens. While there is no uncontrovertible evidence that this is the case, these experiments prove the usefulness of the approach of \cite{khatri2024anatomy} and, more generally, of the usefulness of string diagrams in machine learning.

\section{Future directions and related work}

The amount of implementational detail conveyed by neural circuit diagrams and the diagrams in \cite{khatri2024anatomy} suggests that it should be possible to find a 1-to-1 correspondence between these diagrams and code. It is even argued in \cite{abbott2023robust} that neural circuit diagrams are expressive enough to be considered a high-level language with explicit memory management. Hopefully, future research will lead to the creations of tools that convert neural circuit diagrams into working models, not unlike the library developed by \cite{cruttwell2022categorical} that allows programmers to code using parametric lenses. Another interesting direction of research is the application of methods analogous to the ones of \cite{khatri2024anatomy} to the exploration of other classes of architectures.

We conclude the chapter by briefly discussing a few alternative approaches to the problem of unambiguous representation of neural networks. While these approaches are not categorical (yet), we believe there are fruitful comparisons to be drawn between the latter and the categorical approaches discussed in the previous sections. We also believe that these all these methods can be fruitfully integrated on various level, as discussed below.

\subsection{Tensor networks}

The enormous importance of tensors in quantum physics lead to the development of tensor networks (pioneered by \cite{penrose1971applications}), in which tensor operations are represented as graphs, where tensors are nodes and modes are represented as edges coming out of the nodes. Given the importance of tensors in deep learning, it makes sense to try to adapt tensor networks to represent neural networks. This is done in \cite{xu2023graph} and \cite{taylor2024introduction}: the two papers introduce similar conventions that endow tensor networks with dedicated notation for non-linear activation functions and copy maps, which are not present in ordinary tensor networks. 

\begin{figure}[h]
  \begin{center}
    \includegraphics[width=0.9\textwidth]{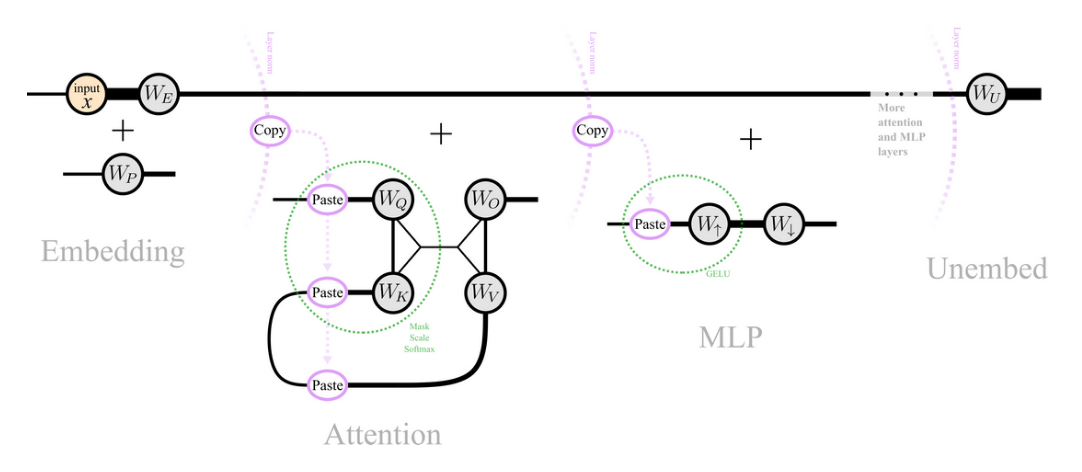}
    \caption[GPT-2 architecture]{Tensor network representing the GPT-2 architecture. (Image taken from \cite{taylor2024introduction}.)}
    \label{fig: gpt2}
  \end{center}
\end{figure}

The main difference between tensor networks and the string diagrams described in this chapter is that the latter represente data as boxes and processes as wires, whereas the former do the opposite. In tensor network an edge linking two tensors represents a tensor contradction: this choice yields intuitive representations for sophisticated tensor operations such as the Tucker product (\cite{xu2023graph}) and for various tensor decompositions tecniques (\cite{taylor2024introduction}), but it also makes it much more difficult to conceptualize how data moves through a model. For instance, the reader can observe \textit{Fig. \ref{fig: gpt2}} to understand how much more difficult it is to represent copy maps in the notation of \cite{taylor2024introduction}. Moreover, while the diagrams of \cite{abbott2023robust} and \cite{khatri2024anatomy} have sophisticated categorical semantics, the notation of \cite{xu2023graph} and \cite{taylor2024introduction} mainly relies on human intuition\footnote{We see no reason why the diagrams in \cite{xu2023graph} and \cite{taylor2024introduction} should not be interpreted under the lens of category theory. Refer to \cite{biamonte2011categorical} for a categorical approach to tensor networks.}.

Despite their weaknesses, tensor networks can be of great usefulness to deep learning practictioners as they can offer clear insight into tensorization processes. Tensorization is an architecture design patter that aims to reduce the number of parameters of a neural network without compromising its expressivity: this can be done by replacing large matrices with appropriately decomposed low-rank tensors. Owing to the redundance of ordinary fully connected layers (see e.g. \cite{sainath2013low}), this strategy leads to remarkable levels of compression (\cite{novikov2015tensorizing}, \cite{xu2023graph}). The improved efficiency comes at the price of human intuition, as it is much more difficult to think in terms of tensor operations across many modes than in terms of matrix-vector multiplication (\cite{xu2023graph}). Diagrams can help bridge this gap.

\subsection{Non-diagrammatic notations}

An alternative to diagrammatic notation is the enhancement of algebraic notation. For instance, \cite{chiang2021named} proposes to give English labels to tensor axes to make it easier to keep track of them. Following this convention, known as \textit{named tensor notation}, the dot-product self-attention mechanism of \cite{vaswani2017attention} can be represented with the equation (which we lift from \cite{chiang2021named}):
\begin{equation}
  \label{eq: attention}
  \mathrm{Attention(Q,K,V) = \left(\nbin{\name{seq}}{\mathrm{softmax}}\frac{Q \ndot{\name{key}} K}{\sqrt{|\mathsf{key}|}}\right)} \ndot{\name{seq}} V,
\end{equation}
where $\mathsf{key}$ represents the axes dedicated to query and key features, and $\mathsf{seq}$ represents the axis of sequence tokens. The authors argue that their named tensor notation is an improvement over the popular Einstein notation for sums as axes with English names allow the reader to immediately understand what kind of information they encode; moreover, they claim that named tensor notation limits repetition of indices and thus makes equations cleaner.

Another viable alternative approach is described in \cite{phuong2022formal}, where it is suggested that neural network architectures should be represented using pseudocode analogous to the one used in algorithm design and analysis (see e.g. \cite{cormen2022introduction}). The authors claim that writing pseudocode distills each architecture down to its fundamental components and removes incidental distractions, all the while preserving the detail necessary for implementation. One disadvantage of using pseudocode is that the structure of the involved tensors and the flow of information are partially obscured, which is detrimental to intuition. More generally, one can argue that pseudocode is readily implementable but hides away mathematics and structure.

\subsection{A multifaceted approach}

In our opinion, the problem of representing neural network architectures is best solved by using a combination of methods. For instance, there is no reason (aside maybe from space concerns that could be mitigated by the use of appendices) why the same architecture cannot be described with a string diagram, a tensor network, named tensor equations, pseudocode, and even code in the same paper. One or more string diagrams might be designed using either of the two approaches described in this chapter; tensor networks and named tensor equations might be employed to clarify the dynamics of complex tensor operations\footnote{Since the edges of tensor networks represent axes, we see no reason why they cannot be themselves labelled, effectively integrating the approaches of \cite{xu2023graph} and \cite{chiang2021named}.}; pseudocode could be provided so that practitioners can readily implement the architecture; finally, providing source code might clarify any ambiguity left by these high-level representations.

\clearpage{\pagestyle{empty}\cleardoublepage}


\chapter*{Conclusions}
\rhead[\fancyplain{}{\bfseries
Conclusions}]{\fancyplain{}{\bfseries\thepage}}
\lhead[\fancyplain{}{\bfseries\thepage}]{\fancyplain{}{\bfseries Conclusions}}
\addcontentsline{toc}{chapter}{Conclusions} 

When discussing applications of category theory outside of pure mathematics, one might ask whether formalizing a concept in categorical terms is really worth the effort. Does a categorical formalization yield any novel insight? Does there exist some aspect of the subject at hand which can be analyzed with category theory more easily or elegantly than without?  Experiments confirm that it possible to elegantly model gradient-based learning with parametric lenses; they confirm that thinking in terms of integral transforms leads to better algorithmic alignment; finally, they confirm that functor learning techniques preserve expressivity while reducing the number of parameters. It also appears clear that string diagrams can effectively describe neural network architectures. Hence, in the light of the many categorical approaches discussed in this thesis, we believe that categorical thinking can indeed be productive even in an applied field like deep learning.

Category theory is not the only area of mathematics that has been proposed to attack the open problems of deep learning and machine learning in general. For instance, both topology (see e.g. \cite{hensel2021survey}) and mathematical physics (see e.g. \cite{roberts2022principles}) seem to be good contenders, and linear algebra and probability theory are so foundational to the field that it is hard to imagine that they will not yield important advancements in the future. However, category theory holds a position of high regard in the hierarchy of mathematics because it can be used to build bridges between disparate fields. Thus, while no prediction about the future can be certain, it seems reasonable to expect that, on top of providing important techniques on its own, category theory will be used to build bridges between non-categorical approaches. Hopefully, bringing all these contributions together will lead us to a general theory of deep learning and machine learning at large.

\clearpage{\pagestyle{empty}\cleardoublepage}


\bibliographystyle{alpha}
{\bibliography{references}}
\rhead[\fancyplain{}{\bfseries \:Bibliography}]{\fancyplain{}{\bfseries\thepage}} 
\addcontentsline{toc}{chapter}{Bibliography}

\clearpage{\pagestyle{empty}\cleardoublepage}


\chapter*{Acknowledgements}

\thispagestyle{empty}

I wish to acknowledge the essential role my advisor Professor F. Zanasi has had in guiding me through the process of writing this thesis. He introduced me to applied category theory, to machine learning, and to the world of academic research in general. I will be forever grateful for it. I also wish to thank my family and my friends, who supported me throughout my academic journey, and without whom I would not have been here writing. To all of you, my most sincere gratitude.

\end{document}